\documentclass[12pt]{article}

\makeatletter
\def\newleaf{\newpage
\newcount\tmp
\tmp=\c@page
\divide\tmp by 2
\multiply\tmp by 2
\ifnum\c@page=\tmp
~\newpage
\fi
}
\makeatother

\expandafter\ifx\csname proofnewpage\endcsname\relax

\fi

\def\color[#1]#2{}

\long\def\nop#1{}

\def\comment{\edef\cps{\the\parskip} \parskip=0.5cm \begingroup \tt}

\def\noproof#1{
\def\proof{{#1}\iffalse}
\let\qed=\fi
}

\begingroup
\makeatletter
\global\def\fakelabel#1#2{
\expandafter\ifx\csname fakelabelsome\endcsname\relax
\let\fakelabelsome\par
\AtEndDocument{\typeout{}}
\fi
\AtEndDocument{\typeout{NOTE: #1 is a fake label, marked #2}\typeout{}}
\@ifundefined {r@#1}
{\global\@namedef{r@#1}{#2}}
{}
}
\endgroup

\long\def\figurearrow#1#2{
\newbox\before
\setbox\before=\hbox{#1}
\newbox\after
\setbox\after=\hbox{#2}
\newdimen\vdim
\vdim=\ht\before
\ifdim\vdim<\the\ht\after
  \vdim=\ht\after
\fi
\hbox{
\vbox to \the\vdim{\vfill\box\before\vfill}
\vbox to \the\vdim{\vfill\hbox to 3cm{\hfill\Huge$\Rightarrow$\hfill}\vfill}
\vbox to \the\vdim{\vfill\box\after\vfill}
}
}

\expandafter\ifx\csname shortcite\endcsname\relax

\fi

\newbox\current

\long\def\plframebox#1{
\setbox\current\vbox{#1}		% set box

\vbox to \ht\current {\hrule\vss
\hbox to \wd\current {%
\vrule \hss\box\current\hss \vrule}
\vss\hrule }
}

\long\def\eatpar#1{%
\ifx#1\par                      % se il token e' \par
\let\nextmove=\eatpar           % rimetti \eatpar in coda
\else
\let\nextmove=#1%               altrimenti, rimetti il token in coda
\fi
\noexpand\nextmove%             il token o \eatpar viene rimesso in coda
}

\def\modifymargins#1#2{
\newdimen\addtoh
\newdimen\addtow
\addtoh=#1
\addtow=#2

\advance\topmargin by -\addtoh
\multiply\addtoh by 2
\advance\textheight by \addtoh

\advance\oddsidemargin by -\addtow
\advance\evensidemargin by -\addtow
\multiply\addtow by 2
\advance\textwidth by \addtow
}

\begingroup
\catcode`\~=11
\gdef\centertilde#1{\lower #1pt\hbox{~}}
\endgroup

\newcount\currenttime
\newcount\hour
\newcount\minute

\def\printtime{%
\currenttime=\time
\hour=\currenttime
\divide\hour by 60
\minute=-\hour
\multiply\minute by 60
\advance\minute by \currenttime
\the\hour:\ifnum\minute<10 0\fi\the\minute
}

\begingroup
\makeatletter
\global\let\@@date=\@date
\gdef\@date{\@@date\ --- \printtime}
\endgroup

\def\oggi{\number\day\space 
\ifcase\month\or
Gennaio\or Febbraio\or Marzo\or Aprile\or Maggio\or Giugno\or
Luglio\or Agosto\or Settembre\or Ottobre\or Novembre\or Dicembre\fi
\space \number\year}

\newcounter{rmexample}

\def\proof{\noindent {\sl Proof.\ \ }}

\def\qed{\hfill{\boxit{}}
  \ifdim\lastskip<\medskipamount \removelastskip\penalty55\medskip\fi}
\def\qedn#1{\hfill{\boxit{}$_#1$}
  \ifdim\lastskip<\medskipamount \removelastskip\penalty55\medskip\fi}
\long\def\boxit#1{\vbox{\hrule\hbox{\vrule\kern3pt
                  \vbox{\kern3pt#1\kern3pt}\kern3pt\vrule}\hrule}}

\def\true{{\sf true}}
\def\false{{\sf false}}

\def\np{{\rm NP}}

\def\Dp{${\rm D}^p$}

\def\bh#1{\if#1{}{\rm BH}\else\mbox{BH$_{#1}$}\fi}
\def\S#1{\mbox{$\Sigma^p_{#1}$}}

\let\cedilla=\c
\def\c{\mbox{$\leadsto$}}

\def\profont{\sf}

\def\x3c{{\profont x3c}}

\def\possnewtheorem#1#2{
\expandafter\ifx\csname #1\endcsname\relax
\newtheorem{#1}{#2}
\fi
}

\def\possnewtheoremthree#1[#2]#3{
\expandafter\ifx\csname #1\endcsname\relax
\newtheorem{#1}[#2]{#3}
\fi
}

\possnewtheorem{theorem}{Theorem}
\possnewtheorem{corollary}{Corollary}
\possnewtheorem{lemma}{Lemma}
\possnewtheoremthree{proposition}[theorem]{Proposition}
\possnewtheorem{definition}{Definition}
\possnewtheorem{question}{Question}
\possnewtheorem{example}{Example}
\possnewtheorem{nontheorem}{Counterexample}
\possnewtheorem{property}{Property}
\possnewtheorem{assumption}{Assumption}
\possnewtheorem{conjecture}{Conjecture}
\possnewtheorem{notation}{Notation}
\newenvironment{theorem*}[1]{{\noindent \bf Theorem~#1}\begin{it}}{\end{it}\

}

\def\after#1#2{#1~{\sf after}~#2}

\expandafter\ifx\csname ttytty\endcsname\relax\modifymargins{60pt}{40pt}
\else\modifymargins{60pt}{0pt}\def\_{\tt\char 95}
\raggedright\parindent 40pt\fi

\possnewtheorem{algorithm}{Algorithm}

\title{Reconstructing a single-head formula to facilitate logical forgetting}
\author{Paolo Liberatore}

\begin{document}

\maketitle

\iffalse

{\bf fare le cose segnate come "fare:"}

{\bf controllare le cose segnate come "nota:"}

{\bf fare:} rimuovere la sinossi e le mini-sinossi

{\bf nota:} il lemma set-implies-set ha come precondizione $x \not\in P'$,
quindi il contrario $x \in P'$ e' una alternativa

{\bf nota:} SCL, BCL, UCL, ITERATION e il valore di IT non includono
tautologie; questa e' anche una precondizione del lemma set-implies-set e del
lemma di costruzione

{\bf nota:} la definizione di $F^x$ e' $F$ senza le clausole che contengono
$x$, con qualsiasi segno; va bene se viene usato il lemma set-implies-set, che
usa la definizione giusta

{\bf nota:} tutte le formule qui sono definite Horn; in common viene poi
dimostrato che si puo' fare forget di Horn generali riducendole a definite e
poi introducendo delle clausole unitarie negative; essendo negative si possono
rimpiazzare la loro variabili con false

{\bf fare:} articolo a parte per loop segregation

{\bf fare:} verificare che ci siano tutti i file python in tests/ in centre

{\bf fare:} decidere se single-head o singlehead: quale e' usato in
singlehead/?

{\bf fare:} $\leq_F$ e' un preorder o un partial order?

\fi

%input{synopsis.tex}
%input{notation.tex}

\begin{abstract}

Logical forgetting may take exponential time in general, but it does not when
its input is a single-head propositional definite Horn formula. Single-head
means that no variable is the head of multiple clauses. An algorithm to make a
formula single-head if possible is shown. It improves over a previous one by
being complete: it always finds a single-head formula equivalent to the given
one if any.

\end{abstract}

\section{Introduction}

% {\em [general introduction about forgetting]}

Logical forgetting is removing some variables from a logical formula while
preserving its information regarding the others~\cite{lang-etal-03}. Seen from
a different angle, it is restricting a formula to keep only its information
about some of its variables~\cite{delg-17}.

Each of the two interpretations has its own applications. Removing information
allow
{} reducing memory requirements~\cite{eite-kern-19},
{} simplifying reasoning~\cite{delg-wang-15,erde-ferr-07,wang-etal-05}
and
{} clarifying the relationship about variables~\cite{delg-17}.
Restricting a formula on some variables allows
{} formalizing the limited knowledge of agents~\cite{fagi-etal-95,raja-etal-14},
{} ensuring privacy~\cite{gonc-etal-17}
and
{} removing inconsistency~\cite{lang-marq-10}.

Initially studied in first-order logic~\cite{lin-reit-94,zhou-zhan-11}, logical
forgetting has found its way into many logics such as
{} propositional logic~\cite{bool-54,moin-07,delg-17},
{} answer set programming~\cite{wang-etal-14,gonc-etal-16},
{} description logics~\cite{kone-etal-09,eite-06},
{} modal logics~\cite{vand-etal-09},
{} logics about actions~\cite{erde-ferr-07,raja-etal-14},
{} temporal logic~\cite{feng-etal-20},
{} defeasible logic~\cite{anto-etal-12}
and
{} belief revision~\cite{naya-chen-lin-07}.

% {\em [summary of forgetsize]}

Contrary to intuition, forgetting may increase size instead of decreasing
it~\cite{libe-20}. This phenomenon is harmful to the many application scenarios
where size reduction is important if not the sole aim of forgetting: increasing
efficiency of reasoning~\cite{delg-wang-15,erde-ferr-07,wang-etal-05},
summarizing, reusing and clarifying information~\cite{delg-17}, dealing with
information overload~\cite{eite-kern-19}, tailoring information to specific
applications~\cite{eite-06}, dealing with memory bounds~\cite{fagi-etal-95} or
more generally with limitations in the ability of
reasoning~\cite{raja-etal-14}.

To complicate matters, forgetting is not uniquely determined: a specific
algorithm for forgetting may produce a large formula that is equivalent to a
small one. The latter is as good as the former at representing information, but
better in terms of size. The question is not whether forget increases size, but
whether every possible mechanism for forgetting does. Several equivalent
formulae express forgetting the same variables from the same formula $F$. The
question is whether one of them is smaller than $F$, or than a target size
bound.

Giving an answer to this question is harder than the typical problems in
propositional reasoning: it is in \S{3} and is \Dp{2}-hard. The Horn
restriction lowers complexity of one level, to membership in \S{2} and hardness
in \Dp~\cite{libe-20}.

% {\em [Horn case]}

Such a simplification in complexity is typical to the propositional Horn
restriction~\cite{mako-87}. It is one of the reasons that makes it of interest,
the other being its expressivity. Consistency, inference and equivalence only
take polynomial time. At the same time, commonly-used statements like ``if a, b
and c then d'' and ``e, g, and f are impossible at the same time'' are allowed.
The Horn restriction is studied in, for example,
{} description logic~\cite{krot-etal-13},
{} default reasoning~\cite{eite-luka-00},
{} fuzzy logic~\cite{belo-vych-06},
{} belief revision~\cite{zhua-pagn-12},
{} transaction logic~\cite{bonn-kife-94}
and
{} nonmonotonic reasoning~\cite{gott-92-e}.

% {\em [summary of common]}

Restricting to the Horn case lowers the complexity of many operations. Is this
the case for forgetting?

A way of forgetting a variable is to resolve all pairs of clauses over that
variable and to remove the resolved clauses. The required space may be
exponential in the number of variables to forget, even if the final result is
polynomially large.

Something better can be done in the definite Horn case: replace the variable
with the body of a clause with the variable in its head. This nondeterministic
algorithm only requires polynomial working space~\cite{libe-20-a}. It extends
from the definite to the general Horn case~\cite{libe-20-a}.

A restriction that makes this method not only polynomial in space but also in
time is the single-head condition. If clauses do not share heads, the choice of
the clause to replace a variable is moot. What was in general nondeterministic
becomes deterministic. Complexity decreases from nondeterministic polynomial
time to deterministic polynomial time. The problem is tractable.

% {\em [summary of singlehead]}

Whether a formula is single-head or not is easy to check by scanning the
clauses and storing their heads; if one of them is found again, the formula is
not single-head.

It is easy, but misses the semantical nature of forgetting.

Forgetting variables is restricting information to the other variables, and
information is independent on the syntax. Forgetting the same variables from
two equivalent formulae gives two equivalent formulae. The initial information
is the same, what to forget is the same. What results should be the same, apart
from its syntactic form. Forgetting $c$ from
{} $\{a \rightarrow b, b \rightarrow c, c \rightarrow d\}$
and forgetting it from
{} $\{a \rightarrow b, b \rightarrow c, c \rightarrow d, a \rightarrow c\}$
produces equivalent formulae since the two formulae are equivalent. Yet, only
the former is single-head. Only the former takes advantage of polynomiality in
time. This needs not to be, since the result is the same. The second formula
can be translated into the first, making forgetting polynomial.

Every formula that is equivalent to a single-head formula is amenable to this
procedure: make it single-head, then forget.

The second step, forgetting, is easy since the formula is single-head. What
about the first?

While checking whether a formula is single-head is straightforward, checking
whether it is equivalent to a single-head formula is not. Many formulae are
equivalent to a given one and are therefore to be checked for multiple heads.
The problem would simplify if it amounted to checking a simple condition on the
formula itself or its models rather than all its equivalent formulae, but
finding such a condition proved difficult~\cite{libe-20-b}.

A polynomial algorithm that often turns a formula in single-head form if
possible is to select the minimal bodies for each given head according to a
certain order. It is polynomial but incomplete: some single-head equivalent
formulae are not recognized as such~\cite{libe-20-b}.

% {\em [summary of this article]}

The present article restarts the quest from a different direction: instead of
fixing the heads and searching for the bodies, it fixes the bodies and searches
for the heads. The overall algorithm is complete: it turns a formula in
single-head form if possible. The price is its running time, which may be
exponential.

This may look discouraging since the target is an efficient algorithm for
forgetting. Two points are still in favor of the new algorithm: first, it is
exponential only when it works on equivalent sets of variables; second, each
step can be stopped at any time, making the algorithm incomplete but fast.

The first point derives from the structure of the algorithm: it is a loop over
the bodies of the formula; for each, its equivalent sets of variables are
candidate bodies for clauses of the single-head form. The overall loop
comprises only a polynomial number of steps. Each step is exponential because a
formula may make a set of variables equivalent to exponentially many others. If
not, the algorithm is polynomial in time.

The second point is a consequence of what each step is required to do: change
some clauses of the original formula to make each head only occurs once. If
such an endeavor turns out to be too hard, the original clauses can be just
left unfiltered. Still better, some original clauses may be added to complete
the best set found so far. This part of the output formula will not be
single-head, but the rest may be. The forgetting algorithm may still benefit
from the change, since it is exponential in the number of duplicated heads.

% {\em [the sections]}

The article is organized as follows. Section~\ref{previously} restates the
definitions and results obtained in previous
articles~\cite{libe-20-a,libe-20-b}. Section~\ref{construction} presents the
overall algorithm. It repeatedly calls a function implemented in
Section~\ref{iteration}. This completes the algorithm, which is correct and
complete but not always polynomial-time. Given that a previous algorithm was
incomplete but polynomial, the question is whether the problem itself is
tractable; this is investigated in Section~\ref{future}. The Python
implementation of the program is described in Section~\ref{python}.

\section{Previously}
\label{previously}

A summary of the past work on single-head equivalence is in the Introduction.
This section contains the technical definitions and results of previous
articles~\cite{libe-20-a,libe-20-b} used in this article.

The first one is a quite obvious lemma that proves that entailing a variable is
only possible via a clause that has that variable as its head~\cite{libe-20-a}.

\begin{lemma}
\label{set-implies-set}

If $F$ is a definite Horn formula, the following three conditions are
equivalent, where $P' \rightarrow x$ is not a tautology ($x \not \in P'$).

\begin{enumerate}

\item $F \models P' \rightarrow x$;

\item $F^x \cup P' \models P$ where $P \rightarrow x \in F$;

\item $F \cup P' \models P$ where $P \rightarrow x \in F$.

\end{enumerate}

\end{lemma}

An order over the sets of variables~\cite{libe-20-a} is central to the
algorithm presented in the present article.

\begin{definition}
\label{leq-f}

The ordering $\leq_F$ that compares two sets of variables where $F$ is a
formula is defined by $A \leq_F B$ if and only if $F \models B \rightarrow A$.

\end{definition}

The set $BCN(B,F)$ contains all consequences of the set of variables $B$
according to formula $F$~\cite{libe-20-a}.

\begin{definition}
\label{bcn}

The set $BCN(B,F)$, where $B$ is a set of variables and $F$ is a formula, is
defined as
{} $BCN(B,F) = BCN(B,F) = \{x \mid F \cup B \models x\}$.

\end{definition}

The intended meaning of $BCN(B,F)$ is the set of consequences of $B$. Formally,
$BCN(B,F)$ meets its aim. Informally, a consequence is something that follows
from a premise. Some deduction is presumed involved in the process. Yet,
$BCN(B,F)$ may contain variables that do not follow from $B$, they are just in
$B$. No inference ever derives it from $B$.

Excluding all variables of $B$ is not a correct solution: $BCN(B,F) \backslash
B$ is too small. It does not include the variables of $B$ that follow from $B$
thanks for some derivation.

For example, both $x$ and $y$ are in $BCN(B,F)$ if $B = \{x,y\}$ and $F = \{y
\rightarrow z, z \rightarrow y\}$. Yet, $x$ is in $BCN(B,F)$ only because it is
in $B$. Instead, $y$ can be deduced from $B$ via $y \rightarrow z$ and $z
\rightarrow y$. This distinction proved important to the previous, incomplete
algorithm for single-head equivalence~\cite{libe-20-b}.

\begin{definition}
\label{rcn}

For every set of variables $B$ and formula $F$, the {\em real consequences} of
$F$ are
{} $RCN(B,F) = \{x \mid F \cup (BCN(B,F) \backslash \{x\}) \models x\}$.

\end{definition}

The set of all consequences of $B = \{x,y\}$ is $BCN(B,F) = \{x,y,z\}$. The
difference between $x$ and $y$ is that $x$ cannot be recovered once removed
from it while $y$ can. In other words, $B$ derives something that derives $y$
back; it does not for $x$. Therefore, $y$ a real consequence, $x$ is not:
$RCN(B,F) = \{y,z\}$.

This example also shows how to determine $RCN(B,F)$ when $F$ is a definite Horn
formula: start from $B$ and add to it all variables $x$ such that $B'
\rightarrow x \in F$ with $B' \subseteq B$. Only the variables that are added
form $RCN(B,F)$. The algorithm is presented in full and proved correct in the
same article where $RCN(B,F)$ is introduced~\cite{libe-20-b}. The algorithm
also determines the set of clauses used in the process:
{} $\{B' \rightarrow x \in F \mid F \models B \rightarrow B'\}$.
The variant of this set where tautologies are excluded is defined in the
present article as $UCL(B,F)$. Since this set only contains the clauses of $F$,
if $F$ does not contain tautologies the difference disappears.

The only variables that may be consequences of a formula without being real
consequences of it are the variables of $B$. This is formally stated by the
following lemma~\cite{libe-20-b}.

\begin{lemma}
\label{b-rcn}

For every formula $F$ and set of variables $B$, it holds
{} $BCN(B,F) = A \cup RCN(B,F)$.

\end{lemma}

\section{The reconstruction algorithm}
\label{construction}

A formula may have multiple clauses with the same head and still be equivalent
to one that has not. Such a formula deceives the algorithm based on replacing
heads with their bodies by sending it on a wild goose chase across multiple
nondeterministic branches that eventually produce the same clause. Turning the
formula into its single-head form wherever possible avoids such a waste of
time.

The approximate algorithm shown in a previous article~\cite{libe-20-b} produces
a single-head formula that may be equivalent to $F$. It gives no guarantee, in
general: it may fail even if the formula is single-head equivalent.

The algorithm presented in this section does not suffer from this drawback.
When run on formula that has a single-head equivalent form, it finds it. The
trade-off is efficiency: it does not always finish in polynomial time.

\subsection{Bird's-eye view of the reconstruction algorithm}

The algorithm tries to reconstruct the formula clause by clause. It starts with
an empty formula and adds clauses to it until it becomes equivalent to the
input formula. Not adding clauses with the same head makes the resulting
formula single-head.

An example tells which clauses are added. The aim is to make the formula under
construction equivalent to the input formula. If the input formula contains $ab
\rightarrow x$, the formula under construction has to entail it to achieve
equivalence. This is the same as $x$ being implied by $\{a,b\}$.

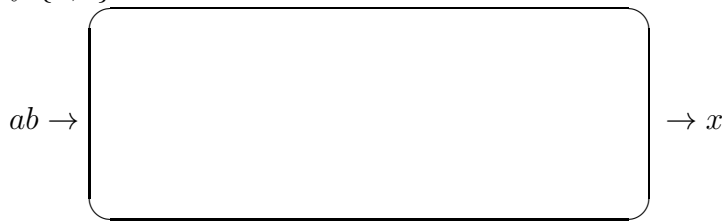
\setlength{\unitlength}{5000sp}%
\begingroup\makeatletter\ifx\SetFigFont\undefined%
\gdef\SetFigFont#1#2#3#4#5{%
  \reset@font\fontsize{#1}{#2pt}%
  \fontfamily{#3}\fontseries{#4}\fontshape{#5}%
  \selectfont}%
\fi\endgroup%
\begin{picture}(3255,1074)(3586,-4123)
\thinlines
{\color[rgb]{0,0,0}\put(3931,-4006){\oval(210,210)[bl]}
\put(3931,-3166){\oval(210,210)[tl]}
\put(6496,-4006){\oval(210,210)[br]}
\put(6496,-3166){\oval(210,210)[tr]}
\put(3931,-4111){\line( 1, 0){2565}}
\put(3931,-3061){\line( 1, 0){2565}}
\put(3826,-4006){\line( 0, 1){840}}
\put(6601,-4006){\line( 0, 1){840}}
}%
\put(3601,-3661){\makebox(0,0)[b]{\smash{{\SetFigFont{12}{24.0}
{\rmdefault}{\mddefault}{\updefault}{\color[rgb]{0,0,0}$ab \rightarrow $}%
}}}}
\put(6826,-3661){\makebox(0,0)[b]{\smash{{\SetFigFont{12}{24.0}
{\rmdefault}{\mddefault}{\updefault}{\color[rgb]{0,0,0}$ \rightarrow x$}%
}}}}
\end{picture}%
\nop{
      +---------------------------+
      |                           |
ab -> |                           | -> x
      |                           |
      +---------------------------+
}

What is in the rectangle? The implication of $x$ from $\{a,b\}$ is required in
the formula under construction. This formula aims at replicating the input
formula. It cannot contain clauses not entailed by that. The answer is that the
rectangle contains clauses entailed by the input formula.

Since all clauses are definite Horn, this implication is realized by forward
chaining: $a$ and $b$ derive some variables that derive some others that derive
others that derive $x$. For example, $a$ and $b$ derive $c$ and $d$, which
derive $e$ and $f$ which derive $y$ which derives $x$. Each individual
derivation is realized by a clause.

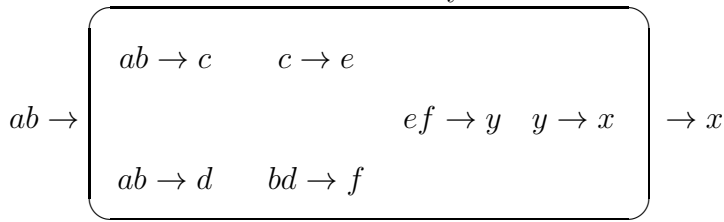
\setlength{\unitlength}{5000sp}%
\begingroup\makeatletter\ifx\SetFigFont\undefined%
\gdef\SetFigFont#1#2#3#4#5{%
  \reset@font\fontsize{#1}{#2pt}%
  \fontfamily{#3}\fontseries{#4}\fontshape{#5}%
  \selectfont}%
\fi\endgroup%
\begin{picture}(3255,1074)(3586,-4123)
\thinlines
{\color[rgb]{0,0,0}\put(3931,-4006){\oval(210,210)[bl]}
\put(3931,-3166){\oval(210,210)[tl]}
\put(6496,-4006){\oval(210,210)[br]}
\put(6496,-3166){\oval(210,210)[tr]}
\put(3931,-4111){\line( 1, 0){2565}}
\put(3931,-3061){\line( 1, 0){2565}}
\put(3826,-4006){\line( 0, 1){840}}
\put(6601,-4006){\line( 0, 1){840}}
}%
\put(3601,-3661){\makebox(0,0)[b]{\smash{{\SetFigFont{12}{24.0}
{\rmdefault}{\mddefault}{\updefault}{\color[rgb]{0,0,0}$ab \rightarrow $}%
}}}}
\put(6826,-3661){\makebox(0,0)[b]{\smash{{\SetFigFont{12}{24.0}
{\rmdefault}{\mddefault}{\updefault}{\color[rgb]{0,0,0}$ \rightarrow x$}%
}}}}
\put(4201,-3361){\makebox(0,0)[b]{\smash{{\SetFigFont{12}{24.0}
{\rmdefault}{\mddefault}{\updefault}{\color[rgb]{0,0,0}$ab \rightarrow c$}%
}}}}
\put(4201,-3961){\makebox(0,0)[b]{\smash{{\SetFigFont{12}{24.0}
{\rmdefault}{\mddefault}{\updefault}{\color[rgb]{0,0,0}$ab \rightarrow d$}%
}}}}
\put(4951,-3361){\makebox(0,0)[b]{\smash{{\SetFigFont{12}{24.0}
{\rmdefault}{\mddefault}{\updefault}{\color[rgb]{0,0,0}$c \rightarrow e$}%
}}}}
\put(5626,-3661){\makebox(0,0)[b]{\smash{{\SetFigFont{12}{24.0}
{\rmdefault}{\mddefault}{\updefault}{\color[rgb]{0,0,0}$ef \rightarrow y$}%
}}}}
\put(6226,-3661){\makebox(0,0)[b]{\smash{{\SetFigFont{12}{24.0}
{\rmdefault}{\mddefault}{\updefault}{\color[rgb]{0,0,0}$y \rightarrow x$}%
}}}}
\put(4951,-3961){\makebox(0,0)[b]{\smash{{\SetFigFont{12}{24.0}
{\rmdefault}{\mddefault}{\updefault}{\color[rgb]{0,0,0}$bd \rightarrow f$}%
}}}}
\end{picture}%
\nop{
      +---------------------------+
      | ab->c   c->e              |
ab -> |                ef->y y->x | -> x
      | ab->d   bd->f             |
      +---------------------------+
}

The first step of such a derivation has $a$ and $b$ imply another variable by
means of a clause like $ab \rightarrow c$. Without such a clause, forward
chaining does not even start. Nothing derives from $\{a,b\}$ without. Not even
a single variable to derive others.

Some clauses in the rectangle are like $ab \rightarrow c$ and $ab \rightarrow
d$, with $\{a,b\}$ as their body. Maybe not two, but at least one. Not zero. At
least a clause in the rectangle has body $\{a,b\}$. Maybe a subset like
$\{a\}$, but this case will be discussed later. For the moment, some clauses on
the left edge of the rectangle have $\{a,b\}$ as their bodies.

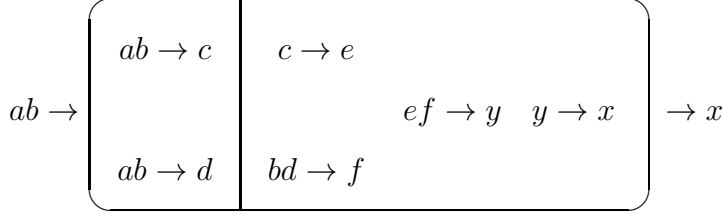
\setlength{\unitlength}{5000sp}%
\begingroup\makeatletter\ifx\SetFigFont\undefined%
\gdef\SetFigFont#1#2#3#4#5{%
  \reset@font\fontsize{#1}{#2pt}%
  \fontfamily{#3}\fontseries{#4}\fontshape{#5}%
  \selectfont}%
\fi\endgroup%
\begin{picture}(3255,1074)(3586,-4123)
\thinlines
{\color[rgb]{0,0,0}\put(3931,-4006){\oval(210,210)[bl]}
\put(3931,-3166){\oval(210,210)[tl]}
\put(6496,-4006){\oval(210,210)[br]}
\put(6496,-3166){\oval(210,210)[tr]}
\put(3931,-4111){\line( 1, 0){2565}}
\put(3931,-3061){\line( 1, 0){2565}}
\put(3826,-4006){\line( 0, 1){840}}
\put(6601,-4006){\line( 0, 1){840}}
}%
\put(3601,-3661){\makebox(0,0)[b]{\smash{{\SetFigFont{12}{24.0}
{\rmdefault}{\mddefault}{\updefault}{\color[rgb]{0,0,0}$ab \rightarrow $}%
}}}}
\put(6826,-3661){\makebox(0,0)[b]{\smash{{\SetFigFont{12}{24.0}
{\rmdefault}{\mddefault}{\updefault}{\color[rgb]{0,0,0}$ \rightarrow x$}%
}}}}
\put(4201,-3361){\makebox(0,0)[b]{\smash{{\SetFigFont{12}{24.0}
{\rmdefault}{\mddefault}{\updefault}{\color[rgb]{0,0,0}$ab \rightarrow c$}%
}}}}
\put(4201,-3961){\makebox(0,0)[b]{\smash{{\SetFigFont{12}{24.0}
{\rmdefault}{\mddefault}{\updefault}{\color[rgb]{0,0,0}$ab \rightarrow d$}%
}}}}
\put(4951,-3361){\makebox(0,0)[b]{\smash{{\SetFigFont{12}{24.0}
{\rmdefault}{\mddefault}{\updefault}{\color[rgb]{0,0,0}$c \rightarrow e$}%
}}}}
\put(5626,-3661){\makebox(0,0)[b]{\smash{{\SetFigFont{12}{24.0}
{\rmdefault}{\mddefault}{\updefault}{\color[rgb]{0,0,0}$ef \rightarrow y$}%
}}}}
\put(6226,-3661){\makebox(0,0)[b]{\smash{{\SetFigFont{12}{24.0}
{\rmdefault}{\mddefault}{\updefault}{\color[rgb]{0,0,0}$y \rightarrow x$}%
}}}}
\put(4951,-3961){\makebox(0,0)[b]{\smash{{\SetFigFont{12}{24.0}
{\rmdefault}{\mddefault}{\updefault}{\color[rgb]{0,0,0}$bd \rightarrow f$}%
}}}}
{\color[rgb]{0,0,0}\put(4576,-3061){\line( 0,-1){1050}}
}%
\end{picture}%
\nop{
      +-------+-------------------+
      | ab->c | c->e              |
ab -> |       |        ef->y y->x | -> x
      | ab->d | bd->f             |
      +-------+-------------------+
}

A first stage contains $ab \rightarrow c$ and $ab \rightarrow d$; a second
contains the following clauses used in the derivation of $x$. The key is
induction. The formula under construction is assumed to already imply the
clauses in the second stage. Adding $ab \rightarrow c$ and $ab \rightarrow d$
completes the derivation.

The inductive assumption is that the formula under construction implies all
clauses in the second stage. The addition of $ab \rightarrow c$ and $ab
\rightarrow d$ makes it imply $ab \rightarrow x$ as well. Such an induction is
only possible if these two clauses are added only when the others are already
implied. Adding them makes $ab \rightarrow x$ entailed. This entailment is the
induction claim. Which is the induction assumption for the next step.

The induction step that makes $ab \rightarrow x$ entailed requires $c
\rightarrow e$ to be entailed, which is only possible if the induction step
that makes $c \rightarrow e$ entailed is in the past. The same for the other
clauses in the second stage: when the clauses of body $ab$ are added, all
clauses of body $c$, $bd$, etc. are already. Key to induction is the order of
addition of clauses.

How the single-head property is achieved is still untalked of. In the example,
once $c \rightarrow e$ is in, all other sets of variables entailing $e$ should
do so by first entailing $c$. For example, if $ab \rightarrow e$ is in the
input formula, it is entailed by the formula under construction thanks to $ab
\rightarrow c$ and $c \rightarrow e$; no need for duplicated heads. If $ab$
entailed $e$ but not $c$, the input formula is not single-head equivalent.

\

All of this works flawlessly because of the simplicity of the input formula.
Not in general.

Induction works because the algorithm tries to entail $ab \rightarrow x$ only
when all clauses whose bodies are entailed by $ab$ are already entailed. Such
an assumption requires a caveat: these bodies are entailed by $ab$ but do not
entail it. It does not work on bodies that are not just entailed by $ab$ but
also equivalent to it. Extending the induction assumption to them makes it
circular rather than inductive. On the example: if $gh$ is equivalent to $ab$,
the induction step for $ab$ follows that of $gh$; but the step for $gh$ follows
that of $ab$ for the same reason.

Equivalent sets cannot be worked on inductively. Even worse, they cannot be
worked on separately as well. When the step that adds $ab \rightarrow x$ is
over, the following steps rely on its inductive claim. A following step for the
body $ij$ may require $ab \rightarrow x$ to be entailed, if $ij$ entails $ab$.
Because of equivalence, $ij$ also entails $gh$. The induction assumption also
requires the clauses of body $gh$ to be entailed. The clauses of head $ab$ and
$gh$ to be found together.

This also tells what would happen if the first stage of derivation contained $a
\rightarrow c$ instead of $ab \rightarrow c$, a strict subset of $ab$ instead
of $ab$ itself. By monotonicity, $a$ is entailed by $ab$; if it does not entail
it, $a \rightarrow c$ is already in the formula under construction thanks to
the inductive assumption. Otherwise, $ab$ and $a$ are equivalent. The induction
step where $ab \rightarrow x$ is achieved also achieves $a \rightarrow c$.

Equivalent bodies also require special care to ensure the single-head property.
Searching for the clauses of body $ab$ and $gh$ at the same time may result in
two clauses such as $ab \rightarrow c$ and $gh \rightarrow c$. Such a solution
may as well satisfy the inductive claim, but is not acceptable. As a previous
article illustrates for pages and pages~\cite{libe-20-b}, just adding $gh
\rightarrow c$ and hoping for equivalence to give $ab \rightarrow c$ does not
work; this very clause $ab \rightarrow c$ may be necessary to ensure
equivalence.

The algorithm in this section turns to a mysterious function to produce the
necessary clauses of body $ab$---or equivalent. The next section unveils the
mystery.

\subsection{Worm's eye view of the reconstruction algorithm}

The key to induction is the entailment between sets of variables. The required
conditions are long to write: ``a set variables entails another without being
entailed by it''. Sentences involving one of these segments may still be
readable, but two or three make them unparsable. An order between sets used in
a previous article~\cite{libe-20-b} shortens wording: $A \leq_F B$ if $F
\models B \rightarrow A$. If $A$ entails $B$ according to $F$ but is not
entailed by it, then $A <_F B$. If they entail each other, then $A \equiv_F B$.

The algorithm sorts the bodies of $F$ according to this order, neglecting their
heads. The minimal bodies are considered first. The set of clauses with these
bodies are turned into single-head form. With this set fixed, the bodies that
are minimal among the remaining ones are processed in the same way. The
procedure ends when no body is left.

If $A$ is the only minimal body, the first step finds all clauses $A
\rightarrow x$ entailed by $A$. If the second-minimal body is $B$, the second
step finds some clauses $B \rightarrow y$; not all of them, since some may be
entailed thanks to the clauses produced in the first step. The same is done in
the remaining steps.

No backtracking is required. The clauses generated by a step are in the final
formula regardless of what happens in the following steps. This is not the
reason why the algorithm may take exponential time.

The reason is that each step may take exponential time. When $A \equiv_F B$,
the heads for $A$ and $B$ have to be generated in the same step. Sometimes this
is easy, sometimes it is not. In the worst case, the possibilities increase
exponentially with the number of equivalent sets.

Technically, the algorithm reconstructs the formula starting from the clauses
of minimal bodies:

\begin{itemize}

\item the formula is initially empty;

\item a body $A$ that only entails equivalent bodies is chosen;

\item enough clauses $P \rightarrow x$ with $P \equiv_F A$ are selected so that
the reconstructed formula entails all clauses of the given formula with a body
equivalent to $A$;

\item a body $B$ that only entails bodies entailed by $A$ or equivalent to $B$
is chosen;

\item enough clauses $P \rightarrow x$ with $P \equiv_F B$ are selected so that
the reconstructed formula entails all clauses of the given formula with a body
equivalent to $B$;

\item a body $C$ that only entails bodies entailed by $A$ and $B$ or equivalent
to $C$ is chosen;

\item \ldots

\end{itemize}

Since bodies are considered in increasing order, when a body $B$ is considered
all bodies $A <_F B$ have already been. This implies that all clauses $P
\rightarrow x$ with $P \equiv_F A$ are already entailed by the formula under
construction.

\subsection{The construction lemma}

The reconstruction algorithm looks like a typical backtracking mechanism: some
clauses $P \rightarrow x$ with $P \equiv_F A$ are chosen; the same for $B$,
$C$, and so on; choosing clauses for a following set $D$ may prove impossible,
showing that the choice for $A$ was wrong. Backtracking to $A$ is necessary.

It is not.

The reason is that each step generates enough clauses to entail all clauses
with a body equivalent to the given one. Different choices may be possible, but
the result always entails all these clauses. To entail all these clauses, their
heads are required to be in the chosen clauses. Different choices differ in
their bodies, but not in their heads and consequences. If a following choice
proves impossible, backtracking is of no help since it can only replace a set
of clauses with an equivalent one, with the same heads. The same impossible
choice is faced again.

\begin{lemma}
\label{reconstruction}

If $F$ is a single-head formula such that $F \models A \rightarrow x$ with $x
\not\in A$ and $F \not\models A \rightarrow B$, then $F$ does not contain $B
\rightarrow x$.

\end{lemma}

\proof Since $F \models A \rightarrow x$ and $x \not\in A$,
Lemma~\ref{set-implies-set} implies that $F$ contains a clause $P \rightarrow
x$ such that $F \models A \rightarrow P$. The set $P$ differs from $B$ since
$F$ entails $A \rightarrow P$ but not $A \rightarrow B$. As a result, $B
\rightarrow x$ is not the same as $P \rightarrow x$. Since $F$ contains the
latter clause and is single-head, it does not contain the former.~\qed

Why does this lemma negate the need for backtracking? Backtracking kicks in
when adding $B \rightarrow x$ is necessary but another clause $A \rightarrow x$
is already in the formula under construction. Because of the order of the
steps, $B$ is either strictly larger than $A$ or incomparable to it. The
condition $F \not\models A \rightarrow B$ holds in both. The lemma proves that
$B \rightarrow x$ is not in any single-head formula equivalent to $F$. Adding
this clause to the formula under construction is not only useless, it is
impossible if that formula has to be single-head. Which clauses the step for
$A$ added is irrelevant.

\subsection{One iteration of the reconstruction algorithm}

An iteration of the reconstruction algorithm picks a body $B$ only when all
bodies $A <_F B$ have already been processed by a previous iteration. As a
result, all clauses $A' \rightarrow x$ with $A' \equiv_F A$ are already
entailed by the formula under construction. This is the fundamental invariant
of the algorithm.

When looking at the iteration for a body $B$, the bodies $A$ such that $A <_F
B$ are called {\em strictly entailed bodies} to simplify wording. Since $B$ is
fixed, ``all strictly entailed bodies do something'' is simpler than ``all
bodies $A$ such that $A <_F B$ do something''.

When processing $B$, the formula under construction is the result of the
previous iterations. The set $A$ of a previous iteration cannot be greater or
equal than $B$; therefore, either $A$ is incomparable with $B$ or is less than
it. If $A$ is incomparable with $B$, then $F \not\models B \rightarrow A$; the
clauses of body $A$ or equivalent are not relevant to implications from $B$. If
$A <_F B$ then $F \models B \rightarrow A$; the clauses of body $A$ or
equivalent to $A$ help entailing variables from $B$. The construction lemma
ensures that their heads $x$ are not valid candidates as heads for $B$. The
following definitions express these ideas.

\begin{definition}

Given a formula $F$, the {\em strictly entailed bodies} of $B$ are the sets $A$
such that $A \rightarrow x \in F$ for some $x$ and $A <_F B$. The clauses and
free heads of such bodies are:

\begin{description}

\item[clauses of strictly entailed bodies:]

\[
SCL(B,F) = \{ A \rightarrow x \mid
	x \not\in A ,~
	F \models A \rightarrow x ,~
	F \models B \rightarrow A ,~
	F \not\models A \rightarrow B
\}
\]

\item[heads free from strictly entailed bodies:]

\[
SFREE(B,F) = \{ x \mid \not\exists A \rightarrow x \in SCL(B,F) \}
\]

\end{description}

\end{definition}

The iteration for a body $B$ produces some clauses $B' \rightarrow x$ with $B'
\equiv B$. The construction lemma limits the choices of the heads to $x \in
SFREE(B,F)$. All clauses of $SCL(B,F)$ are already entailed by the formula
under construction. They are the only ones relevant to finding new clauses $B'
\rightarrow x$. The goal of the iteration is to entail the following clauses.

\begin{definition}

Given a set of variables $B$ and a formula $F$,
the {\em clauses of equivalent bodies} are:

\[
BCL(B,F) = \{ A \rightarrow x \mid
	x \not\in A,
	F \models A \rightarrow x ,~
	F \models B \rightarrow A ,~
	F \models A \rightarrow B
\}
\]

\end{definition}

The algorithm progressively builds a formula equivalent to the given one. It is
initially empty. Every step hinges around the body $B$ of a clause. The main
invariant is that the formula under construction already entails $SCL(B,F)$;
the aim is to make it entail $BCL(B,F)$ as well. This makes the invariant true
again in the following steps, since $BCL(B,F)$ is part of $SCL(C,F)$ if $B <_F
C$. The $SFREE(B,F)$ set simplifies this task by restricting the possible heads
of the new clauses.

Every step adds some new clauses to the formula; which ones they are is
explained in a later section. This one is about the overall algorithm. To it,
what matters of a single step is only what it produces, not how. What it
produces is formalized as follows.

\begin{definition}
\label{condition-iteration}

A {\em valid iteration} is a function $ITERATION(B,F)$ that satisfies the
following two conditions for every set of variables $B$ and formula $F$:

\begin{description}

\item[choice:]
$ITERATION(B,F)$ is a set of non-tautological clauses $B' \rightarrow x$
such that

\begin{itemize}

\item $F \models B' \rightarrow x$,
\item $F \models B \equiv B'$ and
\item $x \in SFREE(B,F)$;

\end{itemize}

% and there exists $B''$ such that $x \in B''$ and $F \models B'' \equiv B$?
% but this is implied by $F \models B' \rightarrow x$;
% still allows regular implications like $B \rightarrow x$,
% since $B \equiv B \cup \{x\}$

\item[entailment:]
$SCL(B,F) \cup ITERATION(B,F) \models BCL(B,F)$.

\end{description}

\end{definition}

The first condition (choice) limits the possible clauses $B' \rightarrow x$
generated at each step. The second (entailment) requires them to satisfy the
invariant for the next steps (entailment of $BCL(B,F)$) given the invariant for
the current (entailment of $SCL(B,F)$).

A minimal requirement for the steps of the algorithm is that they succeed when
the formula is already single-head. In spite of its weakness, this property is
useful for proving that the algorithm works on single-head equivalent formulae.

\begin{lemma}
\label{iteration-from-singlehead}

If $F$ is a single-head formula,
{} $ITERATION(B,F) =
{} \{ B' \rightarrow x \in F \mid x \not\in B' ,~ F \models B \equiv B' \}$
is a valid iteration function.

\end{lemma}

\proof The given function $ITERATION(B,F)$ is shown to satisfy both the choice
and the entailment condition, the two parts of the definition of a valid
iteration function (Definition~\ref{condition-iteration}).

The choice condition concerns the individual clauses of $ITERATION(B,F)$. They
are not tautological by construction. The other requirements are proved one by
one for each $B' \rightarrow x \in ITERATION(B,F)$.

\begin{itemize}

\item The first is $F \models B' \rightarrow x$.
It holds because $B' \rightarrow x \in F$
is part of the definition of this function $ITERATION(B,F)$.

\item The second is $F \models B \equiv B'$.
It is part of the definition of this function $ITERATION(B,F)$.

\item The third is $x \in SFREE(B,F)$. Its converse $x \not\in SFREE(B,F)$
means that $SCL(B,F)$ contains a clause $A \rightarrow x$. The definition of $A
\rightarrow x \in SCL(B,F)$ is $x \not\in A$, $F \models A \rightarrow x$, $F
\models B \rightarrow A$ and $F \not\models A \rightarrow B$. The latter
implies $F \not\models A \rightarrow B'$ since $F \models B \equiv B'$.
Lemma~\ref{reconstruction} proves that $F \models A \rightarrow x$, $x \not\in
A$ and $F \not\models A \rightarrow B'$ imply $B' \rightarrow x \not\in F$,
which contradicts $B' \rightarrow x \in F$.

\end{itemize}

The given function $ITERATION(B,F)$ also satisfies the entailment condition of
Definition~\ref{condition-iteration}:
{} $SCL(B,F) \cup ITERATION(B,F) \models BCL(B,F)$.
This is implied by
{} $SCL(B,F) \cup ITERATION(B,F) \models SCL(B,F) \cup BCL(B,F)$,
which is defined by
{} $SCL(B,F) \cup ITERATION(B,F) \models B' \rightarrow x$
{} if $x \not\in B'$,
{}    $F \models B' \rightarrow x$,
{}    $F \models B \rightarrow B'$ and
{}    either $F \not\models B' \rightarrow B$ or $F \models B' \rightarrow B$.
Either one choice of alternative is always true, which simplifies the condition
to
{} $SCL(B,F) \cup ITERATION(B,F) \models B' \rightarrow x$
{} if $x \not\in B'$,
{}    $F \models B' \rightarrow x$ and
{}    $F \models B \rightarrow B'$.
What is actually proved is a more general condition:
{} for every $G \subseteq F$,
{} $SCL(B,F) \cup ITERATION(B,F) \models B' \rightarrow x$ holds
{} if $x \not\in B'$,
{}    $G \models B' \rightarrow x$ and
{}    $F \models B \rightarrow B'$.
The introduction of $G$ allows for induction over the size of $G$.

If $G$ is empty, it entails $B' \rightarrow x$ only if $x \in B'$. Such a
clause does not satisfy the other requirement $x \not\in B'$. The condition is
met. This is the base case of induction.

The induction step requires proving the claim for an arbitrary $G \subseteq F$
assuming it for every formula smaller than $G$.

The claim is that
{}    $G \subseteq F$, 
{}    $x \not\in B'$,
{}    $G \models B' \rightarrow x$ and
{}    $F \models B \rightarrow B'$
imply
{} $SCL(B,F) \cup ITERATION(B,F) \models B' \rightarrow x$.
The induction assumption is the same for every $G' \subset G$.

The premises $x \not\in B'$ and $G \models B' \rightarrow x$ allow applying
Lemma~\ref{set-implies-set}: $G^x \models B' \rightarrow B''$ holds for some
clause $B'' \rightarrow x \in G$.

A consequence of $B'' \rightarrow x \in G$ is $B'' \rightarrow x \in F$ since
$G \subseteq F$. For the same reason, $G^x \models B' \rightarrow B''$ implies
$F \models B' \rightarrow B''$. With the other assumption $F \models B
\rightarrow B'$, transitivity implies $F \models B \rightarrow B''$. The
converse implication $B'' \rightarrow B$ may be entailed by $F$ or not. If $F
\models B'' \rightarrow B$ holds then either $x$ is in $B''$ or $B''
\rightarrow x$ is in $ITERATION(B,F)$; in both cases, $B'' \rightarrow x$ is
entailed by
{} $SCL(B,F) \cup ITERATION(B,F)$. If $F \not\models B'' \rightarrow B$ then
either $x$ is in $B''$ or $B'' \rightarrow x$ is in $SCL(B,F)$; in both cases,
$B'' \rightarrow x$ is entailed by
{} $SCL(B,F) \cup ITERATION(B,F)$.
The conclusion of this paragraph is
{} $SCL(B,F) \cup ITERATION(B,F) \models B'' \rightarrow x$.

% note: membership of $B'' \rightarrow x$ to $ITERATION(B,F)$ requires $B''
% \rightarrow x$ to belong to $F$, while memership to $SCL(B,F)$ requires only
% to be entailed

Induction proves that $SCL(B,F) \cup ITERATION(B,F)$ also entails $B'
\rightarrow B''$. This condition is the same as
{} $SCL(B,F) \cup ITERATION(B,F) \models B' \rightarrow b$
for every $b \in B''$. Since $G^x$ entails $B' \rightarrow B''$, it entails $B'
\rightarrow b$ for every $b \in B''$. If $b \in B'$ then $B' \rightarrow b$ is
a tautology and is therefore entailed by
{} $SCL(B,F) \cup ITERATION(B,F)$.
Otherwise, $b \not\in B'$. Since $B'' \rightarrow x$ contains $x$, it is not in
$G^x$; since $G$ contains $B'' \rightarrow x$ instead, $G^x$ is a proper subset
of $G$. The induction assumption is the same as the induction claim with $G^x$
in place of $G$, that is,
{}    $G^x \subseteq F$, 
{}    $x \not\in B'$,
{}    $G^x \models B' \rightarrow b$ and
{}    $F \models B \rightarrow B'$
imply
{} $SCL(B,F) \cup ITERATION(B,F) \models B' \rightarrow b$.
Since all premises hold, the claim holds as well:
{} $SCL(B,F) \cup ITERATION(B,F) \models B' \rightarrow b$.
This is the case for every $b \in B''$. A consequence is
{} $SCL(B,F) \cup ITERATION(B,F) \models B' \rightarrow B''$.

The conclusion
{} $SCL(B,F) \cup ITERATION(B,F) \models B' \rightarrow B''$
joins
{} $SCL(B,F) \cup ITERATION(B,F) \models B'' \rightarrow x$
to prove
{} $SCL(B,F) \cup ITERATION(B,F) \models B' \rightarrow x$.
This is the induction claim.~\qed

The specific function $ITERATION(B,F)$ of this lemma gives only clauses of $F$.
It depends on the syntax of $F$. Yet, the validity of an iteration function is
semantical: only depends on entailments, equivalences and the three functions
$SCL(B,F)$, $SFREE(B,F)$ and $BCL(B,F)$, which are all defined in terms of
entailments. Therefore, if $ITERATION(B,F)$ is valid, the same function applied
to another formula $ITERATION(B,F')$ is still valid if $F \equiv F'$.

\begin{lemma}
\label{iteration-equivalent}

If $ITERATION(B,F)$ is a valid iteration function and $F \equiv F'$, then
$ITERATION(B,F') = ITERATION(B,F)$ is a valid iteration function.

\end{lemma}

\proof Since $ITERATION(B,F)$ is valid iteration function, it produces a set of
clauses $B' \rightarrow x$ such that $x \not\in B'$, $F \models B \rightarrow
x$, $F \models B \equiv B'$, $x \in SFREE(B,F)$ and all of them plus $SCL(B,F)$
entail all clauses $B' \rightarrow x$ with $F \models B \equiv B'$ that are
entailed by $F$. In turn, $SCL(B,F)$ is defined as the set of clauses $B''
\rightarrow x$ such that $F \models B'' \rightarrow x$, $F \models B
\rightarrow B''$ and $F \not\models B'' \rightarrow x$. Finally, $SFREE(B,F)$
is defined as the set of variables $x$ such that no clause $B'' \rightarrow x
\in SCL(B,F)$.

All these conditions are based on what $F$ entails. Since $F$ is equivalent to
$F'$, they also hold when $F'$ is in place of $F$. Therefore, $ITERATION(B,F')
= ITERATION(B,F)$ is also a valid iteration function.~\qed

The two lemmas together prove that if $F$ is single-head equivalent then it has
a valid iteration function: the one for a single-head formula that is
equivalent to $F$.

\begin{lemma}
\label{equivalent-valid}

If $F$ is equivalent to the single-head formula $F'$, then
{} $ITERATION(B,F) = \{ B' \rightarrow x \in F' \mid F' \models B \equiv B' \}$
is a valid iteration function.

\end{lemma}

\proof By Lemma~\ref{iteration-from-singlehead}, $ITERATION(B,F')$ is a valid
iteration function for $F'$. By Lemma~\ref{iteration-equivalent}, it is also
valid for $F$.~\qed

As a result, every single-head equivalent formula has a valid iteration
function.

\begin{corollary}
\label{all-singlehead-valid}

Every single-head equivalent formula has a valid iteration function.

\end{corollary}

The proofs of the three lemmas show how to find a valid iteration function:
switch from the single-head equivalent formula to a single-head formula that is
equivalent to it and then extract the necessary clauses from it.

This mechanism works in theory, and is indeed used in proofs that requires a
valid iteration function to exists. It does not work in practice, since the
single-head formula equivalent to the given one is not known. Rather the
opposite: the whole point of valid iteration functions is to be used in
reconstruction algorithm to find a single-head formula equivalent to the given
one. The iteration function is a means to the aim of finding a single-head
formula, not the other way around.

In practice, finding a valid iteration function is no easy task. An entire
following section is devoted to that. For the moment, a valid iteration
function is assumed available.

\subsection{The reconstruction algorithm}

Assuming the availability of a valid iteration function $ITERATION(B,F)$, the
reconstruction algorithm is as follows.

\

\begin{algorithm}[Reconstruction algorithm]
\label{algorithm-reconstruction}

\begin{enumerate}

\item $G = \emptyset$

\item $P = P(X)$ \# set of all subsets of $X$

\item while $P \not= \emptyset$:

\begin{enumerate}

\item choose a $\leq_F$-minimal $B \in P$
\label{algorithm-reconstruction-choice}
\newline
(a set such that $A <_F B$ does not hold for any $A \in P$)

\item $G = G \cup ITERATION(B,F)$

\item $P = P \backslash \{B' \mid F \models B \equiv B'\}$

\end{enumerate}

\item return $G$

\end{enumerate}

\end{algorithm}

\

The algorithm is non-deterministic, but works for every possible
nondeterministic choices. It is not deterministic because many sets $B$ may be
minimal each time. The final result depends on these choices because
$ITERATION(B,F)$ may differ from $ITERATION(B',F)$ even if $B \equiv_F B'$.
Yet, an equivalent single-head formula is found anyway, if any.

As an example, a valid iteration function for
{} $F = \{a \rightarrow b, b \rightarrow a, bc \rightarrow d\}$
has the values
{} $ITERATION(\{a,c\},F) = \{bc \rightarrow d\}$
and
{} $ITERATION(\{b,c\},F) = \{ac \rightarrow d\}$.
The reconstruction algorithm produces either $bc \rightarrow d$ or $ac
\rightarrow d$ depending on whether it chooses $\{a,c\}$ or $\{b,c\}$ as the set
$B$ of an iteration of its main loop. Either way, the result is single-head and
equivalent to $F$.

The following lemmas state some properties of the algorithm that are
independent on the nondeterministic choices. For example, proving termination
means that the algorithm terminates for all possible nondeterministic choices.

\begin{lemma}
\label{reconstruction-terminates}

The reconstruction algorithm (Algorithm~\ref{algorithm-reconstruction}) always
terminates.

\end{lemma}

\proof The set $P$ becomes smaller at each iteration, since a choice of $B$ is
always possible if $P$ is not empty because $\leq_F$ is a partial order. As a
result, at some point $P$ becomes empty and the algorithm terminates.~\qed

The second property of the reconstruction algorithm is that it never calculates
$ITERATION(B,F)$ for equivalent sets $B$. Still better, it calculates
$ITERATION(B,F)$ exactly once for each $B$ among its equivalent ones. This is
proved by the following two lemmas.

\begin{lemma}
\label{reconstruction-one-equivalent}

For every set of variables $B'$, the reconstruction algorithm
(Algorithm~\ref{algorithm-reconstruction}) calculates $ITERATION(B,F)$ for some
$B$ such that $F \models B \equiv B'$.

\end{lemma}

\proof Initially, $B' \in P$ since $P$ contains all sets of variables. The
algorithm terminates by Lemma~\ref{reconstruction-terminates}. It only
terminates when $P$ is empty. Therefore, $B'$ is removed from $P$ during the
run of the algorithm. Removal is only done by $P = P \backslash \{B' \mid F
\models B \equiv B'\}$. This operation is preceded by $G = G \cup
ITERATION(B,F)$.~\qed

\begin{lemma}
\label{reconstruction-no-equivalent}

The reconstruction algorithm (Algorithm~\ref{algorithm-reconstruction}) does
not calculate both $ITERATION(B,F)$ and $ITERATION(B',F)$ if $F \models B
\equiv B'$.

\end{lemma}

\proof Let $B$ and $B'$ be two sets such that the reconstruction algorithm
calculates $ITERATION(B,F)$ and $ITERATION(B',F)$. This implies that the
algorithm chooses $B$ in an iteration and $B'$ in another. By symmetry, the
iteration of $B$ can be assumed to come first. This iteration ends with $P = P
\backslash \{B' \mid F \models B \equiv B'\}$. When the iteration of $B'$
begins, no set equivalent to $B$ according to $F$ is in $P$ any longer. Since
$B'$ is in $P$, it is not equivalent to $B$. This is a contradiction.~\qed

When the reconstruction algorithm extracts a set $B$ from $P$, all sets $A$
with $A <_F B$ have been removed in a past iteration. During that, $G$ was
increased to entail all clauses $A \rightarrow x$ entailed by $F$, which make
$SCL(B,F)$ because of $A <_F B$. That $G$ entails $SCL(B,F)$ is the fundamental
invariant of the algorithm.

\begin{lemma}
\label{reconstruction-invariant}

If $ITERATION(B,F)$ is a valid iteration function, $G \models SCL(B,F)$ holds
during the whole execution of the reconstruction algorithm
(Algorithm~\ref{algorithm-reconstruction}).

\end{lemma}

\proof The proof links the values of the variables at different iterations. The
values of $P$, $B$ and $G$ at the $i$-th iteration of the algorithm are denoted
$P_i$, $B_i$ and $G_i$. More precisely, these are the values right after
Step~\ref{algorithm-reconstruction-choice} of the $i$-th iteration.

The claim $G \models SCL(B,F)$ is proved by induction over the number of
iterations of the loop.

In the first iteration, $P_1$ contains all sets of variables, $G_1$ is the
empty set and $B_1$ is a set of $P_1$ such that $P_1$ contains no set $A <_F
B_1$. The values of $P_1$ and $B_1$ imply that no set such that $A <_F B_1$
exists. As a result, $SCL(B_1,F)$ is empty and therefore entailed even by the
empty formula $G_1$.

The induction claim is $G_i \models SCL(B_i,F)$; the induction assumption is
$G_j \models SCL(B_j,F)$ for every $j < i$.

The claim $G_i \models SCL(B_i,F)$ is $G_i \models A \rightarrow x$ if $x
\not\in A$, $F \models A \rightarrow x$ and $A <_F B$. The set $B_i$ is an
element of $P_i$ such that $P_i$ contains no element that is strictly less than
$B_i$. Since $A$ is strictly less than $B_i$, it is not in $P_i$. Going back to
the first iteration, $P_1$ contains all sets of variables, including $A$.
Therefore, $A$ is removed at some iteration $j < i$. The only removal
instruction is
{} $P_{j+1} = P_j \backslash \{B' \mid F \models B_j \equiv B'\}$.
It removes $A$; therefore, $F \models B_j \equiv A$ holds. With the assumptions
that $A \rightarrow x$ is not tautologic and is entailed by $F$, it proves that
$A \rightarrow x$ belongs to $BCL(B_j,F)$.

The induction assumption is $G_j \models SCL(B_j,F)$. A consequence is that
$G_j \cup ITERATION(B_j,F)$ entails $SCL(B_j,F) \cup ITERATION(B_j,F)$. By
definition of a valid iteration function, the latter entails $BCL(B_j,F)$,
which includes $A \rightarrow x$. This clause is therefore entailed by $G_j
\cup ITERATION(B_j,F)$, the same as $G_{j+1}$.

Since $j < i$ and $G$ monotonically increases, $G_{j+1} \subseteq G_i$. As a
result, $G_i \models A \rightarrow x$, which is the claim.~\qed

% Every iteration of the algorithm makes $G$ entail $BCL(B,F)$. Since an
% iteration is run for every set $B$, at the end $G$ entails all of $F$. Since
% $G$ is only added clauses entailed by $F$, it is eventually equivalent to it.

The reconstruction algorithm generates a formula that is equivalent to $F$.
This holds for every valid iteration function and nondeterministic choices.

\begin{lemma}
\label{reconstruction-equivalent}

If $ITERATION(B,F)$ is a valid iteration function, the formula produced by the
reconstruction algorithm is equivalent to $F$.

\end{lemma}

\proof The algorithm returns $G$, which is only added clauses of
$ITERATION(B,F)$. All of them are entailed by $F$ by the definition of valid
iteration functions. This proves that the formula returned by the algorithm
contains only clauses entailed by $F$.

It is now proved to imply all clauses $B' \rightarrow x$ entailed by $F$. By
Lemma~\ref{reconstruction-one-equivalent}, the reconstruction algorithm
calculates $ITERATION(B,F)$ for some $B$ such that $F \models B \equiv B'$. A
consequence of the invariant $G \models SCL(B,F)$ proved by
Lemma~\ref{reconstruction-invariant} is that $G \cup ITERATION(B,F)$ entails
$SCL(B,F) \cup ITERATION(B,F)$, which entails $B' \rightarrow x$ by the
definition of valid iteration. By transitivity, $G \cup ITERATION(B,F)$ entails
$B' \rightarrow x$. After $G = G \cup ITERATION(B,F)$, it is $G$ that entails
$B' \rightarrow x$. Since $G$ is never removed clauses, at the end of the
algorithm $G$ still entails $B' \rightarrow x$.~\qed

The algorithm is correct: it always terminates and always outputs a formula
equivalent to the given one. What is missing is its completeness: if the input
formula is equivalent to a single-head formula, that formula is output for some
valid iteration function.

\begin{lemma}
\label{all-singlehead-equivalent}

If $F$ is equivalent to a single-head formula $F'$, for some valid iteration
function the reconstruction algorithm
(Algorithm~\ref{algorithm-reconstruction}) returns $F'$.

\end{lemma}

\proof By Lemma~\ref{equivalent-valid},
{} $ITERATION(B,F) =
{}	\{ B' \rightarrow x \in F' \mid F' \models B \equiv B' \}$
is a valid iteration. The algorithm always terminates thanks to
Lemma~\ref{reconstruction-equivalent}. It returns only clauses from
$ITERATION(B,F)$; since these are all clauses of $F'$, the return value is a
subset of $F'$.

The claim is proved by showing that all clauses of $F'$ are returned. Let $B'
\rightarrow x \in F'$. By Lemma~\ref{reconstruction-one-equivalent}, the
algorithm calculates $ITERATION(B,F)$ for some $B$ such that $F \models B
\equiv B'$. By construction, $B' \rightarrow x$ is in $ITERATION(B,F)$. The
algorithm returns the union of all sets $ITERATION(B,F)$ it calculates, and $B'
\rightarrow x$ is in this union.~\qed

This lemma tells that every single-head formula equivalent to the input can be
returned by the reconstruction algorithm using the appropriate iteration
function. Yet, an inappropriate iteration function may make it return a formula
that is still equivalent to the original but is not single-head. An additional
property of the iteration function is required to avoid such an outcome.

\begin{lemma}
\label{singlehead-local}

If $ITERATION(B,F)$ is a valid iteration function whose values never contain
each two clauses with the same head then the reconstruction algorithm
(Algorithm~\ref{algorithm-reconstruction}) returns a single-head formula if and
only if $F$ is equivalent to a single-head formula.

\end{lemma}

\proof Let $F'$ be the formula returned by the reconstruction algorithm on $F$.
By Lemma~\ref{reconstruction-equivalent}, $F \equiv F'$.

If $F'$ is single-head, then $F$ is single-head equivalent by definition.

Otherwise, $F'$ is not single-head. The following proof shows that $F$ is
equivalent to no single-head formula.

Let $B \rightarrow x$ and $C \rightarrow x$ be two clauses returned by the
algorithm. They were added to $G$ in different iterations, since by assumption
$ITERATION(B,F)$ does not contain two clauses with the same head. Let $B'$ and
$C'$ be the values of $B$ in the iterations that produce them:

\begin{eqnarray*}
B \rightarrow x & \in & ITERATION(B',F) \mbox{ where } F \models B \equiv B' \\
C \rightarrow x & \in & ITERATION(C',F) \mbox{ where } F \models C \equiv C'
\end{eqnarray*}

By Lemma~\ref{reconstruction-no-equivalent}, the reconstruction algorithm does
not determine both $ITERATION(B',F)$ and $ITERATION(C',F)$ if $F \models B'
\equiv C'$. Therefore, $F \not\models B \equiv C$.

Let $F'$ be a single-head formula that is equivalent to $F$, and $A \rightarrow
x$ be its only clause with head $x$. Because of its equivalence with $F$, it
entails both $B \rightarrow x$ and $C \rightarrow x$. Since these clauses are
produced by a valid iteration function, they are not tautologic: $x \not\in B$
and $x \not\in C$. Lemma~\ref{set-implies-set} tells that $F'$ contains $B''
\rightarrow x$ and $C'' \rightarrow x$ such that $F' \models B \rightarrow B''$
and $F' \models C \rightarrow C''$. Since the only clause of $F'$ that has $x$
in the head is $A \rightarrow x$, both $B''$ and $C''$ coincide with $A$. As a
result, $F'$ entails $B \rightarrow A$ and $C \rightarrow A$.

Since $F$ is equivalent to $F'$, it also does. This is the same as $A \leq_F B$
and $A \leq_F C$. These orderings are not strict. Otherwise, $A <_F B$ implies
$A <_F B'$, which implies $A \rightarrow x \in SCL(B',F)$, which implies $x
\not\in SFREE(B',F)$, which contradicts $B \rightarrow x \in ITERATION(B',F)$
since the latter includes $x \in SFREE(B',F)$. The same applies to $A <_F C$.
As a result, $F \models A \equiv B$ and $F \models A \equiv C$. The consequence
$F \models B \equiv C$ contradicts $F \not\models B \equiv C$, which was
previously proved.~\qed

This lemma completes the proof of soundness and correctness of the
reconstruction algorithm: it terminates with a single-head formula, if any; if
any exists, it returns it.

\subsection{Efficiency of the reconstruction algorithm}

The reconstruction algorithm works flawlessly in theory, but in practice its
iterations may be exponentially many because so many are the sets of variables.
A small change solves the problem: instead of all sets of variables, only
process the bodies of clauses of the formula. The algorithm produces the same
result because it runs like using a tie-breaking rule: if more than one set is
minimal, prefer one such that $ITERATION(B,F) = \emptyset$; if none is such,
choose a set that is the body of a clause of $F$. The following two lemmas show
this rule valid.

\begin{lemma}
\label{iteration-preconditions}

If $ITERATION(B,F) \not= \emptyset$ then $F \models B \equiv B''$ holds for
some clause $B'' \rightarrow x \in F$.

\end{lemma}

\proof Let $B' \rightarrow x$ be a clause of $ITERATION(B,F)$. The definition
of valid iteration includes $x \not\in B'$, $F \models B' \equiv B$, $F \models
B' \rightarrow x$ and $x \in SFREE(B,F)$.

From $F \models B' \rightarrow x$ and $x \not\in B'$
Lemma~\ref{set-implies-set} implies that $F$ contains a clause $B'' \rightarrow
x$ such that $F \models B' \rightarrow B''$. The converse implication $F
\models B'' \rightarrow B'$ may hold or not. If it does, then $F$ contains a
clause $B'' \rightarrow x$ with $F \models B' \equiv B''$. Since $F \models B
\equiv B'$, it follows $F \models B \equiv B''$ and the claim is proved.

The other case $F \not\models B'' \rightarrow B'$ is shown to lead to
contradiction. Since $F \models B \equiv B'$, the conditions $F \not\models B''
\rightarrow B'$ and $F \models B' \rightarrow B''$ extend from $B'$ to $B$.
This change turns them into $F \not\models B'' \rightarrow B$ and $F \models B
\rightarrow B''$. This is the definition of $B'' <_F B$. Since $B'' \rightarrow
x$ is in $F$, it is not a tautology and is entailed by $F$. With $B'' <_F B$,
these conditions define $B'' \rightarrow x \in SCL(B,F)$. This implies $x
\not\in SFREE(B,F)$, which conflicts with $x \in SFREE(B,F)$.~\qed

The choice of the minimal set $B$ in the construction algorithm is
non-deterministic. All lemmas proved so far hold for every possible
nondeterministic choices. Therefore, they also hold when restricting the
choices in whichever way. The following lemma tells that the candidate sets are
of two kinds: those which would generate an empty set of clauses, and those
equivalent to the body of a clause in the original formula.

\begin{lemma}
\label{precondition-or-empty}

At every step of the reconstruction algorithm
(Algorithm~\ref{algorithm-reconstruction}) if $P$ is not empty it contains
either a $\leq_F$-minimal set $B$ such that $ITERATION(B,F) = \emptyset$ or a
$\leq_F$-minimal set $B$ such that $B \rightarrow x \in F$.

\end{lemma}

\proof Since $P$ is not empty, it has at least a minimal element $B$ because
$\leq_F$ is a partial order. If $ITERATION(B,F)$ is empty, the claim is proved.
Otherwise, $ITERATION(B,F) \not= \emptyset$, and
Lemma~\ref{iteration-preconditions} applies: $F$ contains a clause $B''
\rightarrow x$ such that $F \models B \equiv B''$. Its body $B''$ is
$<_F$-minimal in $P$ because it is equivalent to $B$, which is minimal. Remains
to prove that $B''$ is in $P$.

By contradiction, $B'' \not\in P$ is assumed. Since $P$ initially contains all
sets of variables, $B''$ was removed at some point. The only operation removing
elements from $P$ removes all sets equivalent to a given one. Since it removes
$B''$ it also removes $B$ since it is equivalent to $B''$. This contradicts the
assumption that $B$ is in $P$.~\qed

If $ITERATION(B,F)$ is empty the iteration that chooses $B$ is irrelevant to
the final result since $G$ remains the same. The sets equivalent to $B$ could
just be removed from $P$. The other case is that $B$ is equivalent to a body
$B''$ of a clause of $F$. Since $B$ is minimal, also $B''$ is minimal.
Therefore, the algorithm could as well choose $B''$ instead of $B$. This means
that the algorithm may be limited to choosing sets of variables that are bodies
of clauses of $F$. The other sets either do not change $G$ or can be replaced
by them.

\begin{algorithm}[Precondition-reconstruction algorithm]
\label{algorithm-precondition-reconstruction}

\begin{enumerate}

\item $G = \emptyset$

\item $P = \{B \mid B \rightarrow x \in F \mbox{ for some variable x}\}$

\item while $P \not= \emptyset$:

\begin{enumerate}

\item choose a $<_F$-minimal $B \in P$
\newline
(a set such that $A <_F B$ does not hold for any $A \in P$)

\item $G = G \cup ITERATION(B,F)$

\item $P = P \backslash \{B' \mid F \models B \equiv B'\}$

\end{enumerate}

\item return $G$

\end{enumerate}

\end{algorithm}

The algorithm is almost the same as the basic reconstruction algorithm
(Algorithm~\ref{algorithm-reconstruction}). The only difference is that $P$ is
no longer initialized with all sets of variables, but only with the bodies of
the clauses of $F$. Its initial size is reduced from exponential to linear.
Since each iterations removes at least $B$, it makes $P$ strictly decrease. They
are therefore linearly many. The algorithm is still not polynomial-time since
$ITERATION(B,F)$ may not be polynomial.

This new, efficient version of the reconstruction algorithm is like the
original one with a restriction of the nondeterministic choices: if
$ITERATION(B,F) = \emptyset$ holds for some minimal sets $B$ of $P$, one of
them is chosen; otherwise, the body of a clause of $F$ is chosen.
Lemma~\ref{precondition-or-empty} guarantees that $P$ always contains a minimal
set of either kind.

The difference is that the original algorithm removes all sets $B'$ such that
$F \models B \equiv B'$ when $ITERATION(B,F) = \emptyset$; the new algorithm
virtually removes $B$ only. The following lemma shows that the other sets $B'$
also satisfy $ITERATION(B',F) = \emptyset$, and are therefore removed right
after $B$.

\begin{lemma}

If $ITERATION(B,F) = \emptyset$ and $F \models B \equiv B'$ then
$ITERATION(B',F) = \emptyset$.

\end{lemma}

\proof The proof shows contradictory the existence of a clause $B'' \rightarrow
x$ in $ITERATION(B',F)$. The definition of a valid iteration function includes
$x \not\in B''$, $F \models B'' \rightarrow x$, $F \models B \equiv B''$ and $x
\in SFREE(B,F)$. The first three conditions define $B'' \rightarrow x \in
BCL(B',F)$.

Since $BCL(B',F)$ and $SFREE(B',F)$ are defined from what $F$ entails, they
coincide with $BCL(B,F)$ and $SFREE(B,F)$ since $F \models B \equiv B'$. This
implies $B'' \rightarrow x \in BCL(B,F)$ and $x \in SFREE(B,F)$.

The definition of $ITERATION(B,F)$ includes
{} $SCL(B,F) \cup ITERATION(B,F) \models BCL(B,F)$.
Since $ITERATION(B,F)$ is empty, this is the same as $SCL(B,F) \models
BCL(B,F)$. As a result, $SCL(B,F)$ implies $B'' \rightarrow x$ while not
containing any positive occurrence of $x$ since $x \in SFREE(B,F)$. This is only
possible if $B'' \rightarrow x$ is tautologic, but $x \in B''$ contradicts $x
\not\in B''$.~\qed

The restriction of the nondeterministic choices does not affect the lemmas
about Algorithm~\ref{algorithm-reconstruction} proved so far. While not
explicitly mentioned, they do not assume anything about the nondeterministic
choices. As a result, they hold regardless of them. They hold even restricting
the nondeterministic choice in whichever way.

As a result, the new algorithm always terminates, always returns a formula
equivalent to the given one, and that formula is single-head if and only if any
is equivalent to the original, provided that $ITERATION(B,F)$ does not contain
two clauses of the same head for any $B$.

\

The modified reconstruction algorithm performs the loop a number of times equal
to the number of clauses of the input formula. More precisely, to the number of
different bodies in the formula; still, a linear number of times. If each
iteration turns out to be polynomial, establishing single-head equivalence
would be proved polynomial.

The problem is indeed in $ITERATION(B,F)$. A brute-force implementation is to
loop over all sets of clauses $B' \rightarrow x$ with $B' \equiv_F B$ and $x
\in SFREE(B,F)$, checking if one satisfies the conditions of validity.
Unfortunately, the sets $B'$ may be exponentially many. This obvious solution
is unfeasible.

Whether $ITERATION(B,F)$ can be determined in polynomial time is an open
problem. The following section gives an algorithm that may run in polynomial
time but is exponential-time in the worst case.

\subsection{Formulae that are not single-head equivalent}

All is good when the formula is single-head equivalent: an appropriate valid
iteration function makes the reconstruction algorithm return a single-head
formula that is equivalent to the given one. This is what happens when the
formula is single-head equivalent. What if it is not?

An easy and wrong answer is that the algorithm returns a formula that is not
single-head. A closer look at the lemmas shows that this is not always the
case.

Lemma~\ref{reconstruction-equivalent} tells that the reconstruction algorithm
returns a formula that is equivalent to the given one, but has a requirement: a
valid iteration function. That one exists is only guaranteed for single-head
formulae, not in general. If the formula is not single-head equivalent then
either:

\begin{itemize}

\item no valid iteration function exists;

\item a valid iteration function exists; it makes the algorithm generate a
formula equivalent to the given one by Lemma~\ref{reconstruction-equivalent}
and therefore not single-head.

\end{itemize}

Lemma~\ref{reconstruction-equivalent} splits the second case in two: either the
iteration function has some values containing two clauses with the same head or
it does not. All three cases are possible.

An example of a formula that has no valid iteration function is
{} $F = \{a \rightarrow b, b \rightarrow c, c \rightarrow b\}$,
in the following figure and in the {\tt inloop.py} test file of the {\tt
singlehead.py} program. While $ITERATION(\{b\},F)$ and $ITERATION(\{c\},F)$
satisfy the definition of validity if equal to $\{b \rightarrow c, c
\rightarrow b\}$, no value for $ITERATION(\{a\},F)$ does. Indeed, $SCL(\{a\},F)
= \{b \rightarrow c, c \rightarrow b\}$, which implies $SFREE(\{a\},F) =
\{a\}$. As a result, $ITERATION(\{a\},F)$ could only be empty. This is not
possible because $BCL(\{a\},F)$ contains $a \rightarrow c$ and $b \rightarrow
c$, which are not entailed by $SCL(\{a\},F)$.

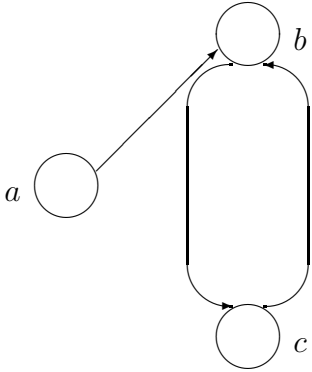
\setlength{\unitlength}{5000sp}%
\begingroup\makeatletter\ifx\SetFigFont\undefined%
\gdef\SetFigFont#1#2#3#4#5{%
  \reset@font\fontsize{#1}{#2pt}%
  \fontfamily{#3}\fontseries{#4}\fontshape{#5}%
  \selectfont}%
\fi\endgroup%
\begin{picture}(1449,1850)(6811,-6536)
\thinlines
{\color[rgb]{0,0,0}\put(7876,-5611){\oval(450,1200)[tl]}
\put(7876,-5611){\oval(450,1200)[bl]}
\put(7876,-6211){\vector( 1, 0){0}}
}%
{\color[rgb]{0,0,0}\put(8026,-5611){\oval(450,1200)[br]}
\put(8026,-5611){\oval(450,1200)[tr]}
\put(8026,-5011){\vector(-1, 0){0}}
}%
{\color[rgb]{0,0,0}\put(7051,-5611){\circle{336}}
}%
{\color[rgb]{0,0,0}\put(7951,-4861){\circle{336}}
}%
{\color[rgb]{0,0,0}\put(7951,-6361){\circle{336}}
}%
{\color[rgb]{0,0,0}\put(7201,-5536){\vector( 1, 1){600}}
}%
\put(6826,-5686){\makebox(0,0)[rb]{\smash{{\SetFigFont{12}{24.0}
{\rmdefault}{\mddefault}{\updefault}{\color[rgb]{0,0,0}$a$}%
}}}}
\put(8176,-4936){\makebox(0,0)[lb]{\smash{{\SetFigFont{12}{24.0}
{\rmdefault}{\mddefault}{\updefault}{\color[rgb]{0,0,0}$b$}%
}}}}
\put(8176,-6436){\makebox(0,0)[lb]{\smash{{\SetFigFont{12}{24.0}
{\rmdefault}{\mddefault}{\updefault}{\color[rgb]{0,0,0}$c$}%
}}}}
\end{picture}%
\nop{
a -> b <-> c
}

An example of a formula that has a valid iteration function but all of them
have values with two clauses with the same head is
{} $F = \{x \rightarrow a, a \rightarrow d, x \rightarrow b, b \rightarrow c,
{}        ac \rightarrow x, bd \rightarrow x \}$.
This formula is in the {\tt disjointnotsingle.py} test file of the {\tt
singlehead.py} program. The following figure shows it. A valid iteration
function is
{} $ITERATION(\{a\},F) = \{a \rightarrow d\}$,
{} $ITERATION(\{b\},F) = \{b \rightarrow c\}$,
{} $ITERATION(\{x\},F) =
{}  ITERATION(\{a,b\},F) =
{}  ITERATION(\{a,c\},F) =
{}  ITERATION(\{d,b\},F) = 
{} \{x \rightarrow a, x \rightarrow b, ac \rightarrow x, bd \rightarrow x\}$,
but the last values contain two clauses with the same head $x$. This is always
the case. In particular, if $B=\{x\}$ then $ITERATION(B,F)$ is required to
entail $x \rightarrow a$, $x \rightarrow b$, $ac \rightarrow x$ and $bd
\rightarrow x$, but $SCL(B,F) = \{a \rightarrow d, b \rightarrow c\}$ does not
help at all. Three heads $x$, $a$ and $b$ do not suffice for four bodies.

%
%  two cycles of single variables, running in different directions 
%
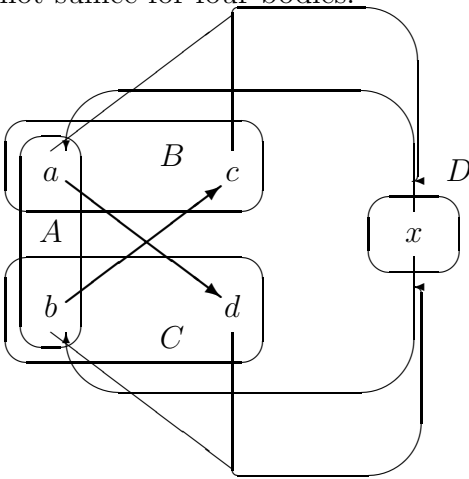
\setlength{\unitlength}{5000sp}%
\begingroup\makeatletter\ifx\SetFigFont\undefined%
\gdef\SetFigFont#1#2#3#4#5{%
  \reset@font\fontsize{#1}{#2pt}%
  \fontfamily{#3}\fontseries{#4}\fontshape{#5}%
  \selectfont}%
\fi\endgroup%
\begin{picture}(2277,2342)(7564,-6931)
\thinlines
{\color[rgb]{0,0,0}\put(7876,-6211){\vector( 0, 1){0}}
\put(8614,-6211){\oval(1476,600)[bl]}
\put(8614,-5836){\oval(1974,1350)[br]}
}%
{\color[rgb]{0,0,0}\put(8627,-5611){\oval(1948,1200)[tr]}
\put(8627,-5311){\oval(1502,600)[tl]}
\put(7876,-5311){\vector( 0,-1){0}}
}%
{\color[rgb]{0,0,0}\put(8926,-6886){\oval(450, 74)[bl]}
\put(8926,-6211){\oval(1424,1424)[br]}
\put(9601,-6211){\oval( 74,450)[tr]}
\put(9601,-5986){\vector(-1, 0){0}}
}%
{\color[rgb]{0,0,0}\put(9601,-5461){\vector(-1, 0){0}}
\put(9601,-5294){\oval( 40,334)[br]}
\put(8926,-5294){\oval(1390,1390)[tr]}
\put(8926,-4636){\oval(450, 74)[tl]}
}%
{\color[rgb]{0,0,0}\put(7756,-6181){\oval(210,210)[bl]}
\put(7756,-5341){\oval(210,210)[tl]}
\put(7846,-6181){\oval(210,210)[br]}
\put(7846,-5341){\oval(210,210)[tr]}
\put(7756,-6286){\line( 1, 0){ 90}}
\put(7756,-5236){\line( 1, 0){ 90}}
\put(7651,-6181){\line( 0, 1){840}}
\put(7951,-6181){\line( 0, 1){840}}
}%
{\color[rgb]{0,0,0}\put(9481,-5806){\oval(210,210)[bl]}
\put(9481,-5641){\oval(210,210)[tl]}
\put(9721,-5806){\oval(210,210)[br]}
\put(9721,-5641){\oval(210,210)[tr]}
\put(9481,-5911){\line( 1, 0){240}}
\put(9481,-5536){\line( 1, 0){240}}
\put(9376,-5806){\line( 0, 1){165}}
\put(9826,-5806){\line( 0, 1){165}}
}%
{\color[rgb]{0,0,0}\put(7681,-5506){\oval(210,210)[bl]}
\put(7681,-5266){\oval(210,210)[tl]}
\put(8746,-5506){\oval(210,210)[br]}
\put(8746,-5266){\oval(210,210)[tr]}
\put(7681,-5611){\line( 1, 0){1065}}
\put(7681,-5161){\line( 1, 0){1065}}
\put(7576,-5506){\line( 0, 1){240}}
\put(8851,-5506){\line( 0, 1){240}}
}%
{\color[rgb]{0,0,0}\put(7681,-6256){\oval(210,210)[bl]}
\put(7681,-5941){\oval(210,210)[tl]}
\put(8746,-6256){\oval(210,210)[br]}
\put(8746,-5941){\oval(210,210)[tr]}
\put(7681,-6361){\line( 1, 0){1065}}
\put(7681,-5836){\line( 1, 0){1065}}
\put(7576,-6256){\line( 0, 1){315}}
\put(8851,-6256){\line( 0, 1){315}}
}%
\thicklines
{\color[rgb]{0,0,0}\put(7876,-6061){\vector( 4, 3){768}}
}%
{\color[rgb]{0,0,0}\put(7876,-5461){\vector( 4,-3){768}}
}%
\thinlines
{\color[rgb]{0,0,0}\put(7801,-5311){\line( 4, 3){900}}
\put(8701,-4636){\line( 0,-1){675}}
}%
{\color[rgb]{0,0,0}\put(7801,-6211){\line( 4,-3){900}}
\put(8701,-6886){\line( 0, 1){675}}
}%
\put(7801,-5461){\makebox(0,0)[b]{\smash{{\SetFigFont{12}{24.0}
{\rmdefault}{\mddefault}{\updefault}{\color[rgb]{0,0,0}$a$}%
}}}}
\put(8701,-5461){\makebox(0,0)[b]{\smash{{\SetFigFont{12}{24.0}
{\rmdefault}{\mddefault}{\updefault}{\color[rgb]{0,0,0}$c$}%
}}}}
\put(7801,-6136){\makebox(0,0)[b]{\smash{{\SetFigFont{12}{24.0}
{\rmdefault}{\mddefault}{\updefault}{\color[rgb]{0,0,0}$b$}%
}}}}
\put(8701,-6136){\makebox(0,0)[b]{\smash{{\SetFigFont{12}{24.0}
{\rmdefault}{\mddefault}{\updefault}{\color[rgb]{0,0,0}$d$}%
}}}}
\put(9601,-5761){\makebox(0,0)[b]{\smash{{\SetFigFont{12}{24.0}
{\rmdefault}{\mddefault}{\updefault}{\color[rgb]{0,0,0}$x$}%
}}}}
\put(9826,-5461){\makebox(0,0)[b]{\smash{{\SetFigFont{12}{24.0}
{\rmdefault}{\mddefault}{\updefault}{\color[rgb]{0,0,0}$D$}%
}}}}
\put(8401,-6286){\makebox(0,0)[b]{\smash{{\SetFigFont{12}{24.0}
{\rmdefault}{\mddefault}{\updefault}{\color[rgb]{0,0,0}$C$}%
}}}}
\put(8401,-5386){\makebox(0,0)[b]{\smash{{\SetFigFont{12}{24.0}
{\rmdefault}{\mddefault}{\updefault}{\color[rgb]{0,0,0}$B$}%
}}}}
\put(7801,-5761){\makebox(0,0)[b]{\smash{{\SetFigFont{12}{24.0}
{\rmdefault}{\mddefault}{\updefault}{\color[rgb]{0,0,0}$A$}%
}}}}
\end{picture}%
\nop{
+----------------------+
|                      |
|   +--------+-----+   |
|   |        |     |   |
+-> a -\ +-> c     V --+               B=ac
        X    +---> x           A=ab            D=x
+-> b -/ +-> d     ^ --+               C=bd
|   |        |     |   |
|   +--------+-----+   |
|                      |
+----------------------+
}

An example of a formula that is not single-head equivalent but has a valid
iteration function that never returns a set containing two clauses with the
same head is
{} $F = \{a \rightarrow c, b \rightarrow c\}$,
in the following figure and in the {\tt samehead.py} test file of the {\tt
singlehead.py} program. Its only valid iteration function is defined by
{} $ITERATION(\{a\},F) = \{a \rightarrow c\}$
and
{} $ITERATION(\{b\},F) = \{b \rightarrow c\}$.
The reconstruction algorithm either processes $\{a\}$ first and $\{b\}$ second
or vice versa. In both cases it adds both $ITERATION(\{a\},F)$ and
$ITERATION(\{b\},F)$ to the resulting formula, which ends up containing two
clauses with head $c$.

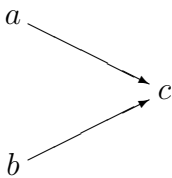
\setlength{\unitlength}{5000sp}%
\begingroup\makeatletter\ifx\SetFigFont\undefined%
\gdef\SetFigFont#1#2#3#4#5{%
  \reset@font\fontsize{#1}{#2pt}%
  \fontfamily{#3}\fontseries{#4}\fontshape{#5}%
  \selectfont}%
\fi\endgroup%
\begin{picture}(780,885)(5011,-4051)
\thinlines
{\color[rgb]{0,0,0}\put(5101,-3961){\vector( 2, 1){600}}
}%
{\color[rgb]{0,0,0}\put(5101,-3286){\vector( 2,-1){600}}
}%
\put(5026,-4036){\makebox(0,0)[b]{\smash{{\SetFigFont{12}{24.0}
{\rmdefault}{\mddefault}{\updefault}{\color[rgb]{0,0,0}$b$}%
}}}}
\put(5026,-3286){\makebox(0,0)[b]{\smash{{\SetFigFont{12}{24.0}
{\rmdefault}{\mddefault}{\updefault}{\color[rgb]{0,0,0}$a$}%
}}}}
\put(5776,-3661){\makebox(0,0)[b]{\smash{{\SetFigFont{12}{24.0}
{\rmdefault}{\mddefault}{\updefault}{\color[rgb]{0,0,0}$c$}%
}}}}
\end{picture}%
\nop{
a ---> c <--- b
}

The algorithm as implemented in the {\tt singlehead.py} program does not deal
with these three cases separately. It needs not. The optimizations in the
search for an iteration function turn all three cases into the first, where no
valid iteration function exists.

The reason is that $ITERATION(B,F)$ is not calculated in advance, but during
the iteration for $B$. A number of candidate sets of clauses are tried.
Excluding the ones with duplicated heads is actually a simplification. So is
excluding the ones that contain some heads of another value $ITERATION(A,F)$
calculated before, which happens in two cases: when no valid iteration function
exists and when the output formula would not be single-head.

\section{The iteration function}
\label{iteration}

The reconstruction algorithm uses a function $ITERATION(B,F)$ at each
iteration. Only its properties have been given so far. If $F$ is single-head
equivalent such a function is easily proved to exist, but how to calculate it
is an entirely different story. This long section is exclusively devoted to
this problem.

At least, the value of the function $ITERATION(B,F)$ for $B$ is independent of
that for $B'$. Even if $B \equiv_F B'$, but also if $B \not\equiv_F B'$, it is
not required to be the same. This is a simplification because the computation
of $ITERATION(B,F)$ is only affected by the set of variables $B$, not by any
other set $B'$.

The requirements on $ITERATION(B,F)$ are listed in
Definition~\ref{condition-iteration}: is a set of non-tautological clauses $B'
\rightarrow x$ such that $F \models B' \rightarrow x$, $B' \equiv_F B$, $x \in
SFREE(B,F)$ and $SCL(B,F) \cup ITERATION(B,F) \models BCL(B,F)$.

The algorithm presented in this section is implemented in the Python program
{\tt singlehead.py}: it first determines the heads of the clauses; it then
calculates the set of all possible bodies to be associated with them; for each
association, the resulting set is checked. The first two steps automatically
guarantee the first three conditions, the second checks the fourth.

The heads are uniquely determined. This is not the case for the bodies, nor it
is the case for their associations with the heads. Efficiency depends on how
many alternatives can be excluded in advance, and how fast they can be
discarded if they are not valid.

\subsection{Heads in search of bodies}

The heads of the clauses of $ITERATION(B,F)$ are uniquely determined. The
condition of validity implies an exact set of heads; two valid iteration
functions do not differ on their heads. If $ITERATION(B,F)$ is also required
not to contain duplicated heads, it must contain exactly a clause for each
head.

The starting point is $BCN(B,F) = \{x \mid F \models B \rightarrow x\}$, the
set of variables entailed by $F$ and $B$. A valid iteration function
$ITERATION(B,F)$ contains only clauses with head in $BCN(B,F)$. Not the other
way around, because:

\begin{itemize}

\item $ITERATION(B,F)$ does not contain heads in $SFREE(B,F)$ but $BCN(B,F)$
does;

\item $ITERATION(B,F)$ does not contain heads that are in all sets equivalent
to $B$, but $BCN(B,F)$ does; if a clause $B' \rightarrow x$ is in
$ITERATION(B,F)$ then $B' \equiv_F B$; if $x$ is in all sets equivalent to $B$
then $x \in B'$; the clause $B' \rightarrow x$ is true but tautological.

\end{itemize}

The second point may look moot. As a matter of facts, it is not. An example
clarifies why $ITERATION(B,F)$ cannot include tautologies. The formula is
{} $F = \{ab \rightarrow c, ac \rightarrow b, d \rightarrow a\}$,
in the following figure and in the {\tt equiall.py} test file. The sets
$\{a,b\}$ and $\{a,c\}$ are equivalent. The clauses $ab \rightarrow a$ and $ac
\rightarrow a$ are tautological. While they look harmless, useless but
harmless, they are not. Once in $ITERATION(\{a,b\},F)$, they conflict with
$ITERATION(\{d\},F)$, which needs $d \rightarrow a$. The formula returned by
the reconstruction algorithm would not be single-head while $F$ is single-head
equivalent, being single-head. Variables like $a$ are not possible heads for
$\{a,b\}$ although they are in $BCN(\{a,b\},F)$ and may be in
$SFREE(\{a,b\},F)$. Excluding tautologies excludes them.

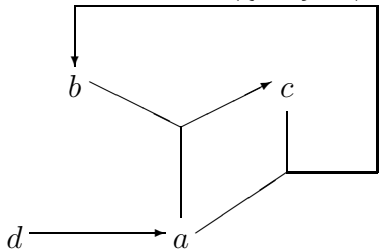
\setlength{\unitlength}{5000sp}%
\begingroup\makeatletter\ifx\SetFigFont\undefined%
\gdef\SetFigFont#1#2#3#4#5{%
  \reset@font\fontsize{#1}{#2pt}%
  \fontfamily{#3}\fontseries{#4}\fontshape{#5}%
  \selectfont}%
\fi\endgroup%
\begin{picture}(1834,1384)(5079,-4433)
\thinlines
{\color[rgb]{0,0,0}\put(5476,-3436){\line( 2,-1){450}}
\put(5926,-3661){\line( 0,-1){450}}
}%
{\color[rgb]{0,0,0}\put(5926,-3661){\vector( 2, 1){450}}
}%
{\color[rgb]{0,0,0}\put(6001,-4186){\line( 3, 2){450}}
\put(6451,-3886){\line( 0, 1){300}}
}%
{\color[rgb]{0,0,0}\put(6451,-3886){\line( 1, 0){450}}
\put(6901,-3886){\line( 0, 1){825}}
\put(6901,-3061){\line(-1, 0){1500}}
\put(5401,-3061){\vector( 0,-1){300}}
}%
{\color[rgb]{0,0,0}\put(5176,-4186){\vector( 1, 0){675}}
}%
\put(5101,-4261){\makebox(0,0)[b]{\smash{{\SetFigFont{12}{24.0}
{\rmdefault}{\mddefault}{\updefault}{\color[rgb]{0,0,0}$d$}%
}}}}
\put(5401,-3511){\makebox(0,0)[b]{\smash{{\SetFigFont{12}{24.0}
{\rmdefault}{\mddefault}{\updefault}{\color[rgb]{0,0,0}$b$}%
}}}}
\put(5926,-4261){\makebox(0,0)[b]{\smash{{\SetFigFont{12}{24.0}
{\rmdefault}{\mddefault}{\updefault}{\color[rgb]{0,0,0}$a$}%
}}}}
\put(6451,-3511){\makebox(0,0)[b]{\smash{{\SetFigFont{12}{24.0}
{\rmdefault}{\mddefault}{\updefault}{\color[rgb]{0,0,0}$c$}%
}}}}
\end{picture}%
\nop{
   +---------------+
   |               |
   V               |
   b ---+---> c ---+
        |          |
        |          |
d ----> a ---------+
}

Finding such variables is easy using the real consequences~\cite{libe-20-b}:
{} $RCN(B,F) = \{x \mid F \cup (BCN(B,F) \backslash \{x\}) \models x\}$.
They are the variables that follow from $B$ because of some clauses of $F$, not
just because they are in $B$.

\begin{lemma}
\label{bcn-rcn}

If $x \in BCN(B,F) \backslash RCN(B,F)$ and $F \models B \equiv B'$, then $x
\in B'$.

\end{lemma}

\proof The contrary of the claim is proved contradictory: $x \not\in B'$.

Since $x$ is in $BCN(B,F) \backslash RCN(B,F)$, it is in $BCN(B,F)$ but 
not in $RCN(B,F)$.

The first condition $x \in BCN(B,F)$ is defined as $F \cup B \models x$. With
the assumption $F \models B \equiv B'$, it implies $F \cup B' \models x$.

The second condition $x \not\in RCN(B,F)$ is defined as
{} $F \cup (BCN(B,F) \backslash \{x\}) \not\models x$.
The assumption $F \models B \equiv B'$ implies $F \models B \rightarrow B'$,
which in turn implies $B' \subseteq BCN(B,F)$. Because of the additional
assumption $x \not\in B'$, this containment strengthens to $B' \subseteq
BCN(B,F) \backslash \{x\}$. Since
{} $F \cup (BCN(B,F) \backslash \{x\}) \not\models x$,
the entailment
{} $F \cup B' \not\models x$
follows by monotonicity, contradicting $F \cup B' \models x$.~\qed

This lemma tells that a variable in $BCN(B,F)$ but not in $RCN(B,F)$ belongs to
all sets $B'$ that are equivalent to $B$ according to $F$. The following lemma
proves something in the opposite direction: the variables in $RCN(B,F)$ may be
heads of $ITERATION(B,F)$.

\begin{lemma}
\label{rcn-bcl}

If $x \in RCN(B,F)$ then $B' \rightarrow x \in BCL(B,F)$ for some $B'$.

\end{lemma}

\proof Since $F \cup B$ implies every variable in $BCN(B,F)$ by definition, it
implies $BCN(B,F)$. A consequence of the deduction theorem is $F \models B
\rightarrow BCN(B,F)$. Every variable in $B$ is entailed by $F \cup B$ and is
therefore in $BCN(B,F)$. A consequence of $B \subseteq BCN(B,F)$ is $F \models
BCN(B,F) \rightarrow B$. Equivalence is therefore proved: $F \models B \equiv
BCN(B,F)$.

The assumption $x \in RCN(B,F)$ is defined as
{} $F \cup (BCN(B,F) \backslash \{x\}) \models x$.
Combined with the tautological entailment
{} $F \cup (BCN(B,F) \backslash \{x\}) \models BCN(B,F) \backslash \{x\}$,
it implies
{} $F \cup (BCN(B,F) \backslash \{x\}) \models
{}    BCN(B,F) \backslash \{x\} \cup \{x\}$.
This is the same as
{} $F \cup (BCN(B,F) \backslash \{x\}) \models BCN(B,F)$,
which is the same as
{} $F \models BCN(B,F) \backslash \{x\} \rightarrow BCN(B,F)$.
The converse implication is a consequence of monotony. This proves the
equivalence $F \models BCN(B,F) \backslash \{x\} \equiv BCN(B,F)$. By
transitivity of equivalence,
{} $F \models B \equiv BCN(B,F) \backslash \{x\}$.

The set $B'$ required by the claim of the lemma is $BCN(B,F) \backslash \{x\}$.
The final consequence of the last paragraph is $F \models B \equiv B'$. Another
entailment proved is
{} $F \cup (BCN(B,F) \backslash \{x\}) \models x$,
which is the same as $F \models B' \rightarrow x$. The last requirement for $B'
\rightarrow x$ to be in $BCL(B,F)$ is $x \not\in B'$, which is the case because
$B'$ is by construction the result of removing $x$ from another set.~\qed

This is not exactly the reverse of the previous lemma, in that it does not
precisely prove that $RCN(B,F)$ contains the variables that are in some sets
equivalent to $B$ but not all. Yet, it allows for a simple way for determining
the heads of the clauses in $ITERATION(B,F)$.

\begin{lemma}
\label{iteration-heads}

The heads of the clauses in $ITERATION(B,F)$ for a valid iteration function is
{} $HEADS(B,F) = RCN(B,F) \cap SFREE(B,F)$.

\end{lemma}

\proof The proof comprises two parts: first, if $B' \rightarrow x \in
ITERATION(B,F)$ then $x \in HEADS(B,F)$; second, if $x \in HEADS(B,F)$ then $B'
\rightarrow x \in ITERATION(B,F)$ for some set of variables $B'$.

The first part assumes $B' \rightarrow x \in ITERATION(B,F)$. The definition of
valid iteration includes $x \not\in B'$, $F \models B \equiv B'$, $F \models B'
\rightarrow x$ and $x \in SFREE(B,F)$. A consequence of $F \models B \equiv B'$
is that every element of $B'$ is entailed by $F \cup B$, and is therefore in
$BCN(B,F)$. This proves the containment $B' \subseteq BCN(B,F)$. Since $x$ is
not in $B'$, this containment strengthens to $B' \subseteq BCN(B,F) \backslash
\{x\}$. The assumption $F \models B' \rightarrow x$ implies $F \cup B' \models
x$. Since $B' \subseteq BCN(B,F) \backslash \{x\}$, this implies $F \cup
(BCN(B,F) \backslash \{x\}) \models x$. This entailment defines $x \in
RCN(B,F)$. Since $x \in SFREE(B,F)$ also holds, the claim $x \in HEADS(B,F)$
follows.

The second part of the proof assumes $x \in HEADS(B,F)$. This is defined as $x
\in RCN(B,F)$ and $x \in SFREE(B,F)$. By Lemma~\ref{rcn-bcl}, the first implies
$B' \rightarrow x \in BCL(B,F)$ for some $B'$. By the assumption of validity of
the iteration function, $SCL(B,F) \cup ITERATION(B,F)$ implies $B' \rightarrow
x$. By Lemma~\ref{set-implies-set}, $SCL(B,F) \cup ITERATION(B,F)$ contains a
clause $B'' \rightarrow x$. Since $x \in SFREE(B,F)$, this clause is not in
$SCL(B,F)$, and is therefore in $ITERATION(B,F)$.~\qed

The name of this function $HEADS(B,F)$ needs a clarification. The aim of
$ITERATION(B,F)$ is to reproduce $BCL(B,F)$. Since $HEADS(B,F)$ are the heads
of $ITERATION(B,F)$, they are the heads needed to reproduce $BCL(B,F)$. Not.

A part of the definition of $ITERATION(B,F)$ is that its heads are all in
$SFREE(B,F)$. This restriction presumes the formula single-head equivalent: it
only works if variables that are not heads in $SCL(B,F)$ suffice as heads of
clauses that reproduce $BCL(B,F)$. If they do not, $ITERATION(B,F)$ is unable
to imply $BCL(B,F)$. The following formula exemplifies such a situation.

\[
F = \{a \rightarrow x, x \rightarrow b, b \rightarrow x\}
\]

The heads of $SCL(\{a\},F)$ are $\{b,x\}$. The other variables are
$SFREE(\{a\},F) = \{a\}$. Therefore, $HEADS(\{a\},F) = \emptyset$ since $a
\not\in RCN(\{a\},F)$. The heads of $SCL(\{a\},F)$ and $ITERATION(\{a\},F)$ are
disjoint by definition, this is not the problem. The problem is that the
available heads $HEADS(\{a\},F) = \emptyset$ are insufficient for building
clauses that entail $BCL(B,F)$.

\

The first step for obtaining $ITERATION(B,F)$ is generating its heads. Finding
the bodies and attaching them to these heads is the hard part of the
reconstruction algorithm. Finding the heads is not.

\subsection{Every body on its own heads}

A side effect of Lemma~\ref{iteration-heads} is an equivalent condition to
single-head equivalence. The following is a preliminary lemma.

\begin{lemma}
\label{reconstruction-heads}

If a valid iteration function is such that $ITERATION(B,F)$ does not contain
multiple clauses with the same head for any $B$, the return value of the
reconstruction algorithm is single-head if and only if $HEADS(B,F)$ and
$HEADS(B',F)$ do not overlap for any $B$ and $B'$ such that $F \not\models B
\equiv B'$.

\end{lemma}

\proof One direction of the claim is that if the reconstruction algorithm
returns two clauses with the same head then $HEAD(B,F) \cap HEAD(B',F) \not=
\emptyset$ for some $B$ and $B'$ such that $F \not\models B \equiv B'$. The
reconstruction algorithm returns the union of all sets $ITERATION(B,F)$ it
calculates. Since this union contains two clauses of the same head, one comes
from $ITERATION(B,F)$ and one from $ITERATION(B',F)$ because no single set
$ITERATION(B,F)$ contains multiple clauses with the same head.
Lemma~\ref{reconstruction-no-equivalent} tells that the reconstruction
algorithm does not determine both $ITERATION(B,F)$ and $ITERATION(B',F)$ if $F
\models B \equiv B'$. This proves $F \not\models B \equiv B'$. The heads of
$ITERATION(B,F)$ and $ITERATION(B',F)$ are respectively $HEADS(B,F)$ and
$HEADS(B',F)$ by Lemma~\ref{iteration-heads}. These sets overlap because
$ITERATION(B,F)$ contains a clause with the same head as one of
$ITERATION(B',F)$. This proves $HEADS(B,F) \cap HEADS(B',F) \not= \emptyset$
with $F \not\models B \equiv B'$.

The other direction of the claim is that $HEADS(B,F) \cap HEADS(B',F) \not=
\emptyset$ with $F \not\models B \equiv B'$ implies that the reconstruction
algorithm returns two clauses with the same head.
Lemma~\ref{reconstruction-one-equivalent} states that the reconstruction
algorithm determines $ITERATION(B'',F)$ for some $B''$ such that $F \models B
\equiv B''$. The heads of the clauses in $ITERATION(B'',F)$ are $HEADS(B'',F)$,
which is the same as $HEADS(B,F)$ because $F \models B \equiv B''$. For the
same reason, the reconstruction algorithm also determines $ITERATION(B''',F)$
with $F \models B' \equiv B'''$, and its heads are $HEADS(B',F)$. Let $x \in
HEADS(B,F) \cap HEADS(B',F)$. Both $ITERATION(B'',F)$ and $ITERATION(B''',F)$
contain a clause of head $x$. The body cannot be the same because the first is
equivalent to $B''$ and the second to $B'''$, and these sets are respectively
equivalent to $B$ and $B'$, which are not equivalent by assumption. As a
result, these two clauses have the same head but different bodies.~\qed

The condition that the sets $HEADS(B,F)$ are disjoint is logical, not
algorithmic. It is independent of the reconstruction algorithm. It depends on
the definition of $HEADS(B,F)$ which is based on $RCN(B,F)$ and $SFREE(B,F)$,
both defined in terms of entailments. It turns the overall search for a
single-head formula equivalent to $F$ into the local search for the clauses of
$ITERATION(B,F)$.

\begin{lemma}
\label{disjoint-heads-all}

A formula $F$ is equivalent to a single-head formula if and only if a valid
iteration function is such that $ITERATION(B,F)$ does not contain two clauses
with the same head for any $B$ and $HEADS(B,F) \cap HEADS(B',F)$ is empty for
every $B$ and $B'$ such that $F \not\models B \equiv B'$.

\end{lemma}

\proof The long statement of the lemma is shortened by naming its conditions:
{} (a) a valid iteration function is such that $ITERATION(B,F)$
{}     does not contain two clauses with the same head for any $B$;
{} (b) $HEADS(B,F) \cap HEADS(B',F)$ is empty
{}     for every $B$ and $B'$ such that $F \not\models B \equiv B'$.
The claim is that $F$ is single-head equivalent if and only if both (a) and (b)
hold.

The first direction of the claim assumes (a) and (b) and proves $F$ single-head
equivalent. Lemma~\ref{reconstruction-heads} proves that the return value of
the reconstruction algorithm is single-head from (a) and (b).
Lemma~\ref{reconstruction-equivalent} proves that the reconstruction algorithm
returns a formula equivalent to $F$ when it uses a valid iteration function,
which is part of (a). This proves that $F$ is single-head equivalent, as
required.

The other direction of the claim assumes $F$ single-head equivalent and proves
(a) and (b).

Since $F$ is equivalent to a single-head formula $F'$,
Lemma~\ref{equivalent-valid} applies:
{} $ITERATION(B,F) = \{ B' \rightarrow x \in F' \mid F' \models B \equiv B' \}$
is a valid iteration function.

It is proved to meet condition (a) by contradiction. Let $B'' \rightarrow x$
and $B''' \rightarrow x$ be two different clauses of $ITERATION(B,F)$. They are
in $F'$ because $ITERATION(B,F)$ only contains clauses of $F'$. This implies
that $F'$ is not single-head, while it is assumed to be.

Condition (b) is proved by contradiction. Its converse is $x \in HEADS(B,F)$
and $x \in HEADS(B',F)$ with $F \not\models B \equiv B'$.
Lemma~\ref{iteration-heads} proves that the heads of $ITERATION(B,F)$ and
$ITERATION(B',F)$ are respectively $HEADS(B,F)$ and $HEADS(B',F)$. Therefore, a
clause $B'' \rightarrow x$ is in $ITERATION(B,F)$ and a clause $B'''
\rightarrow x$ is in $ITERATION(B',F)$. Both sets comprise only clauses of $F'$
by construction. Therefore, both clauses are in $F'$. The definition of
iteration function includes $F \models B \equiv B''$ and $F \models B' \equiv
B'''$. The assumption $F \not\models B \equiv B'$ implies $F \not\models B''
\equiv B'''$, which implies $B'' \not= B'''$. Two clauses of the same head but
different bodies are in $F'$, which was assumed single-head.~\qed

The condition as stated by the lemma is infeasible to check since exponentially
many sets of variables exist. However, only the sets that are bodies of clauses
of the formula need to be checked.

\begin{lemma}
\label{disjoint-heads}

A formula $F$ is equivalent to a single-head formula if and only if there
exists a valid $ITERATION(B,F)$ that never produces two clauses with the same
head and $HEADS(B,F) \cap HEADS(B',F)$ is empty for every clauses $B
\rightarrow x$ and $B' \rightarrow y$ of $F$ with $F \not\models B \equiv B'$.

\end{lemma}

\proof If $F$ is single-head equivalent, $HEADS(B,F) \cap HEADS(B',F)$ is empty
for every pair of sets $B$ and $B'$ that are not equivalent according to
$\equiv_F$ by Lemma~\ref{disjoint-heads-all}. It is therefore empty in the
particular case where $B$ and $B'$ are bodies of clauses of $F$. This is one
direction of the claim.

The other direction requires proving that $F$ is single-head equivalent from
the other properties mentioned in the claim. The first condition includes the
existence of a valid iteration function $ITERATION(B,F)$. The second is that
$HEADS(B,F) \cap HEADS(B',F)$ is empty whenever $F \not\models B \equiv B'$ and
$B$ and $B'$ are bodies of clauses of $F$. When a valid iteration function
exists, the second extends to the case where $B$ and $B'$ that are not bodies
of clauses of $F$. The claim follows by Lemma~\ref{disjoint-heads-all}.

By Lemma~\ref{iteration-preconditions}, either $ITERATION(B,F) = \emptyset$ or
$F \models B \equiv B''$ hold for some clause $B'' \rightarrow x \in F$. If
$ITERATION(B,F)$ is empty so is $HEADS(B,F)$ because the latter is the set of
the heads of the former by Lemma~\ref{iteration-heads}. In this case,
$HEADS(B,F) \cap HEADS(B',F) = \emptyset$ follows. The same holds for $B'$ by
symmetry. The claim is therefore proved if either $B$ or $B'$ are not each
$\equiv_F$-equivalent to a body of $F$.

Remains to prove it if both $B$ and $B'$ are equivalent to a body of $F$ each.
Let $B''$ and $B'''$ be these bodies. Since $B$ and $B'$ are
$\equiv_F$-equivalent, so are $B''$ and $B''$. Since $F \models B'' \equiv
B'''$ and $HEADS(B,F)$ is defined from what $F$ entails, it coincides with
$HEADS(B''',F)$. For the same reason, $HEADS(B',F)$ coincides with
$HEADS(B'''',F)$. Since $B''$ and $B'''$ are bodies of clauses of $F$,
$HEADS(B'',F) \cap HEADS(B''',F) = \emptyset$ holds by assumption. It implies
$HEADS(B,F) \cap HEADS(B',F) = \emptyset$, the claim.~\qed

\subsection{Head-on}

Lemma~\ref{iteration-heads} tells the heads of the clauses that may be in
$ITERATION(B,F)$, but also helps to find their bodies: since the heads are
known and the clauses are consequences of $F$, the bodies are found by bounding
resolution to the heads.

\begin{lemma}
\label{head-resolution}

If $F \models B' \rightarrow x$ and $x \not\in B'$, a clause with head $x$ and
body contained in $B'$ is obtained by repeatedly resolving a non-tautological
clause of $F$ with clauses of $F^x$.

\end{lemma}

\proof Lemma~\ref{set-implies-set} applied to $F \models B' \rightarrow x$ and
$x \not\in B'$ proves $F^x \models B' \rightarrow B$ for some $B \rightarrow x
\in F$. This is the clause of $F$ in the statement of the lemma. The rest of
the proof shows that it is not a tautology and that it resolves with a sequence
of clauses of $F^x$, resulting in a subclause of $B' \rightarrow x$.

The clause $B \rightarrow x$ is first proved not to be a tautology. The
entailment $F^x \models B' \rightarrow B$ is the same as $F^x \cup \{B'\}
\models B$. The formula $F^x \cup \{B'\}$ does not contain $x$ by the
definition of $F^x$ and the assumption that $B' \rightarrow x$ is not a
tautology ($x \not\in B'$). Since it is definite Horn and does not contain $x$
it satisfied by the model that sets all variables to true but $x$: all clauses
are satisfied since they contain at least a positive variable, which is not
$x$. This model does not satisfy $x$, proving $F^x \cup \{B'\} \models x$
impossible and disproving $x \in B$.

Remains to show how to resolve $B \rightarrow x$ with the clauses of $F^x$. It
immediately follows from a property proved by induction.

\

The induction is on the size of a formula $G$. The property is: if $G$ entails
$B' \rightarrow C$ then $C \cup D \rightarrow x$ resolves with some clauses of
$G$ into $B'' \cup D \rightarrow x$ where $B''$ is a subset of $B'$. The clause
$C \cup D \rightarrow x$ is not required to be in $G$.

Since induction is on the size of $G$, the base case is that $G$ is empty. An
empty formula implies $B' \rightarrow C$ only if $C \subseteq B'$. The claim
holds with a zero-length resolution, since $C \cup D \rightarrow x$ already
satisfies $C \subseteq B'$. This proves the base case of induction.

The inductive case is for an arbitrary formula $G$.

If $C$ is a subset of $B'$, the claim again holds because a zero-length
resolution turns $C \cup D \rightarrow x$ into itself, which satisfies $C
\subseteq B'$.

If $C$ is not a subset of $B'$, then $C \backslash B'$ contains some variables.
Let $c \in C \backslash B'$ be one of them: $c \in C$ and $c \not\in B'$. The
first condition $c \in C$ makes $G \models B' \rightarrow C$ include $G \models
B' \rightarrow c$. The second condition $c \not\in B'$ and this conclusion are
the premises of Lemma~\ref{set-implies-set}, which proves $G^c \models B'
\rightarrow E$ for some clause $E \rightarrow c \in G$.

Since $c \in C$, this clause $E \rightarrow c$ of $G$ resolves with $C \cup D
\rightarrow x$ into
{} $C \backslash \{c\} \cup E \cup D \rightarrow x$.
The entailment $G^c \models B' \rightarrow E$ allows applying the inductive
assumption:
{} $C \backslash \{c\} \cup E \cup D \rightarrow x$
resolves with clauses of $G^c$ into
{} $C \backslash \{c\} \cup B''' \cup D \rightarrow x$,
where $B''' \subseteq B'$.

Overall, $C \cup D \rightarrow x$ resolves with a clause of $G$ into
{} $C \backslash \{c\} \cup E \cup D \rightarrow x$,
which then resolves with clauses of $G^b \subseteq G$ into
{} $C \backslash \{c\} \cup B''' \cup D \rightarrow x$
where $B''' \subseteq B'$. The last clause is similar to the first $C \cup D
\rightarrow x$ in that its head is the same and its body minus $D$ is still
implied by $B'$. Indeed, $G \models B' \rightarrow C \backslash \{c\} \cup
B'''$ follows from $G \models B' \rightarrow C$ and $B''' \subseteq B'$. Yet,
the variables of $C \backslash \{c\} \cup B'''$ not in $B'$ are one less than
those of $C$.

Applying the same mechanism to another variable of $C \backslash \{c\} \cup
B'''$ not in $B'$ if any is therefore possible, and results in another sequence
of resolutions that replace that variable with variables of $B'$. Since the
number of variables not in $B'$ strictly decreases, the process terminates with
a clause containing zero of them. The resulting clause is $B'' \cup D
\rightarrow x$ with $B'' \subseteq B'$.

This proves the induction claim.

\

The premise of the induction claim is $G \models B' \rightarrow C$; its
conclusion is that $C \cup D \rightarrow x$ resolves with clauses of $G$ into
$B'' \cup D \rightarrow x$ with $B'' \subseteq B'$.

The premise is satisfied by $F^x \models B' \rightarrow B$, obtained in the
first paragraph of the proof. The conclusion is that $B \cup \emptyset
\rightarrow x$ resolves with clauses of $F^x$ into $B'' \cup \emptyset
\rightarrow x$ with $B'' \subseteq B'$. Since the first paragraph of the proof
also states $B \rightarrow x \in F$, all parts of the lemma are proved.~\qed

The goal is still to find the clauses of $ITERATION(B,F)$. A previous result
shows that their heads are $HEADS(B,F)$ if the formula is single-head
equivalent. This lemma restricts its bodies. It does not restrict them enough,
however.

For example, since $abc \rightarrow d$ is a clause of
{} $F = \{a \rightarrow b, ac \rightarrow d, abc \rightarrow d\}$,
it is the result of a zero-length resolution starting from itself. Yet, $ac
\rightarrow d$ is obtained in the same way. It implies $abc \rightarrow d$,
making it redundant. If a set of clauses containing $ab \rightarrow d$ fails at
being $ITERATION(B,F)$, the same test with $abc \rightarrow d$ in its place
fails as well.

The inclusion of $abc \rightarrow d$ among the clauses to test is a problem for
two reasons:

\begin{itemize}

\item the bodies generated this way are associated to the heads in $HEADS(B,F)$
to form the candidates for a valid iteration function; the number of these
combinations increases exponentially with the number of bodies;

\item as explained later, a candidate set for $ITERATION(B,F)$ is checked using
the result of this bounded form of resolution; namely, its result on $F$ and
its result on the formula under construction plus the candidate set are
compared; this comparison can be done syntactically, but only if all clauses in
the candidate set are minimal.

\end{itemize}

Retaining only the clauses of minimal bodies solves both problems:

\[
HCLOSE(H,F) = \{B \rightarrow x \mid
	F \models B \rightarrow x ,~
	x \in H \backslash B ,~
	\not\exists B' \subset B ~.~ F \models B' \rightarrow x
\}
\]

This set can be generated by removing all clauses containing others from the
result of the resolution procedure hinged on $H$. The {\tt reconstruct.py}
program does it by an obvious function $minimal(C)$ that returns the clauses of
$C$ whose body does not strictly contain others. The algorithm that generates
$HCLOSE(B,F)$ calls it multiple times to reduce the set of clauses it works on.

\

\begin{algorithm}
\label{hclose-algorithm}

\noindent $hclose(H, F)$

\begin{enumerate}

\item $C = \{B \rightarrow x \in F \mid x \in H \backslash B\}$

\item $C = minimal(C)$
\label{step-hclose-minimize-initial}

\item $R = \emptyset$

\item $T = C \backslash R$

\item while $T \not= \emptyset$:

\begin{enumerate}

\item for $B \rightarrow x \in T$:

\begin{itemize}

\item[] for $B'' \rightarrow y \in F$:

\begin{itemize}

\item[] if $B' \rightarrow x \in resolve(B'' \rightarrow y, B \rightarrow x)$
and $x \not\in B'$:

\begin{itemize}

\item[] $C = C \cup \{B' \rightarrow x\}$

\end{itemize}

\end{itemize}

\end{itemize}

\item $C = minimal(C)$
\label{step-hclose-minimize-loop}

\item $R = R \cup T$

\item $T = C \backslash R$

\end{enumerate}

\item return $C$
\label{step-hclose-return}

\end{enumerate}

\end{algorithm}

\

Since the main loop of this algorithm is a ``while'' statement on a set $T$
that is changed in its iterations, its termination is not obvious. The
following lemma proves it.

\begin{lemma}

The {\tt hclose()} algorithm (Algorithm~\ref{hclose-algorithm}) always
terminates.

\end{lemma}

\proof If a clause is in $T$, it is added to $R$ in the next iteration of the
main loop by the instruction $R = R \cup T$ and removed from $T$ by $T = C
\backslash R$.

The clause remains in $R$ because this set is never removed elements. It is
never inserted again in $T$ since this set is only changed by $T = C \backslash
R$. This proves that $T$ does not contain the same clause in two different
iterations.

The number of clauses deriving from resolution from a fixed formula is finite.
Since all clauses added to $T$ derive from resolution, if the iterations
overcome this finite number, $T$ is empty in at least one of them. This
condition terminates the loop.~\qed

Let $HCLOSEALL(H,F)$ be like $HCLOSE(H,F)$ without minimality:
{} $HCLOSEALL(H,F) = \{B \rightarrow x \mid
{}	F \models B \rightarrow x ,~
{}	x \in H \backslash B \}$.

The minimal subclauses of $HCLOSEALL(H,F)$ equivalently define $HCLOSE(H,F)$.

\begin{lemma}
\label{hcloseall-hclose}

The minimal subclauses of $HCLOSEALL(H,F)$ are $HCLOSE(H,F)$.

\end{lemma}

\proof The definition of $B \rightarrow x \in HCLOSE(H,F)$ is $F \models B
\rightarrow x$, $x \in H \backslash B$ and no subclause $B' \rightarrow x$ of
$B \rightarrow x$ is entailed by $F$. The same clause $B \rightarrow x$ is
minimal in $HCLOSEALL(H,F)$ if $F \models B \rightarrow x$, $x \in H \backslash
B$ and no subclause $B' \rightarrow x$ of $B \rightarrow x$ is in
$HCLOSEALL(H,F)$. The definition $B' \rightarrow x \in HCLOSEALL(H,F)$
comprises $F \models B' \rightarrow x$ but also $x \in H \backslash B'$. The
latter is however not a restriction: if $B' \rightarrow x$ is a subclause of $B
\rightarrow x$, which satisfies $x \in H \backslash B$, it also satisfies $x
\in H \backslash B'$ since $B' \subseteq B$.~\qed

This set $HCLOSEALL(H,F)$ is not interesting by itself. It is only used as an
indirect means to prove the correctness of the algorithm. The first step in
that direction proves that the algorithm only produces clauses of
$HCLOSEALL(H,F)$. Not all of them, though. Yet, for each clause of
$HCLOSEALL(H,F)$ it produces a subclause. Since the only subclause in
$HCLOSEALL(H,F)$ of a minimal clause of $HCLOSEALL(H,F)$ is itself, this proves
that the algorithm generates it. Finally, the last statement $C = minimal(C)$
excludes all clauses that are not minimal.

The first step of the proof proves that Algorithm~\ref{hclose-algorithm} only
returns clauses of $HCLOSEALL(H,F)$.

\begin{lemma}
\label{hcloseall-every}

Every clause produced by the {\tt hclose()} algorithm
(Algorithm~\ref{hclose-algorithm}) is in $HCLOSEALL(H,F)$.

\end{lemma}

\proof The clauses in $HCLOSEALL(H,F)$ are defined by being entailed by $F$, by
having their head in $H$ and not being tautologies. The claim is proved by
showing the all clauses in $C$ satisfy these conditions throughout the
execution of the algorithm.

The set $C$ is initialized by
{} $C = \{B \rightarrow x \in F \mid x \in H \backslash B\}$,
and $x \in H \backslash B$ implies $x \not\in B$. It is then changed only by
{} $C = C \cup \{B' \rightarrow x\}$,
which is only executed if $x \not\in B'$. This proves that $C$ never contains
tautologies.

The other two properties are proved by showing that they are initially
satisfied and are never later falsified: every clause in $C$ is entailed by $F$
and its head is in $H$.

The set $C$ is initialized by
{} $C = \{B \rightarrow x \in F \mid x \in H \backslash B\}$.
All clauses in $F$ are entailed by $F$, and $x \in H \backslash B$ includes $x
\in H$.

A clause $B' \rightarrow x$ is added to $C$ only if it is the result of
resolving a clause $B \rightarrow x$ of $C$ with a clause $B'' \rightarrow y$
of $F$. Since all clauses of $C$ are entailed by $F$, so is $B' \rightarrow x$.
Since all clauses of $C$ have head in $H$, the head $x$ of $B' \rightarrow x$
is in $H$.

This proves that every clause in $C$ is in $HCLOSEALL(H,F)$. Since the
algorithm returns $C$, it only returns clauses of $HCLOSEALL(H,F)$.~\qed

The algorithm returns clauses of $HCLOSEALL(H,F)$, but not all of them. As
expected, since $C$ is minimized. Yet, a subclause of every clause of
$HCLOSEALL(H,F)$ is returned.

\begin{lemma}
\label{hcloseall-subclause}

A subclause of every clause in $HCLOSEALL(H,F)$ is returned by the {\tt
hclose()} algorithm (Algorithm~\ref{hclose-algorithm}).

\end{lemma}

\proof The definition of $B' \rightarrow x \in HCLOSEALL(H,F)$ is $F \models B'
\rightarrow x$ and $x \in H \backslash B'$. The latter implies $x \not\in B'$.
Lemma~\ref{head-resolution} proves that a subclause of $B' \rightarrow x$ is
obtained by iteratively resolving a clause $B \rightarrow x \in F$ such that $x
\not\in B$ with clauses of $F^x$.

The algorithm initializes $C$ with the non-tautological clauses of $F$ with
head in $H$. This includes $B \rightarrow x$ since $x$ is in $H$ but not in
$B$. The algorithm then iteratively adds to $C$ the resolvent of clauses of $C$
with clauses of $F$, but also removes clauses by $C = minimize(C)$.

For each clause in the derivation, one of its subclauses is proved to be in $C$
in the last step of the algorithm.

This property is proved iteratively: for the first clause of the derivation,
and then for every resolvent.

The first clause $B \rightarrow x$ of the derivation is shown to satisfy this
property. Being a clause of $F$ such that $x \in H \backslash B$, it is added
to $C$ by
{} $C = \{B \rightarrow x \mid
{}	F \models B \rightarrow x ,~ x \in H \backslash B\}$.
The only instruction removing clauses from $C$ is $C = minimal(C)$, which only
removes clauses that strictly contain other clauses of $C$. In turn, these
contained clauses are removed only if $C$ contains strict subclauses of them.
By induction on the size of the subclause, $C$ contains a subclause of $B
\rightarrow x$ at the end of the execution of the algorithm.

An arbitrary clause $B'' \rightarrow x$ of the derivation is inductively
assumed to satisfy the property: one of its subclauses $B''' \rightarrow x$ is
in $C$ at the end of execution. The claim is proved on the following clause in
the derivation.

Let $D \rightarrow b$ with $b \in B''$ be the clause of $F^x$ that resolves
with $B'' \rightarrow x$. Their resolvent is
{} $B'' \backslash \{b\} \cup D \rightarrow x$.
Since $B''' \rightarrow x$ is a subclause of $B'' \rightarrow x$, its body
$B'''$ is a subset of $B''$. It may contain $b$ or not.

If $B'''$ does not contain $b$ then $B''' \subseteq B'' \backslash \{b\}$, and
$B''' \rightarrow x$ is itself a subclause of
{} $B'' \backslash \{b\} \cup D \rightarrow x$,
and is by assumption in $C$ at the end of execution.

If $B'''$ contains $b$ then $B''' \rightarrow x$ resolves with $D \rightarrow
b$, producing
{} $B''' \backslash \{b\} \cup D \rightarrow x$.
This is a subclause of
{} $B'' \backslash \{b\} \cup D \rightarrow x$
since $B''' \subseteq B''$. A subclause of this resolvent is proved to be in
$C$ at the end of execution.

Since $B''' \rightarrow x$ is in $C$, it has been added by either
{} $C = \{B \rightarrow x \mid F
{}	\models B \rightarrow x ,~ x \in H \backslash B\}$
or
{} $C = C \cup \{B' \rightarrow x\}$.
In the first case, $B''' \rightarrow x$ is also copied to $T$. In the second,
since $R$ accumulates the values of $T$, either $T$ already contained $B'''
\rightarrow x$ at some previous point or contains it after $T = C \backslash
R$. Either way, at some point $B''' \rightarrow x$ is in $T$. At the next
iteration, it is resolved with all clauses of $F$, including $D \rightarrow b$.
This clause does not contain $x$ since it is in $F^x$. Neither does $B'''$
since $B''' \rightarrow x$ is in $C$, which is never added a tautology. Their
resolvent
{} $B''' \backslash \{b\} \cup D \rightarrow x$
is therefore not a tautology and is therefore added to $C$. It is removed if
and when $C$ contains one of its proper subclauses. Which is only removed if
and when $C$ contains one of its proper subclauses. This inductively proves
that a subclause of
{} $B''' \backslash \{b\} \cup D \rightarrow x$,
which is a subclause of
{} $B'' \backslash \{b\} \cup D \rightarrow x$,
is in $C$ at the end of execution.

This shows that a subclause of every clause in the derivation is in the final
value of $C$. This includes the final clause of the derivation: $B' \rightarrow
x$. Since $B' \rightarrow x$ is an arbitrary clause of $HCLOSEALL(H,F)$, this
proves that a subclause of every clause of $HCLOSEALL(H,F)$ is produced by the
algorithm.~\qed

The rest of the proof of correctness of Algorithm~\ref{hclose-algorithm} is
easy.

\begin{lemma}
\label{hclose-every}

The {\tt hclose()} algorithm (Algorithm~\ref{hclose-algorithm}) generates
exactly the minimal subclauses of $HCLOSEALL(H,F)$.

\end{lemma}

\proof For every clause $B' \rightarrow x$ of $HCLOSEALL(H,F)$, the algorithm
generates one of its subclauses by Lemma~\ref{hcloseall-subclause}. Since it is
generated by the algorithm, this subclause $B'' \rightarrow x$ is in
$HCLOSEALL(H,F)$ by Lemma~\ref{hcloseall-every}. Since $B' \rightarrow x$ is
minimal, none of its proper subsets is in $HCLOSEALL(H,F)$. Therefore, the
subclause $B'' \rightarrow x$ of $B' \rightarrow x$ is not a proper subclause:
it coincides with $B' \rightarrow x$. This proves that the algorithm generates
every minimal clause of $HCLOSEALL(H,F)$.

In the other direction, since the algorithm returns all minimal clauses of
$HCLOSEALL(H,F)$, it cannot return any non-minimal one. Indeed, if $B'
\rightarrow x \in HCLOSEALL(H,F)$ contains another clause of $HCLOSEALL(H,F)$,
it also contains a minimal clause of $HCLOSEALL(H,F)$. By
Lemma~\ref{hcloseall-every}, this minimal clause is in $C$ at the end of the
algorithm. Therefore, $B' \rightarrow x$ is removed from $C$ by $C =
minimal(C)$ and is not returned.~\qed

Since the minimal subclauses of $HCLOSEALL(H,F)$ are exactly $HCLOSE(H,F)$ by
Lemma~\ref{hcloseall-hclose}, the correctness of the algorithm is proved.

\begin{corollary}

The {\tt hclose()} algorithm (Algorithm~\ref{hclose-algorithm})
returns $HCLOSE(H,F)$.

\end{corollary}

Having shown how $HCLOSE(H,F)$ can be calculated, it can be used to determine
$ITERATION(B,F)$, which is then used within the reconstruction algorithm to
determine a single-head formula equivalent to a given one if any. How it is
used is the theme of the next section.

\subsection{Body search}

When searching for the value of $ITERATION(B,F)$, Lemma~\ref{iteration-heads}
provides half of the answer: its heads are exactly $HEADS(B,F)$. What about its
bodies? The definition of a valid iteration function only constraints them to
come from $BCL(B,F)$. It is not much of a constraint: being a set of
consequences, $BCL(B,F)$ is not bounded by the size of $F$. Not all of it is
necessary, though: if a valid iteration function exists, one such that
$ITERATION(B,F)$ is contained in $HCLOSE(HEADS(B,F),UCL(B,F))$ exists as well.
Proving this statement is the aim of this section.

The first two lemmas are proved with an arbitrary subset of $SCL(B,F) \cup
BCL(B,F)$, which $UCL(B,F)$ is then proved to be.

\begin{lemma}
\label{hclose-scl-bcl}

For every $H$ and every formula $F' \subseteq SCL(B,F) \cup BCL(B,F)$,
it holds $HCLOSE(H,F') \subseteq SCL(B,F) \cup BCL(B,F)$.

\end{lemma}

\proof The condition $B' \rightarrow x \in SCL(B,F) \cup BCL(B,F)$ is the same
as $B' \rightarrow x \in SCL(B,F)$ or $B' \rightarrow x \in BCL(B,F)$. This is
an alternative. The first case is $x \not\in B'$, $F \models B' \rightarrow x$,
$F \models B \rightarrow B'$ and $F \not\models B' \rightarrow B$. The second
is $x \not\in B'$, $F \models B' \rightarrow x$, $F \models B \rightarrow B'$
and $F \models B' \rightarrow B$. The two alternative subconditions $F
\not\models B' \rightarrow B$ and $F \models B' \rightarrow B$ cancel each
other. This proves $B' \rightarrow x \in SCL(B,F) \cup BCL(B,F)$ equivalent to
$x \not\in B'$, $F \models B' \rightarrow x$ and $F \models B \rightarrow B'$.

The premise of the lemma $F' \subseteq SCL(B,F) \cup BCL(B,F)$ is the same as
$x \not\in B'$, $F \models B' \rightarrow x$ and $F \models B \rightarrow B'$
for every clause $B' \rightarrow x$ of $F'$. The claim is the same for every
clause $B' \rightarrow x$ of $HCLOSE(H,F')$.

The definition of $B' \rightarrow x \in HCLOSE(H,F')$ includes $x \in H
\backslash B'$, which implies $x \not\in B'$, the first requirement for $B'
\rightarrow x \in SCL(B,F) \cup BCL(B,F)$. It also includes $F' \models B'
\rightarrow x$. Since $F'$ is contained in $SCL(B,F) \cup BCL(H,F)$, which only
contains clauses entailed by $F$, it is entailed by $F$. A consequence of $F
\models F'$ and $F' \models B' \rightarrow x$ is $F \models B' \rightarrow x$.
This is the second requirement for $B' \rightarrow x \in SCL(B,F) \cup
BCL(B,F)$.

Only $F \models B \rightarrow B'$ remains to be proved. Since $F$ entails $F'$,
which contains $B' \rightarrow x$, it also entails $B' \rightarrow x$. A
subclause of $B' \rightarrow x$ is obtained by resolving clauses of $F$. The
claim $F \models B \rightarrow B'$ is iteratively proved on all clauses in such
a derivation. The base case is proved because $F \models F'$ and $F' \subseteq
SCL(B,F) \cup BCL(B,F)$ imply $F \models B \rightarrow B'$ for every clause $B'
\rightarrow x \in F'$.

The induction step assumes $F \models B \rightarrow B''$ and $F \models B
\rightarrow B'''$ for two clauses $B'' \rightarrow x$ and $B''' \rightarrow y$
obtained from resolution. They resolve if either $x \in B'''$ or $y \in B''$.
The result is respectively $((B'' \backslash \{x\}) \cup B''') \rightarrow y$
or $(B'' \cup (B''' \backslash \{y\})) \rightarrow x$. In both cases the body
of the resulting clause is a subset of $B'' \cup B'''$. The claim is proved
because of $F \models B \rightarrow B''$ and $F \models B \rightarrow
B'''$.~\qed

A specific value for $H$ strengthens the containment proved by the lemma.

\begin{lemma}
\label{hclose-subset-bcl}

For every formula $F' \subseteq SCL(B,F) \cup BCL(B,F)$,
it holds $HCLOSE(HEADS(B,F),F') \subseteq BCL(B,F)$.

\end{lemma}

\proof Since $F' \subseteq SCL(B,F) \cup BCL(B,F)$, Lemma~\ref{hclose-scl-bcl}
proves $HCLOSE(H,F') \subseteq SCL(B,F) \cup BCL(B,F)$ for every set of
variables $H$. The claim is proved by showing that $H = HEADS(B,F)$ forbids
clauses of $SCL(B,F)$ in $HCLOSE(H,F')$.

By contradiction, $B' \rightarrow x$ is assumed to be a clause of $SCL(B,F)$
that is also in $HCLOSE(HEADS(B,F),F')$. The definition of
$HCLOSE(HEADS(B,F),F')$ includes $x \in HEADS(B,F)$. Since $HEADS(B,F) =
RCN(B,F) \cap SFREE(B,F)$, it holds $x \in SFREE(B,F)$. This is defined as the
set of variables that are the head of no clause of $SCL(B,F)$, contradicting
the assumption $B' \rightarrow x \in SCL(B,F)$.~\qed

A set contained in $SCL(B,F) \cup BCL(B,F)$ is $UCL(B,F)$, as proved by
the following lemma.

\begin{lemma}
\label{ucl-f-scl-bcl}

For every formula $F$ and set of variables $B$,
it holds $UCL(B,F) = F \cap (SCL(B,F) \cup BCL(B,F))$.

\end{lemma}

\proof The set $UCL(B,F)$ is defined as the set of clauses $B' \rightarrow x$
of $F$ such that $x \not\in B$ and $F \models B \rightarrow B'$. The entailment
defines $B' \leq_F B$, which is the same as either $B' <_F B$ or $B' \equiv_F
B$. Since every clause of $F$ is entailed by $F$, every clause $B' \rightarrow
x$ of $UCL(B,F)$ satisfies $x \not\in B'$, $F \models B' \rightarrow x$ and
either $B' <_F B$ or $B' \equiv_F B$. Therefore, it is either in $SCL(B,F)$ or
in $BCL(B,F)$.

In the reverse direction, every clause of $SCL(B,F)$ and every clause of
$BCL(B,F)$ satisfies $x \not\in B'$ and $F \models B \rightarrow B'$. This is
the definition of $UCL(B,F)$ if the clause is also in $F$.~\qed

Lemma~\ref{hclose-subset-bcl} proves $HCLOSE(HEADS(B,F),F') \subseteq BCL(B,F)$
for every subset $F'$ of $SCL(B,F) \cup BCL(B,F)$. The last lemma proves that
$UCL(B,F)$ is such a subset. Combining the two lemmas produce the following
corollary.

\begin{corollary}
\label{hclose-bcl}

For every formula $F$ and set of variables $B$, it holds
$HCLOSE(HEADS(B,F),UCL(B,F)) \subseteq BCL(B,F)$.

\end{corollary}

This corollary proves that $HCLOSE(HEADS(B,F),UCL(B,F))$ is a subset of
$BCL(B,F)$ and being a subset of $BCL(B,F)$ is a part of the definition of
$ITERATION(B,F)$. The other parts are now proved to be met by
$HCLOSE(HEADS(B,F),UCL(B,F))$ as well.

This claim requires some preliminary lemmas. The first is the monotonicity of
$UCL(B,F)$ with respect to containment of the formula.

\begin{lemma}
\label{ucl-subformula}

If $F' \subseteq F$ then $UCL(B,F') \subseteq UCL(B,F)$.

\end{lemma}

\proof The definition of $B' \rightarrow x \in UCL(B,F')$ is $B' \rightarrow x
\in F'$, $x \not\in B'$ and $F' \models B \rightarrow B'$. The first condition
implies $B' \rightarrow x \in F$ because $F' \subseteq F$. The third implies $F
\models B \rightarrow B'$ by monotonicity of entailment. The conditions in the
definition of $B' \rightarrow x \in UCL(B,F)$ all hold.~\qed

This lemma allows proving that $UCL(B,F)$ is the part of $F$ that matters when
checking entailment from $B$.

\begin{lemma}
\label{ucl-body}

The following two conditions are equivalent:
$F \models B \rightarrow x$
and
$UCL(B,F) \models B \rightarrow x$.

\end{lemma}

\proof If $x \in B$ then $B \rightarrow x$ is a tautology. It is therefore
entailed by both $F$ and $UCL(B,F)$ and the claim is proved. The rest of the
proof covers the case $x \not\in B$.

If $UCL(B,F)$ implies $B \rightarrow x$ also $F$ does, since $UCL(B,F)$
is a subset of $F$.

Remains to prove the converse. The assumption is $F \models B \rightarrow x$
and the claim is $UCL(B,F) \models B \rightarrow x$.

This is proved by induction on the size of $F$.

The base case is $F = \emptyset$, where $F \models B \rightarrow x$ implies $x
\in B$, a case already proved to satisfy the claim.

The induction step proves that $F \models B \rightarrow x$ implies $UCL(B,F)
\models B \rightarrow x$ assuming the same implication for every $F'$ strictly
smaller than $F$.

The claim is already proved when $x \in B$. Otherwise, $x \not\in B$ and $F
\models B \rightarrow x$ imply by Lemma~\ref{set-implies-set} that $F$ contains
a clause $B' \rightarrow x$ such that $F^x \models B \rightarrow B'$. If $x \in
B'$ then $B' \rightarrow x$ is a tautology and is therefore entailed by
$UCL(B,F)$. Otherwise, $x \not\in B'$. The condition $B' \rightarrow x \in F$
implies $F \models B' \rightarrow x$. The condition $F^x \models B \rightarrow
B'$ implies $F \models B \rightarrow B'$. Otherwise, all conditions for $B'
\rightarrow x$ being in $UCL(B,F)$ are met. Therefore, $UCL(B,F)$ entails
$B' \rightarrow x$.

By definition, $F^x \models B \rightarrow B'$ is the same as $F^x \models B
\rightarrow b$ for every $b \in B'$. Since $F^x$ is smaller than $F$ because it
does not contain $B' \rightarrow x$ at least, the induction assumption applies:
$UCL(B,F^x) \models B \rightarrow b$. Since $F^x \subseteq F$,
Lemma~\ref{ucl-subformula} tells that $UCL(B,F^x) \subseteq UCL(B,F)$. By
monotonicy of entailment, $UCL(B,F) \models B \rightarrow b$ follows. This holds
for every $b \in B'$, implying $UCL(B,F) \models B \rightarrow B'$.

This conclusion $UCL(B,F) \models B \rightarrow B'$ with the previous one
$UCL(B,F) \models B' \rightarrow x$ implies $UCL(B,F) \models B \rightarrow
x$.~\qed

Since $F \models B \rightarrow x$ defines $x \in BCN(B,F)$ and $UCL(B,F)
\models B \rightarrow x$ defines $x \in BCN(B,UCL(B,F))$, an immediate
consequence is the equality of these two sets.

\begin{lemma}
\label{ucl-bcn}

For every formula $F$ and set of variables $B$ it holds
$BCN(B,F) = BCN(B,UCL(B,F))$.

\end{lemma}

\proof By Lemma~\ref{ucl-body}, $F \models B \rightarrow x$ is the same as
$UCL(B,F) \models B \rightarrow x$. The first condition defines $x \in
BCN(B,F)$, the second $x \in BCN(B,UCL(B,F))$.~\qed

Contrary to the other functions defined in this article, $UCL(B,F)$ only
contains clauses of $F$. A clause that is not in $F$ is never in $UCL(B,F)$
even if it is entailed by it. This set is the restriction of $F$ to the case
where $B$ is true. Only the clauses in $UCL(B,F)$ matter when deriving
something from $F \cup B$. They are the only clauses that matter to $BCN(B,F)$,
$RCN(B,F)$, $SCL(B,F)$ and $BCL(B,F)$. If $UCL(B,F)$ replaces $F$, these do not
change. More generally, if something is defined in terms of what $F$ and $B$
entail, it is unaffected by the removal of all clauses not in $UCL(B,F)$ from
$F$.

This claim requires a preliminary result: $UCL()$ is monotonic with respect to
entailment of the set of variables.

\begin{lemma}
\label{ucl-contain}

If $F \models A \rightarrow B$ then $UCL(B,F) \subseteq UCL(A,F)$.

\end{lemma}

\proof By definition, $B' \rightarrow x \in UCL(B,F)$ is $x \not\in B'$, $F
\models B \rightarrow B'$ and $B' \rightarrow x \in F$. The second condition $F
\models B \rightarrow B'$ implies $F \models A \rightarrow B'$ since $F \models
A \rightarrow B$ is assumed. With $x \not\in B'$ and $B' \rightarrow x \in F$,
this condition defines $B' \rightarrow x \in UCL(A,F)$.~\qed

Lemma~\ref{ucl-body} generalizes from clauses whose body is equal to the given
set of variables to clauses whose body is entailed by it.

\begin{lemma}
\label{ucl-only}

If $F \models A \rightarrow B$, then $F \models B \rightarrow x$ is equivalent
to $UCL(A,F) \models B \rightarrow x$.

\end{lemma}

\proof A consequence of $UCL(A,F) \models B \rightarrow x$ is $F \models B
\rightarrow x$ since $UCL(A,F)$ is a subset of $F$. The first direction of the
claim is proved even if $F \models A \rightarrow B$ does not hold.

The other direction is that $F \models B \rightarrow x$ implies $UCL(A,F)
\models B \rightarrow x$ if $F \models A \rightarrow B$. By
Lemma~\ref{ucl-body}, the assumption $F \models B \rightarrow x$ implies
$UCL(B,F) \models B \rightarrow x$. This is not the claim yet because it
contains $UCL(B,F)$ instead of $UCL(A,F)$. However, Lemma~\ref{ucl-contain}
proves $UCL(B,F) \subseteq UCL(A,F)$ from $F \models A \rightarrow B$.
Monotonicity implies $UCL(A,F) \models B \rightarrow x$.~\qed

This lemma proves that checking whether $F$ entails $B \rightarrow x$ where $F
\models A \rightarrow B$ can be safely restricted from $F$ to its subset
$UCL(A,F)$. This set is determined by the algorithm for $RCN(A,F)$ at almost no
additional cost, and may reduce the number of clauses to consider. This
reduction is dramatic on large formulae that contain only a small fraction of
clauses whose bodies are entailed by $A$.

This result applies to the search for bodies of $ITERATION(B,F)$.

\begin{lemma}
\label{hclose-it-subset}

If $ITERATION(B,F)$ is a valid iteration function and $B' \rightarrow x \in
ITERATION(B,F)$, then $HCLOSE(HEADS(B,F),UCL(B,F))$ contains a clause $B''
\rightarrow x$ with $B'' \subseteq B'$.

\end{lemma}

\proof The definition of $B' \rightarrow x \in ITERATION(B,F)$ includes $F
\models B' \rightarrow x$ and $F \models B \equiv B'$. Equivalence implies
entailment: $F \models B \rightarrow B'$. Lemma~\ref{ucl-only} applies: $F
\models B' \rightarrow x$ implies $UCL(B,F) \models B' \rightarrow x$.

Lemma~\ref{iteration-heads} proves $x \in HEADS(B,F)$.

By definition, $HCLOSE(HEADS(B,F),UCL(B,F))$ does not include a clause $B'
\rightarrow x$ such that $UCL(B,F) \models B' \rightarrow x$ and $x \in
HEADS(B,F)$ only if $UCL(B,F) \models B'' \rightarrow x$ with $B'' \subset B'$.
Since the head of this clause is in $HEADS(B,F)$, the same applies to it: it is
not in $HCLOSE(HEADS(B,F),UCL(B,F))$ if a proper subclause of it has the same
property. This inductively proves that a subclause $B'' \rightarrow x$ of $B'
\rightarrow x$ is in $HCLOSE(HEADS(B,F),UCL(B,F))$.~\qed

Corollary~\ref{hclose-bcl} prove that searching for a valid $ITERATION(B,F)$
can be restricted to the subsets of $HCLOSE(HEADS(B,F),UCL(B,F))$.

\begin{lemma}
\label{hclose-it}

If $ITERATION(B,F)$ is a valid iteration function then another valid iteration
function with the same heads in the same number of clauses is a subset of
$HCLOSE(HEADS(B,F),UCL(B,F))$.

\end{lemma}

\proof Lemma~\ref{hclose-it-subset} shows that $B' \rightarrow x \in
ITERATION(B,F)$ implies $B'' \rightarrow x \in HCLOSE(HEADS(B,F),UCL(B,F))$ for
some $B'' \subseteq B'$. This containment implies $B'' \rightarrow x \models B'
\rightarrow x$. As a result, replacing $B' \rightarrow x$ with $B'' \rightarrow
x$ in $ITERATION(B,F)$ results in a formula entailing $ITERATION(B,F)$, and
therefore still entailing $BCL(B,F)$. This is the entailment part of the
definition of a valid iteration function.

This replacement also maintains the choice part. By Corollary~\ref{hclose-bcl},
$HCLOSE(HEADS(B,F),UCL(B,F)) \subseteq BCL(B,F)$. This proves $x \not\in B'$,
$F \models B' \rightarrow x$ and $F \models B \equiv B'$. Remains to prove $x
\in SFREE(B,F)$. It holds because the definition of $B' \rightarrow x \in
HCLOSE(HEADS(B,F),UCL(B,F))$ includes $x \in HEADS(B,F)$, which includes $x \in
SFREE(B,F)$, which implies $B' \rightarrow x \not\in SCL(B,F)$.

This replacement maintains the heads of the clauses because it replaces a
clause with another having the same head.~\qed

This lemma proves that searching for $ITERATION(B,F)$ can be restricted to the
subsets of $HCLOSE(HEADS(B,F),UCL(B,F))$. This avoids the inclusion of clauses
containing others; a further minimization is shown in the next section.

\subsection{Implied by strictly implied subsets}

% the procedure switches from B' to B'' if the first and SCL(B,F) entail the
% second, not the other way around; this is correct because SCL(B,F)uB' |= B''
% implies the reverse implication on clauses: SCL(B,F)u{B''->x} |= B'->x; in
% such cases, B''->x allows entailing everything B'->x does

Lemma~\ref{hclose-it} bounds the search for a valid iteration function to the
subsets of $HCLOSE(HEADS(B,F),UCL(B,F))$. Still, the number of such subsets may
be exponential. Lemma~\ref{iteration-heads} further bounds the search: the
heads of $ITERATION(B,F)$ are not just in $HEADS(B,F)$, they are exactly
$HEADS(B,F)$. The subsets that do not contain a head in $HEADS(B,F)$ can be
skipped. So can the subsets that contain duplicated heads, since the aim is to
find a single-head formula. Yet, these reductions do not always make the number
of candidate subsets polynomial. Every further restriction helps.

Some considerations on $HCLOSE(HEADS(B,F),UCL(B,F))$ provide a direction. It
constrains clauses to be entailed by $UCL(B,F)$, to have head in $HEADS(B,F)$
and a minimal body with respect to set containment. The first two are necessary
because of the definition of validity and because of
Lemma~\ref{iteration-heads}, but why minimality?

A non-minimal body is never necessary because $B'' \rightarrow x$ entails $B'
\rightarrow x$ if $B'' \subset B'$. Everything $B' \rightarrow x$ entails is
also entailed by $B'' \rightarrow x$. Both clauses are acceptable for being in
$ITERATION(B,F)$, they both consume the same head $x$, but $B'' \rightarrow x$
entails everything $B' \rightarrow x$ entails. The latter can be disregarded
since the former is a valid substitute of it in every respect.

The goal is a set of clauses that entails $BCL(B,F)$ with $SCL(B,F)$. If $B''
\subset B'$, then
{} $SCL(B,F) \cup IT \cup \{B'' \rightarrow x\}$
implies
{} $SCL(B,F) \cup IT \cup \{B' \rightarrow x\}$.
What really matters is this implication. When it holds, $BCL(B,F)$ is entailed
by the implicant if it is entailed by the implicate. The implication is a
consequence of set containment, but is not equivalent to it. It may hold even
if set containment does not.

The implication involves the set $IT$, but is used to restrict the possible
clauses to include in $IT$, when $IT$ is not fixed yet. What can be checked
then is whether
{} $SCL(B,F) \cup \{B'' \rightarrow x\}$
implies
{} $SCL(B,F) \cup \{B' \rightarrow x\}$.
This is the same as
{} $SCL(B,F) \cup \{B'' \rightarrow x\} \models B' \rightarrow x$,
which holds if
{} $SCL(B,F) \cup B' \models B''$.
A body $B'$ can be replaced by another $B''$ it entails with $SCL(B,F)$.

\begin{lemma}
\label{minbodies}

If $ITERATION(B,F)$ is a valid iteration function,
{} $B' \rightarrow x \in ITERATION(B,F)$,
{} $B'' \rightarrow x \in BCL(B,F)$ and
{} $SCL(B,F) \cup B' \models B''$ hold,
another valid iteration function is
{} $ITERATION'(B,F) = ITERATION(B,F)
{}   \backslash \{B' \rightarrow x\} \cup \{B'' \rightarrow x\}$.

\end{lemma}

\proof All clauses in $ITERATION'(B,F)$ but $B'' \rightarrow x$ are in
$ITERATION(B,F)$. They satisfy the condition of choice of the definition of
validity, which is now proved for $B'' \rightarrow x$. The assumption $B''
\rightarrow x \in BCL(B,F)$ implies $F \models B'' \rightarrow x$ and $F
\models B \equiv B''$. The condition $x \in SFREE(B,F)$ follows from $B'
\rightarrow x \in ITERATION(B,F)$. This proves the condition of choice.

Remains to prove the entailment condition: $SCL(B,F) \cup ITERATION'(B,F)
\models BLC(B,F)$. By assumption, $SCL(B,F) \cup B' \models B''$. As a result,
{} $SCL(B,F) \cup ITERATION'(B,F) \cup B'$,
being equal to
{} $SCL(B,F) \cup ITERATION(B,F)
{}	\backslash \{B' \rightarrow x\}
{}	\cup \{B'' \rightarrow x\}
{}	\cup B'$,
entails
{} $SCL(B,F) \cup ITERATION(B,F)
{}	\backslash \{B' \rightarrow x\}
{}	\cup \{B'' \rightarrow x\}
{}	\cup B''$,
which entails
{} $x$.
By the deduction theorem,
{} $SCL(B,F) \cup ITERATION'(B,F) \models B' \rightarrow x$.
Since $SCL(B,F) \cup ITERATION'(B,F)$ implies the only clause of
$ITERATION(B,F)$ it does not contain, it implies it all:
{} $SCL(B,F) \cup ITERATION'(B,F) \models SCL(B,F) \cup ITERATION(B,F)$.
Since $ITERATION(B,F)$ is a valid iteration function, it satisfies
{} $SCL(B,F) \cup ITERATION(B,F) \models BCL(B,F)$.
By transitivity,
{} $SCL(B,F) \cup ITERATION'(B,F) \models BCL(B,F)$.~\qed

If $B' \rightarrow x$ and $B'' \rightarrow x$ are both in
$HCLOSE(HEADS(B,F),UCL(B,F))$, the first can be disregarded if $SCL(B,F) \cup
B' \models B''$. This reduces the set of candidates subsets of
$HCLOSE(HEADS(B,F),UCL(B,F))$.

This is similar to what Lemma~\ref{hclose-it} allows when $B'' \subset B'$, but
with an important difference: the reverse containment $B' \subset B''$ is not
possible, the reverse implication from $B''$ to $B'$ is. If $SCL(B,F) \models
B' \equiv B''$, Lemma~\ref{minbodies} allows $B'' \rightarrow x$ to replace $B'
\rightarrow x$ but also the other way around. Such an equivalence is not
possible under set containment: $B' \subset B''$ contradicts $B'' \subset B'$.
If a body is minimal according to set containment, no other body can take its
place; a body that is minimal according to implication can be replaced by an
equivalent one. While uniquely defining $HCLOSE(H,F)$ is possible, extending it
to minimality by entailment is not. Not uniquely. This is why the following
definition only gives a condition of minimality rather than uniquely defining a
set like $HCLOSE(H,F)$.

\begin{definition}

$MINBODIES(U,S)$ is an arbitrary function that returns a subset $R$ of $U$ such
that $B' \rightarrow x \in U$ implies $S \cup B' \models B''$ for some clause
$B'' \rightarrow x \in R$.

\end{definition}

The function $minbodies(U,S)$ in {\tt singlehead.py} meets this definition. It
works on whole clauses $B' \rightarrow x$ and $B'' \rightarrow x$ instead of
their bodies $B'$ and $B''$ only. It starts from a clause of $U$ and repeatedly
checks whether its body and $S$ entail the body of another clause of $U$ with
the same head. The trace of clauses in this and in all runs avoid loops and
repeated search.

The function is called with
{} $U=HCLOSE(HEADS(B,F),UCL(B,F))$
and
{} $S=UCL(B,F) \cap SCL(B,F)$.
The resulting subset of $HCLOSE(HEADS(B,F),UCL(B,F))$ may still include some
clauses that could be removed for two reasons.

\begin{description}

\item[How inference is checked.]

Lemma~\ref{minbodies} allows neglecting a clause if its body and $SCL(B,F)$
entails another body. But the function $minbodies(U,S)$ in {\tt singlehead.py}
does not check $SCL(B,F) \cup B' \models B''$ for every set $B''$. For
simplicity, it only checks whether a single variable in $B'$ resolves with a
single clause of $UCL(B,F) \cap SCL(B,F)$ to produce $B''$.

Using resolution instead of entailment is not a limitation because the only
difference is that resolution may produce a subclause of an entailed clause.
This difference disappears since $B' \rightarrow x$ is subset-minimal: none of
its subsets are entailed.

Whether resolving a variable in $B'$ with a clause in $UCL(B,F) \cap SCL(B,F)$
once is sufficient is another story. A preliminary analysis suggests that this
is not a limitation, and a resolution derivation of $B''$ from $SCL(B,F) \cup
B'$ can always be restructured so that the last step is always the resolution
of a body of $HCLOSE(HEADS(B,F),UCL(B,F))$ with a clause of $UCL(B,F) \cap
SCL(B,F)$. This would prove by iteration that such a derivation is a sequence
of steps like the ones the program follows. Yet, this is unproven.

\item[How minimal elements are searched for.]

The function $minbodies(U,S)$ in {\tt singlehead.py} follows implications among
bodies, avoiding the ones it already visited. The following example ({\tt
minbodies.py}) shows that it may return a non-minimal body this way.

\[
F = \{
	bh \rightarrow c, ch \rightarrow b, ch \rightarrow d,
	x \rightarrow h,
	bde \rightarrow x
\}
\]

The first four clauses make $UCL(B,F) \cap SCL(B,F)$ when $B = \{b, d, e\}$.
They imply
{} $bhe \rightarrow che$,
{} $che \rightarrow bhe$ and
{} $che \rightarrow cde$.
These three bodies $bhe$, $che$ and $cde$ are equivalent according to $F$, but
$cde$ is only entailed by the other two according to $UCL(B,F) \cap SCL(B,F)$.

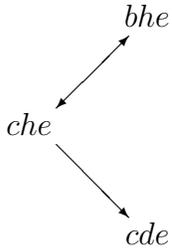
\setlength{\unitlength}{5000sp}%
\begingroup\makeatletter\ifx\SetFigFont\undefined%
\gdef\SetFigFont#1#2#3#4#5{%
  \reset@font\fontsize{#1}{#2pt}%
  \fontfamily{#3}\fontseries{#4}\fontshape{#5}%
  \selectfont}%
\fi\endgroup%
\begin{picture}(615,1245)(5656,-4486)
\thinlines
{\color[rgb]{0,0,0}\put(5806,-3796){\vector(-1,-1){  0}}
\put(5806,-3796){\vector( 1, 1){360}}
}%
{\color[rgb]{0,0,0}\put(5806,-3976){\vector( 1,-1){360}}
}%
\put(5671,-3931){\makebox(0,0)[b]{\smash{{\SetFigFont{12}{24.0}
{\rmdefault}{\mddefault}{\updefault}{\color[rgb]{0,0,0}$che$}%
}}}}
\put(6256,-3391){\makebox(0,0)[b]{\smash{{\SetFigFont{12}{24.0}
{\rmdefault}{\mddefault}{\updefault}{\color[rgb]{0,0,0}$bhe$}%
}}}}
\put(6256,-4471){\makebox(0,0)[b]{\smash{{\SetFigFont{12}{24.0}
{\rmdefault}{\mddefault}{\updefault}{\color[rgb]{0,0,0}$cde$}%
}}}}
\end{picture}%
\nop{
    <---> bhe
che
    ----> cde
}

The only necessary clause is $cde \rightarrow x$, since its body $cde$ is
implied by the other two $che$ and $bhe$ but not the other way around. Yet,
{\tt minbodies(U, S)} may also produce $bhe \rightarrow x$.

If searching starts from $che$, it moves to $bhe$ and stops there because the
only other body it directly entails it is $che$, which it skips as already
analyzed. The clause $bhe \rightarrow x$ is therefore produced even if its body
entails $cde$ and not the other way around.

The problem is still due to single resolutions in place of inference, but not
only. The only necessary body $cde$ could be found from $bhe$ even using a
single resolution at time by not avoiding the bodies already visited like
$che$. This is not done for its computational costs.

\end{description}

These issues may make the function return some clauses that are not minimal.
This only affects the running time of the algorithm, not its correctness. It
increases the number of candidates for $ITERATION(B,F)$ and ultimately the
overall running time, but since all necessary clauses are included the result
is correct anyway.

\subsection{The algorithm, at last}

The overall algorithm for checking whether a formula is single-head equivalent
assembles the pieces built so far. It is the reconstruction algorithm
(Algorithm~\ref{reconstruction}) where $ITERATION(B,F)$ is not given but
calculated employing $HEADS()$, $HCLOSE()$ and $MINBODIES()$.

The reconstruction algorithm iterates over the bodies of the clauses in the
formula, sorted by entailment. It starts from the bodies that entail no other
and continues to the ones entailing only the ones already processed. For every
body $B$, it calculates $ITERATION(B,F)$ and adds it to the formula under
construction $G$. The only missing bit of this algorithm is $ITERATION(B,F)$.
It is found by iterating over the sets of clauses it may contain until a set
that entails $BCL(B,F)$ with $SCL(B,F)$ is found.

\

\begin{algorithm}[Singlehead-autoreconstruction algorithm]
\label{algorithm-autoreconstruction}

\begin{enumerate}

\item $F = \{B \rightarrow x \in F \mid x \not \in B\}$

\item $G = \emptyset$

\item $P = \{B \mid \exists x . B \rightarrow x \in F\}$

\item while $P \not= \emptyset$
\label{autoreconstruction-external}

\begin{enumerate}

\item choose a $<_F$-minimal $B \in P$
\newline
(a set such that $A <_F B$ does not hold for any $A \in P$)
\label{autoreconstruction-choice}

\item $H = HEADS(B,F)$

\item $M = HCLOSE(H,UCL(B,F))$

\item $T = MINBODIES(M,UCL(B,F) \cap SCL(B,F))$

\item $E = \false$

\item for all $IT =
\{C_1 \rightarrow h_1, \ldots, C_m \rightarrow h_m \mid
H = \{h_1,\ldots,h_m\} ,~
\forall i ~.~ h_i \not\in C_i ,~
\exists y_1,\ldots,y_m ~.~
C_1 \rightarrow y_1,\ldots,C_m \rightarrow y_m \in T\}$
\label{autoreconstruction-internal}

\begin{enumerate}

\item if $SCL(B,F) \cup IT \models BCL(B,F)$

\begin{enumerate}

\item $E = \true$

\item break
\label{autoreconstruction-found}

\end{enumerate}

\end{enumerate}

\item if $\neg E$

\begin{enumerate}

\item fail

\end{enumerate}

\item $G = G \cup IT$

\item $P = P \backslash \{B' \mid F \models B \equiv B'\}$

\end{enumerate}

\item return $G$

\end{enumerate}

\end{algorithm}

\

The overall structure is the same as the reconstruction algorithm
(Algorithm~\ref{algorithm-precondition-reconstruction}): a single-head formula
$G$ aims at replicating $F$ by accumulating $ITERATION(B,F)$ during a loop over
the bodies $B$ of $F$ sorted by $\leq_F$. The difference is that
$ITERATION(B,F)$ is not given but found by iterating over sets of clauses $IT$.

This search is the main contributor to complexity in the algorithm, since
exponentially many such sets may exists. This is why a number of hacks for
efficiency are implemented:

\begin{itemize}

\item instead of checking all subsets of $BCL(B,F)$, only the subsets of
$HCLOSE(HEADS(B,F),UCL(B,F))$ are considered;

\item this set is further reduced by $MINBODIES()$;

\item rather than looping over sets of clauses, each variable in $HEADS(B,F)$
is attached a body coming from $MINBODIES()$; this is correct because all these
bodies are equivalent; it is convenient because it automatically excludes all
sets of clauses with missing or duplicated heads;

\item $RCN(B,F)$ and $UCL(B,F)$ are determined and stored for all bodies of $F$
once at the beginning.

\end{itemize}

The algorithm is further improved by other optimizations described in separate
sections because of the length of their proofs. The rest of this section is
about the correctness of Algorithm~\ref{algorithm-autoreconstruction} itself.

\begin{lemma}
\label{it-bcl-each}

All non-tautological clauses with head in $HEADS(B,F)$ and body in
{} $HCLOSE(HEADS(B,F),UCL(B,F))$
are in $BCL(B,F)$.

\end{lemma}

\proof The premise of the lemma is that $B' \rightarrow x$ is a clause such
that $x \not\in B'$, $x \in HEADS(B,F)$ and $B' \rightarrow y \in
HCLOSE(HEADS(B,F),UCL(B,F))$ for some variable $y$.

Corollary~\ref{hclose-bcl} proves that every clause of
$HCLOSE(HEADS(B,F),UCL(B,F))$ is in $BCL(B,F)$. The definition of $B'
\rightarrow y \in BCL(B,F)$ includes $B' \equiv_F B$, which is also part of the
definition of $B' \rightarrow x \in BCL(B,F)$. Another part is $x \not\in B'$,
which holds because $B' \rightarrow x$ is by assumption not a tautology.

The only missing part is $F \models B' \rightarrow x$. The definition of $x \in
HEADS(B,F)$ includes $x \in RCN(B,F)$, which implies $x \in BCN(B,F)$ by
Lemma~\ref{b-rcn}: $F \models B \rightarrow x$. Since the previously proved
fact $B' \equiv_F B$ is defined as $F \models B' \equiv B$, this entailment is
the same as $F \models B' \rightarrow x$, the missing part of the definition of
$B' \rightarrow x \in BCL(B,F)$.~\qed

Since $IT$ comprises only clauses meeting the condition of the previous lemma,
they satisfy its consequence.

\begin{lemma}
\label{it-bcl}

All sets $IT$ in the autoreconstruction algorithm
(Algorithm~\ref{algorithm-autoreconstruction}) are contained in
$BCL(B,F)$.

\end{lemma}

\proof By construction, $IT$ is made of non-tautological clauses with heads in
$HEADS(B,F)$ and bodies in $MINBODIES(HCLOSE(HEADS(B,F),UCL(B,F)),UCL(B,F) \cap
SCL(B,F))$. The latter is a subset of its first argument
$HCLOSE(HEADS(B,F),UCL(B,F))$ by definition. Therefore, all clauses of $IT$
have heads in $HEADS(B,F)$ and bodies in $HCLOSE(HEADS(B,F),UCL(B,F))$ and are
not tautologies. Lemma~\ref{it-bcl-each} prove that they are in
$BCL(B,F)$.~\qed

Since all sets $IT$ are subsets of $BCL(B,F)$, its clauses meet most of the
choice part of the definition of a valid iteration function: $x \not\in B'$, $F
\models B' \rightarrow x$ and $F \models B \equiv B'$. They also meet the
remaining one, $x \in SFREE(B,F)$, since $x \in HEADS(B,F)$.

The entailment part of the definition of validity is not meet in general, but
only if the iteration of the external loop ends.

\begin{lemma}
\label{auto-iteration}

At the end of every iteration of the external loop
(Step~\ref{autoreconstruction-external}) of the autoreconstruction algorithm
(Algorithm~\ref{algorithm-autoreconstruction}), $IT$ satisfies the
conditions for a valid iteration function $ITERATION(B,F)$.

\end{lemma}

\proof Lemma~\ref{it-bcl} proves that $IT$ only contains clauses of $BCL(B,F)$.
By construction, the heads of $IT$ are in $HEADS(B,F)$, a subset of
$SFREE(B,F)$. The choice condition for $IT$ being a valid iteration function is
met.

The entailment condition is $SCL(B,F) \cup IT \models BCL(B,F)$. This may or
may not happen for a specific set $IT$. However, if no set $IT$ satisfies
$SCL(B,F) \cup IT \models BCL(B,F)$ in the internal loop, the algorithm fails.
Therefore, the entailment condition for $IT$ holds if the iteration of the
external loop ends.~\qed

While $IT$ is a valid iteration function if the algorithm does not fail, the
same could be achieved by making the algorithm fail for all sets $IT$. It is
also essential that the algorithm does not fail if a valid iteration function
exists.

\begin{lemma}
\label{auto-heads}

If $F$ has a valid iteration function such that $ITERATION(B,F)$ is
single-head, then $IT$ is equal to the value of a valid iteration function
$ITERATION(B,F)$ at some iteration of the internal loop
(Step~\ref{autoreconstruction-internal}) of the autoreconstruction algorithm
(Algorithm~\ref{algorithm-autoreconstruction}).

\end{lemma}

\proof Lemma~\ref{hclose-it} proves that if $F$ has a valid iteration function
it also has one such that $ITERATION(B,F)$ is a subset of
$HCLOSE(HEADS(B,F),UCL(B,F))$ and has the same heads in the same number of
clauses. Lemma~\ref{minbodies} further restricts this property to the subsets
of $MINBODIES(HCLOSE(HEADS(B,F),UCL(B,F)),UCL(B,F) \cap SCL(B,F))$. Since this
is a valid iteration function, its heads are $HEADS(B,F)$ by
Lemma~\ref{iteration-heads}.

This proves that if $F$ has a valid iteration function such that
$ITERATION(B,F)$ is single-head, also has one such that $ITERATION(B,F)$ have
bodies from $MINBODIES(HCLOSE(HEADS(B,F),UCL(B,F)),UCL(B,F) \cap SCL(B,F))$ and
single-head heads in $HEADS(B,F)$. This other one is still a valid iteration
function; therefore, it contains no tautology. Since the internal loop
(Step~\ref{autoreconstruction-internal}) of the algorithm tries all
tautology-free associations of these heads with these bodies, it finds it.~\qed

The following lemma has an awkward statement but is necessary to avoid
repeating the same argument in multiple proofs.

\begin{lemma}
\label{same-choices}

If the autoreconstruction algorithm
(Algorithm~\ref{algorithm-autoreconstruction}) terminates, the
sequence of values of $G$ is the same as in the reconstruction algorithm
(Algorithm~\ref{algorithm-reconstruction}) for some nondeterministic choices
and the following valid iteration function.

\[
ITERATION(B,F) =
\left\{
\begin{array}{ll}
IT
&
\mbox{at the end of the iteration of $B'$}
\\
&
\mbox{if $B$ is equivalent to a body $B'$ in $F$}
\\
\emptyset
&
\mbox{otherwise}
\end{array}
\right.
\]

\end{lemma}

\proof The iteration function is valid: Lemma~\ref{auto-iteration} proves the
conditions of validity when $B$ is a body in $F$, and these conditions carry
over to equivalent sets because they are semantical;
Lemma~\ref{iteration-preconditions} applied in reverse proves that the value of
every valid iteration function $ITERATION(B,F)$ is empty if $B$ is not
equivalent to a body in $F$.

The allowed choices for the precondition $B$ in
Algorithm~\ref{algorithm-autoreconstruction} are the same as in
Algorithm~\ref{algorithm-precondition-reconstruction}.
Lemma~\ref{auto-iteration} proves that $IT$ is the value of some valid
iteration function $ITERATION(B,F)$ at the end of every iteration of
Algorithm~\ref{algorithm-autoreconstruction}. Since $G$ is initially empty and
is added respectively $IT$ or $ITERATION(B,F)$ at every step, the values of $G$
are the same in the two algorithms.

Lemma~\ref{precondition-or-empty} proves that
Algorithm~\ref{algorithm-reconstruction} can always choose a set $B$ such that
$ITERATION(B,F) = \emptyset$ if $P$ does not contain a $<_F$-minimal body of
$F$. Therefore, Algorithm~\ref{algorithm-reconstruction} can always choose a
set $B$ so that it does not change $G$ when it cannot choose a set $B$ like
Algorithm~\ref{algorithm-precondition-reconstruction} does. The sequence of
values of $G$ is the same because the choices of the first kind do not change
it and the choices of the second kind change it in the same way.~\qed

The correctness of the algorithm can now be proved.

\begin{theorem}
\label{autoreconstruction-correct}

The autoreconstruction algorithm
(Algorithm~\ref{algorithm-autoreconstruction}), returns a single-head formula
equivalent to the input formula if and only if the input formula has any.

\end{theorem}

\proof The proof comprises three steps: first, if $F$ is single-head
equivalent, the algorithm does not fail; second, the output formula (if any) is
equivalent to $F$; third, if $F$ is single-head equivalent the output formula
is single-head. These three facts imply that the algorithm outputs a
single-head version of $F$ if any. The second implies that if $F$ is not
single-head equivalent the output formula is not single-head, if the algorithm
does not fail.

The first step is that the algorithm does not fail if $F$ is single-head
equivalent. If $F$ is single-head equivalent, it has a valid iteration function
whose values are all single-head by Lemma~\ref{equivalent-valid}. This is the
precondition of Lemma~\ref{auto-heads}, which proves that $IT$ holds the value
of a valid iteration function at some iteration of the internal loop of the
algorithm. The definition of validity includes $SCL(B,F) \cup IT \models
BCL(B,F)$. This entailment makes $E$ true, avoiding failure.

The second step is that the output formula is equivalent to the input formula.
Lemma~\ref{same-choices} proves that the values of $G$ in the algorithm are the
same as the values in Algorithm~\ref{algorithm-reconstruction} for a valid
iteration function and some nondeterministic choices. This includes the last
value, which is returned. Algorithm~\ref{algorithm-reconstruction} returns a
formula equivalent to $F$ by Lemma~\ref{reconstruction-equivalent}.

The third step is that the output formula (if any) is single-head if the input
formula is single-head equivalent. The algorithm has been proved to terminate
two paragraphs above; this is the precondition of Lemma~\ref{same-choices},
which proves the validity of a certain iteration function. Its possible values
are $\emptyset$ and the value of $IT$ at the end of some iteration of the
algorithm. Both are single-head: $\emptyset$ because it contains no clause,
$IT$ by construction. Since the output formula is the same as that of
Algorithm~\ref{algorithm-reconstruction} with this iteration function by
Lemma~\ref{same-choices}, and this iteration function has only single-head
values, Lemma~\ref{singlehead-local} applies: the output is single-head since
$F$ is by assumption equivalent to a single-head formula.~\qed

If the formula $F$ is single-head equivalent the algorithm terminates and
outputs a single-head formula equivalent to it. Otherwise, the algorithm does
not output a single-head formula; it may output a formula that is not
single-head or nothing at all because it fails.

\subsection{Tautologies}

The autoreconstruction algorithm (Algorithm~\ref{algorithm-autoreconstruction})
exploits $RCN(B,F)$ and $UCL(B,F)$. A previous article shows an algorithm {\tt
rcnucl()} for calculating $RCN(B,F)$ and
{} $\{B' \rightarrow x \in F \mid F \models B \rightarrow B'\}$.
The second is almost $UCL(B,F)$ but does not exclude tautologies. Since $F$ is
removed all tautologies in the first step of the autoreconstruction algorithm,
this missing condition is moot as implied by $B' \rightarrow x \in F$.

Some optimizations below exploit $UCL(B,G \cup IT)$. The {\tt rcnucl()}
algorithm determines it if $G \cup IT$ does not contain tautologies, which is
proved by the following lemma.

\begin{lemma}
\label{g-it-taut}

The set $G \cup IT$ does not contain tautologies during the execution of the
autoreconstruction algorithm (Algorithm~\ref{algorithm-autoreconstruction}).

\end{lemma}

\proof The sets assigned to $IT$ do not contain tautology because their
definition includes $h_i \not\in C_i$. Since $G$ is initially empty, is only
changed by the instruction $G = G \cup IT$ and $IT$ does not contain tautology,
it does not either.~\qed

% proof by induction: $G$ initially does not contain any tautology; at each
% step of the algorithm, it is inductively assumed not to contain any tautology
% at the start of every iteration; it is changed by setting it to $G \cup IT$;
% this set does not contain tautology because $G$ does not by the induction
% assumption and $IT$ by construction

\subsection{Wild-goose chase}

The algorithm requires $SCL(B,F)$ and $BCL(B,F)$. Being sets of consequences of
$F$, they may be exponentially large. Their direct use is best avoided. The
implementation of the algorithm in {\tt singlehead.py} exploits three variants
to this aim:

\begin{enumerate}

\item in place of $HEADS(B,F)$, whose definition involves the set of heads of
$SCL(B,F)$, it uses the difference between $RCN(B,F)$ and the heads of $G$, the
formula under construction;

\item in place of $UCL(B,F) \cap SCL(B,F)$ it uses $UCL(B,F) \cap U$ where $U$
is the union of $UCL(B,F)$ for all bodies $B$ of $F$ used in the previous
iterations of the algorithm;

\item $SCL(B,F) \cup IT \models BCL(B,F)$ is checked as
{} $HCLOSE(RCN(B,G \cup IT),UCL(B,G \cup IT)) = HCLOSE(RCN(B,F),UCL(B,F))$.

\end{enumerate}

In all three optimizations $G$ takes the place of $SCL(B,F)$. This is correct
as long as the algorithm mimics the reconstruction algorithm with a valid
iteration function.

\begin{lemma}
\label{g-scl}

At every iteration of the autoreconstruction algorithm
(Algorithm~\ref{algorithm-autoreconstruction}),
$G \models SCL(B,F)$.

\end{lemma}

\proof By Lemma~\ref{same-choices}, the values of $G$ in the autoreconstruction
algorithm are the same as in Algorithm~\ref{algorithm-reconstruction} for some
nondeterministic choices and a valid iteration function.
Lemma~\ref{reconstruction-invariant} proves $G \models SCL(B,F)$.~\qed

The algorithm works by iteratively rebuilding the formula from the ground
up---from a body $A$ that do not entail any other to a body $B$ that only
entails $A$ and so on. Each floor is built over the lower ones. If they are
not, it cannot. They have to be finished and solid. In logical terms,
processing $B$ requires $A$ to have already been processed, and its
consequences already being entailed.

The first lemma states that the order of processing of bodies is correct.

\begin{lemma}
\label{less-before}

If $B$ is chosen in a given iteration of the autoreconstruction algorithm
(Algorithm~\ref{algorithm-autoreconstruction})
and $A'$ is the body of a clause
in $F$ such that $A' <_F B$, then a set $A$ such that $A \equiv_F A'$ was
chosen in a previous iteration.

\end{lemma}

\proof The algorithm only chooses sets of variables that are minimal in $P$.
This means that when $B$ is chosen, the set $A'$ is not in $P$. Since $A'$ is
the body of a clause in $F$, it was initially in $P$. As a result, $A'$ has
been removed in a previous iteration. Let $A$ be the set of variables chosen in
that iteration. Since $P$ is only removed elements by the statement $P = P
\backslash \{B' \mid B \cup F \models B \equiv B'\}$, it holds $F \models A
\equiv A'$, which is the same as $A \equiv_F A'$.~\qed

If the algorithm is erecting a building, it needs all floors below $B$ before
building $B$. Floor $B$ is not made in mid-air, it is made over them. The next
lemma proves the converse: no floor that should be over $B$ is built before it.

\begin{lemma}
\label{before-not-greater}

If $A$ is chosen in an iteration of the autoreconstruction algorithm
(Algorithm~\ref{algorithm-autoreconstruction}) that precedes the iteration in
which $B$ is chosen, then $B \not\leq_F A$.

\end{lemma}

\proof Since $B$ is chosen after $A$, it is in $P$ when $A$ is chosen and at
the end of that iteration. Since the algorithm only chooses minimal elements of
$P$, it does not choose $A$ if $B <_F A$. Since $B$ is still in $P$ at the end
of the iteration, after $P = P \backslash \{B' \mid F \models B \equiv B'\}$,
then $A \equiv_F B$ does not hold either. Since $B$ is not strictly less than
$A$ and not equivalent to it, it is not less than or equal to it: $B \not\leq_F
A$.~\qed

\

Adding $IT$ to $G$ makes $G$ entail every clause in $BCL(B,F)$ at the end of
each iteration. As expected and wanted. What would be neither expected nor
wanted is going overboard and adding clauses that are not in $BCL(B,F)$, or
even worse not entailed by $F$. After all, the final aim is to reconstruct $F$
in a controlled way. Controlled means single-head, but equivalence with $F$ is
essential: not producing anything not entailed by $F$.

This unwanted situation is first excluded for $G$.

\begin{lemma}
\label{g-subset-bcl}

At every iteration of the autoreconstruction algorithm
(Algorithm~\ref{algorithm-autoreconstruction}) $G$ is a subset of $\cup
\{BCL(A,F) \mid B \not\leq_F A\}$.

\end{lemma}

\proof Let $A' \rightarrow x$ a clause of $G$. The set $G$ is initially empty,
and it is only changed by $G = G \cup IT$ at the end of every iteration of the
external loop. Therefore, $A' \rightarrow x$ is in $IT$ at the end of an
iteration. Let $A$ be the element of $P$ chosen in
Step~\ref{autoreconstruction-choice} at the beginning of that iteration.
Lemma~\ref{before-not-greater} proves $B \not\leq_F A$. Lemma~\ref{it-bcl}
proves $IT \subseteq BCL(A,F)$.~\qed

Since $G$ is not too large and $IT$ is built appropriately as shown by
Lemma~\ref{it-bcl}, their union $G \cup IT$ is also correct.

\begin{lemma}
\label{f-implies-g-it}

At every iteration of the algorithm,
$F \models G \cup IT$ holds.

\end{lemma}

\proof Lemma~\ref{it-bcl} proves that $IT$ is a subset of $BCL(B,F)$, which is
a set of consequences of $F$. Lemma~\ref{g-subset-bcl} proves that $G$ is
contained in the union of some sets $BCL(A,F)$, which again only contain
consequences of $F$. As a result, all clauses of $G \cup IT$ are entailed by
$F$.~\qed

\subsection{Head back}

The autoreconstruction algorithm uses $HEADS(B,F) = RCN(B,F) \cap SFREE(B,F)$
as the heads of the clauses of $IT$. Since $SFREE(B,F)$ is defined in terms of
$SCL(B,F)$, which is a set of consequences of $F$, it is expensive to
determine. The implemented algorithm replaces it with $RCN(B,F)$ minus the
heads of the clauses in $G$.

This is not always correct. Formula $F = \{a \rightarrow x, b \rightarrow x\}$
is a counterexample. The set $B$ chosen in the first iteration may be $\{a\}$,
and the algorithm sets $G = \{a \rightarrow x\}$. The second set $B$ is then
$\{b\}$. By definition, $SFREE(B,F)$ is the set of all variables that are not
heads of $SCL(B,F)$, the clauses entailed by $F$ whose body are less than $B$;
the only variable $B=\{b\}$ entails is $x$, and no clause in $F$ contains $x$
negative; therefore, $SCL(B,F) = \emptyset$ and $SFREE(B,F) = \{a,b,x\}$. While
{} $HEADS(B,F) = RCN(B,F) \cap SFREE(B,F) = \{x\} \cap \{a,b,x\} = \{x\}$,
the heads of $IT$ in the implemented algorithm are
{} $RCN(B,F) \backslash \{x \mid \exists B . B \rightarrow x \in G\} =
{}  \emptyset$.

Granted, the counterexample formula is not single-head equivalent. And not by
chance: the actual and computed heads only differ on formulae that are not
single-head equivalent. Rather than insisting on the correct set of heads,
efforts are better spent on proving that the final outcome of each iteration is
correct: even when the algorithm uses the wrong sets of heads, it still
disproves the formula single-head equivalent. This is exactly what happens for
the counterexample formula: the set of heads for $IT$ is empty while $BCL(B,F)$
contains $b \rightarrow x$. The iteration ends in failure, as required. This
holds in general, not just in the counterexample.

The first step in this direction is that the heads of $RCN(B,F)$ minus the
heads of $G$ are all in $HEADS(B,F)$.

\begin{lemma}
\label{h-hclose}

During the execution of the reconstruction algorithm
(Algorithm~\ref{algorithm-autoreconstruction})
{} $H = RCN(B,F) \backslash \{x \mid \exists B' . B' \rightarrow x \in G\}$
is a subset of $HEADS(B,F)$.

\end{lemma}

\proof The claim is proved by showing that every variable not in $HEADS(B,F)$
is not in $H$ either.

If $x$ is not in $HEADS(B,F)$ then it is either not in $RCN(B,F)$ or not in
$SFREE(B,F)$. In the first case, $x$ is not in $H$ because $H$ is a subset of
$RCN(B,F)$. In the second case, $x$ is proved not to be in $H$ by showing that
it is the head of a clause of $G$.

The definition of $x \in SFREE(B,F)$ is that a clause $A' \rightarrow x$ is in
$SCL(B,F)$. Being in $SCL(B,F)$ it is not a tautology: $x \not\in A'$. By
Lemma~\ref{g-scl}, $G$ entails all of $SCL(B,F)$, including $A' \rightarrow x$.
This entailment and $x \not\in A'$ are the preconditions of
Lemma~\ref{set-implies-set}, which tells that $G$ contains a clause $A''
\rightarrow x$. Since $x$ is the head of a clause of $G$, it is not in
$H$.~\qed

This lemma proves that $H$ is a subset of $HEADS(B,F)$. An expected consequence
is that the sets $IT$ built over $H$ are also sets that can be built over
$HEADS(B,F)$. This is however not the case because $MINBODIES(U,S)$ is not
necessarily monotonic with respect to its first argument. Yet, the change would
preserve the property proved by Lemma~\ref{f-implies-g-it}: $F$ entails these
sets.

\begin{lemma}
\label{f-implies-g-it-h}

At every step of the autoreconstruction algorithm
(Algorithm~\ref{algorithm-autoreconstruction}), $F$ implies every clause with
head in
{} $H = RCN(B,F) \backslash \{x \mid \exists B' . B' \rightarrow x \in G\}$
and body in
{} $MINBODIES(HCLOSE(H,UCL(B,F)),UCL(B,F) \cap SCL(B,F))$.

\end{lemma}

\proof By Lemma~\ref{h-hclose}, $H$ is a subset of $HEADS(B,F)$. By definition,
{} $MINBODIES(HCLOSE(H,UCL(B,F)),UCL(B,F) \cap SCL(B,F))$
is a subset of its first argument $HCLOSE(H,UCL(B,F))$. The definition of this
set includes $H$ only as $x \in H \backslash B$, which implies $x \in
HEADS(B,F) \backslash B$ since $H \subseteq HEADS(B,F)$: every clause of
$HCLOSE(H,UCL(B,F))$ is in $HCLOSE(HEADS(B,F),UCL(B,F))$. This implies that
{} $MINBODIES(HCLOSE(H,UCL(B,F)),UCL(B,F) \cap SCL(B,F))$
is a subset of
{} $HCLOSE(HEADS(B,F),UCL(B,F))$.
Lemma~\ref{it-bcl-each} proves that all non-tautological clauses with heads in
$HEADS(B,F)$ and bodies in $HCLOSE(HEADS(B,F),UCL(B,F))$ are contained in
$BLC(B,F)$ and are therefore entailed by $F$; the tautological ones are
entailed because they are tautologies.~\qed

These two lemmas allow proving the main property of this section: $RCN(B,F)$
minus the heads of $G$ is a correct replacement of $HEADS(B,F)$ in the
autoreconstruction algorithm. More precisely, if they are not the same the
autoreconstruction algorithm gives the correct answer anyway.

\begin{lemma}
\label{heads-h}

If
{} $H = RCN(B,F) \backslash \{x \mid \exists B' . B' \rightarrow x \in G\}$
is not equal to $HEADS(B,F)$, the formula $F$ is not single-head equivalent and
the autoreconstruction algorithm (Algorithm~\ref{algorithm-autoreconstruction})
running with this modified definition of $H$ fails.

\end{lemma}

\proof The modified and the original algorithms are identical as long as
{} $H = RCN(B,F) \backslash \{x \mid \exists B' . B' \rightarrow x \in G\}$
and $HEADS(B,F)$ coincide. Up to the first moment these sets differ, the two
algorithms are the same. Therefore, they possess the same properties. For
example, $F \models G$ as proved in Lemma~\ref{f-implies-g-it}. At the first
iteration where $H$ and $HEADS(B,F)$ differ, the modified algorithm is proved
to fail, but this is correct because the formula is not single-head equivalent.
The rest of the proof is about this first iteration where $H$ and $HEADS(B,F)$
differ.

The starting point is $H \subseteq HCLOSE(H,F)$ as proved by
Lemma~\ref{h-hclose}. The proof comprises four steps that follow from the
assumption that $H$ and $HEADS(B,F)$ differ:

\begin{enumerate}

\item some variables $x$ of $HEADS(B,F)$ are the head of some clauses of $G$

\item if $x$ is a variable of $HEADS(B,F)$ but not of $H$, and $A' \rightarrow
x \in G$ for some $A'$ then both
{} $F \not\models A' \rightarrow B$ and
{} $F \not\models B \rightarrow A'$
hold;

\item the algorithm fails;

\item $F$ is not equivalent to any single-head formula.

\end{enumerate}

\

The first step proves that if $HEADS(B,F)$ and $H$ differ then some variables
of $HEADS(B,F)$ are the head of some clauses of $G$.

Since $H$ is subset of $HEADS(B,F)$, these two sets can only differ if
$HEADS(B,F)$ contains a variable not in $H$. Being in $HEADS(B,F)$, such a
variable is in $RCN(B,F)$. Since it is not $H$, which is $RCN(B,F)$ minus the
heads of $G$, it is the head of a clause of $G$.

\

The second step proves that if $x$ is a variable of $HEADS(B,F) \backslash H$,
then all clauses $A' \rightarrow x \in G$ satisfy
{} $F \not\models A' \rightarrow B$ and
{} $F \not\models B \rightarrow A'$.

By Lemma~\ref{g-subset-bcl}, $G$ is a subset of $\cup \{BCL(A,F) \mid B
\not\leq_F A\}$. Therefore, $A' \rightarrow x$ is in $BCL(A,F)$ for some $A$
such that $B \not\leq_F A$. By definition of $BCL(A,F)$ the equivalence $A
\equiv_F A'$ holds. With $B \not\leq_F A$, it proves $B \not\leq_F A'$. This is
the first property to prove: $F \not\models A' \rightarrow B$.

Since $x$ is in $HEADS(B,F)$, it is in $SFREE(B,F)$. By definition, no clause
$A' \rightarrow x$ with head $x$ is in $SCL(B,F)$ exists. The definition of $A'
\rightarrow x \in SCL(B,F)$ is $x \not\in A'$, $F \models A' \rightarrow x$ and
$A' <_F B$. If the first two conditions are true, the third is false. Since $A'
\rightarrow x$ is in $G$, it is in $\cup \{BCL(A,F) \mid B \not\leq_F A\}$ by
Lemma~\ref{g-subset-bcl}; since $BCL(A,F)$ does not include tautologies by
definition, $x$ is not in $A'$. Since $A' \rightarrow x \in G$ and $F \models
G$ by Lemma~\ref{f-implies-g-it}, $F \models A' \rightarrow x$. As a result,
$A' <_F B$ is false. Since $A' <_F B$ means $A' \leq_F B$ and $B \not\leq_F
A'$, its negation is either $A' \not\leq_F B$ or $B \leq_F A'$. The second
cannot because $B \not\leq_F A'$, as proved above. Therefore, $A' \not\leq B$.
This implies $F \not\models B \rightarrow A'$, the second property to prove.

\

The third step proves that if $HEADS(B,F)$ and $H$ differ the algorithm fails
if it uses $H$ in place of $HEADS(B,F)$.

As proved in the first step, if $HEADS(B,F)$ and $H$ differ then a variable $x$
in $HEADS(B,F)$ is the head of a clause of $G$. Since $HEADS(B,F)$ is a subset
of $RCN(B,F)$, this variable $x$ is in $RCN(B,F)$. By Lemma~\ref{rcn-bcl}, a
clause $B' \rightarrow x$ is in $BCL(B,F)$.

By contradiction, the algorithm is assumed not to fail. Since the algorithm
fails if $E$ is false, and $E$ is set true only when $SCL(B,F) \cup IT \models
BCL(B,F)$ holds, this entailment is the case for some $IT$. Since $G$ entails
$SCL(B,F)$ by Lemma~\ref{g-scl}, this implies $G \cup IT \models B' \rightarrow
x$. The other premise $x \not\in B'$ of Lemma~\ref{set-implies-set} holds since
$B' \rightarrow x$ is in $BCL(B,F)$, which does not contain tautologies by
definition. The lemma proves the existence of a clause $B'' \rightarrow x \in G
\cup IT$ such that $G \cup IT \models B' \rightarrow B''$. The former can be
rewritten as: either $B'' \rightarrow x \in G$ or $B'' \rightarrow x \in IT$.
The alternative $B'' \rightarrow x \in IT$ is false since the heads of $IT$ are
exactly $H$, and $H$ does not contain any head of $G$ by construction. As a
result, $B'' \rightarrow x \in G$. As proved in the previous step, for all
clauses $B'' \rightarrow x \in G$ it holds $F \not\models B \rightarrow B''$.
Since $B' \rightarrow x \in BCL(B,F)$ includes $F \models B \equiv B'$, this is
the same as $F \not\models B' \rightarrow B''$.
Since $F$ entails $G$ by
Lemma~\ref{f-implies-g-it} and all clauses of $IT$ by
Lemma~\ref{f-implies-g-it-h}, 
the previously proved property
$G \cup IT \models B' \rightarrow B''$ implies $F \models B' \rightarrow B''$.

This is a contradiction. The assumption was that the algorithm does not fail.
Therefore, the algorithm fails.

\

This failure would be a problem if the formula were single-head equivalent, but
this is proved not to be the case. This is the fourth and final step of the
proof.

By assumption, $HEADS(B,F)$ and $H$ differ. The first step of the proof proves
$x \in HEADS(B,F)$ and $A' \rightarrow x \in G$ for a clause $A' \rightarrow
x$. The second step proves $F \not\models B \rightarrow A''$ and $F \not\models
A'' \rightarrow B$ for all clauses $A'' \rightarrow x \in G$ with head in
$HEADS(B,F)$, including $A' \rightarrow x$. Since $G$ does not contain
tautologies by Lemma~\ref{g-it-taut}, $x$ is not in $A'$. Since $F$ entails $G$
by Lemma~\ref{f-implies-g-it}, it entails its clause $A' \rightarrow x$.

Since $x$ is in $HEADS(B,F)$, it is also in $RCN(B,F)$ by definition. By
Lemma~\ref{rcn-bcl}, $BCL(B,F)$ contains a clause $B' \rightarrow x$. This is
defined as $x \not\in B'$, $F \models B' \rightarrow x$ and $F \models B \equiv
B'$.

The proof is by contradiction: $F'$ is assumed to be a single-head formula
equivalent to $F$. Because of equivalence, everything entailed by $F$ is
entailed by $F'$. In particular, $F'$ entails $B' \rightarrow x$ and $A'
\rightarrow x$. Since $x$ is neither in $B'$ nor in $A'$,
Lemma~\ref{set-implies-set} applies: $F$ contains two clauses $B'' \rightarrow
x$ and $A'' \rightarrow x$ such that $F' \models B' \rightarrow B''$ and $F'
\models A' \rightarrow A''$. Since $F$ is removed tautologies in the first step
of the algorithm, these are not tautologies: $x \not\in A''$ and $x \not\in
B''$. Since $F'$ is single-head, $A''$ and $B''$ are the same. Let $C = A'' =
B''$. What holds for $A''$ or $B''$ also holds for $C$. Namely, $x \not\in C$,
$C \rightarrow x \in F'$, $F' \models B' \rightarrow C$ and $F' \models A'
\rightarrow C$. Membership implies entailment: $F' \models C \rightarrow x$. By
equivalence, $F$ entails the same formulae:
{} $C \rightarrow x$,
{} $B' \rightarrow C$ and
{} $A' \rightarrow C$.

If $F \models C \rightarrow B$, since $F \models A' \rightarrow C$ and $F
\models B \equiv B'$, then $F \models A' \rightarrow B$, which is not the case.
As a result, $F \not\models C \rightarrow B$. Since $F \models B \equiv B'$,
the entailment $F \models B' \rightarrow C$ implies $F \models B \rightarrow
C$. All conditions for $C \rightarrow x \in SCL(B,F)$ are met: $x \not\in C$,
$F \models C \rightarrow x$, $F \models B \rightarrow C$ and $F \not\models C
\rightarrow B$.

Since $G$ implies $SCL(B,F)$ by Lemma~\ref{g-scl}, it implies $C \rightarrow
x$. Since $x$ is not in $C$, Lemma~\ref{set-implies-set} proves that $G$
contains a clause $C' \rightarrow x$ such that $G \models C \rightarrow C'$,
which implies $F \models C \rightarrow C'$ because $F$ implies $G$ by
Lemma~\ref{f-implies-g-it}. Since $F \models B' \rightarrow C$ and $F \models B
\equiv B'$, this implies $F \models B \rightarrow C'$. This is a contradiction
since $x \in HEADS(B,F)$ and $C' \rightarrow x \in G$ imply $F \not\models B
\rightarrow C'$ as proved in the second step of the proof.~\qed

In summary, even if $H$ differs from $HEADS(B,F)$ the algorithm is still right:
it fails, which is what it should do since the formula is not single-head
equivalent. This allows using $H$ instead of $HEADS(B,F)$, saving the
calculation of $SCL(B,F)$.

A positive side effect of the change is that success is no longer ambiguous. If
$F$ is not single-head equivalent, the original algorithm may fail or return a
formula that is not single-head. The modified version fails, period.

\begin{lemma}
\label{always-fails}

If $F$ is not equivalent to any single-head formula,
Algorithm~\ref{algorithm-autoreconstruction} with the variant
{} $H = RCN(B,F) \backslash \{x \mid \exists B' . B' \rightarrow x \in G\}$
fails.

\end{lemma}

\proof A difference between
{} $RCN(B,F) \backslash \{x \mid \exists B' . B' \rightarrow x \in G\}$
and $HEADS(B,F)$ makes the modified algorithm fail, as shown by
Lemma~\ref{heads-h}. The claim is therefore proved in this case. Remains to
prove it when
{} $RCN(B,F) \backslash \{x \mid \exists B' . B' \rightarrow x \in G\}$
always coincides with $HEADS(B,F)$.

This is proved by contradiction: the algorithm is assumed not to fail and
{} $RCN(B,F) \backslash \{x \mid \exists B' . B' \rightarrow x \in G\}$
to always coincide with $HEADS(B,F)$; the input formula is proved single-head
equivalent.

Since
{} $RCN(B,F) \backslash \{x \mid \exists B' . B' \rightarrow x \in G\}$
is the same as $HEADS(B,F)$, the modified algorithm is the same as the original
(Algorithm~\ref{algorithm-autoreconstruction}). Since the modified algorithm
does not fail, the original does not fail either. By
Theorem~\ref{autoreconstruction-correct}, the output formula of the original
algorithm is equivalent to $F$. This is therefore also the output formula of
the modified algorithm. The input formula is proved single-head equivalent by
showing that the output formula is single-head.

The claim is proved by inductively showing that $G$ never contains two clauses
with the same head $x$. It initially holds because $G$ is empty. It is
inductively assumed at the beginning of every iteration and proved at the end.
Two cases are considered: $G$ does not contain a clause with head $x$ at the
beginning of the iteration, or it does.

The first case is that $G$ does not contain a clause of head $x$. The only
instruction that changes $G$ is $G = G \cup IT$. Since $G$ does not contain any
clause of head $x$ and $IT$ may at most contain one by construction, $G \cup
IT$ may at most contain one clause of head $x$. This is the inductive claim.

The second case is that $G$ contains a clause of head $x$. It implies that $x$
is not in
{} $H = RCN(B,F) \backslash \{x \mid \exists B' . B' \rightarrow x \in G\}$.
Since $H$ is the set of heads of $IT$, this set does not contain any clause of
head $x$. The instruction $G = G \cup IT$ does not add any clause of head $x$
to $G$, which still contains only one clause of head $x$ after it.~\qed

In the other way around, if the algorithm succeeds the input formula is
single-head equivalent. The converse is proved by
Theorem~\ref{autoreconstruction-correct}. In the modified version of the
algorithm, success and failure alone tell whether the input formula is
single-head or not.

\subsection{Old bodies}

The algorithm uses $UCL(B,F) \cap SCL(B,F)$ as the second argument of
$MINBODIES()$. The first component $UCL(B,F)$ is already known: it is
calculated in advance together with $RCN(B,F)$ for all bodies $B$ in $F$. The
second component $SCL(B,F)$ is instead a problem because it is not bounded by
the size of $F$. The intersection can be rewritten as
{} $\{B' \rightarrow x \in UCL(B,F) \mid B' <_F B\}$,
which however requires two entailment tests for each clause: $F \models B
\rightarrow B'$ and $F \not\models B' \rightarrow B$.

A more efficient way to obtain it is to accumulate all clauses $UCL(B,F)$ along
the way in a set $U$. During the iteration for $B$, this set $U$ contains all
clauses of $F$ with a body that is strictly less than $B$. It may also contain
other clauses, but intersecting with $UCL(B,F)$ removes them.

The algorithm requires only small, localized changes: at the beginning, $U$ is
empty; at the end of each step, $UCL(B,F)$ is added to $U$ by $U = U \cup
UCL(B,F)$; finally, $UCL(B,F) \cap U$ replaces $UCL(B,F) \cap SCL(B,F)$.

\begin{lemma}
\label{accumulate-used}

If $B' \rightarrow x \in UCL(B,F)$, then
$B' \rightarrow x \in SCL(B,F)$
if and only
$B' \rightarrow x \in U$,
where $U$ is the union of $UCL(A,F)$ for
all sets $A$ used in the previous iterations of the autoreconstruction
algorithm
(Algorithm~\ref{algorithm-autoreconstruction}).

\end{lemma}

\proof The proof comprises two parts: $B' \rightarrow x \in UCL(B,F) \cap U$
implies $B' \rightarrow x \in SCL(B,F)$, and $B' \rightarrow x \in UCL(B,F)
\cap SCL(B,F)$ implies $B' \rightarrow x \in U$.

The first part starts from the assumptions $B' \rightarrow x \in UCL(B,F)$ and
$B' \rightarrow x \in U$. The first condition is defined as $B' \rightarrow x
\in F$, $F \models B \rightarrow B'$ and $x \not\in B'$, the second as $B'
\rightarrow x \in UCL(A,F)$ with $A$ chosen before $B$. That $A$ was chosen
before $B$ has two consequences: first, $B <_F A$ is not the case as otherwise
$A$ would not have been minimal in $P$ when it was chosen; second, $B \equiv_F
A$ does not hold either, as otherwise $B$ would have been removed from $P$ in
the iteration for $A$. Since neither $B <_F A$ nor $B \equiv_F A$ hold, $B
\leq_F A$ does not hold either. By definition, $F \not\models A \rightarrow B$
follows. The condition $B' \rightarrow x \in UCL(A,F)$ implies $F \models A
\rightarrow B'$. If $F \models B' \rightarrow B$ then by transitivity $F
\models A \rightarrow B$, which contradicts $F \not\models A \rightarrow B$.
Therefore, the converse $F \not\models B' \rightarrow B$ holds. The conditions
$F \not\models B' \rightarrow B$, $F \models B \rightarrow B'$, $B' \rightarrow
x \in F$ and $x \not\in B'$ imply $B' \rightarrow x \in SCL(B,F)$.

The reverse direction is proved by assuming $B' \rightarrow x \in UCL(B,F)$ and
$B' \rightarrow x \in SCL(B,F)$. These conditions imply $x \not\in B'$, $B'
\rightarrow x \in F$, $F \models B \rightarrow B'$ and $F \not\models B'
\rightarrow B$. The latter two define $B' <_F B$. Since $B'$ is the
precondition of a clause of $F$, it is initially in $P$. When $B$ is chosen it
is no longer in $P$, as otherwise $B$ would not be minimal in $P$. Therefore,
$B'$ has been removed from $P$ in a previous step. Let $A$ be the set of
variables chosen at that step. The only instruction that removes elements from
$P$ is $P = P \backslash \{B' \mid F \models A \equiv B\}$. Therefore, $F
\models A \equiv B'$. Equivalence implies entailment: $F \models A \rightarrow
B'$; with $x \not\in B'$ and $B' \rightarrow x \in F$, this is the definition
of $B' \rightarrow x \in UCL(A,F)$. Therefore, $B' \rightarrow x$ is added to
$U$ at the end of the iteration. Since $U$ is never removed element, it still
contains that clause at the iteration when $B$ is chosen.~\qed

The instruction
{} $T = MINBODIES(M, UCL(B,F) \cap SCL(B,F))$
can therefore be replaced by
{} $T = MINBODIES(M, UCL(B,F) \cap U)$
with the only additional cost of accumulating $UCL(B,F)$ in $U$. This is
correct because the lemma proves that as long as clauses of $UCL(B,F)$ are
concerned, there is no difference between $SCL(B,F)$ and $U$.

\subsection{Cherry picking}

The check
{} $SCL(B,F) \cup IT \models BCL(B,F)$
is the most expensive operation in the inner loop of the algorithm. As a
dominating operation, it should be as efficient as possible. Instead, it
requires two sets that are not even bounded by the size of the original
formula: $SCL(B,F)$ and $BCL(B,F)$; they contain consequences of $F$, not only
clauses of $F$. This section reformulates the check to remove $SCL(B,F)$, the
next $BCL(B,F)$.

% aim: prove SCL(B,F) u IT |= BCL(B,F) the same as UCL(B,G u IT) |= BCL(B,F)
%
% 1. SCL(B,F) is equivalent to UCL(B,G,F)
% 2. UCL(B,G u IT) |= BCL(B,F) implies SCL(B,F) u IT |= BCL(B,F)
% 3. SCL(B,F) u IT |= BCL(B,F) implies UCL(B,G u IT) |= BCL(B,F)

Since the formula under construction $G$ entails $SCL(B,F)$ by
Lemma~\ref{g-scl}, it could be used in its place. But while $G$ contains a
subset equivalent to $SCL(B,F)$, it also contains other clauses. In particular,
it contains clauses that are not in $BCL(B,G)$ and are therefore irrelevant
when the precondition is $B$. This is a setback in efficiency when $F$ contains
many bodies that do not entail each other. Restricting to $UCL(B,G)$ excludes
the extra clauses, but is not correct in general.

\[
F = \{ab \rightarrow c, c \rightarrow d\}
\]

The algorithm chooses $B=\{c\}$ as the first body since it is entailed by the
other $\{a,b\}$ but not the other way around. The first iteration ends with $G
= \{c \rightarrow d\}$. The algorithm then moves onto $B = \{a,b\}$. The set
$UCL(B,G)$ contains the clauses of $G$ whose body is entailed by $B$ according
to $G$. The key here is ``according to $G$'': since $G$ does not contain $ab
\rightarrow c$, the clause $c \rightarrow d$ does not have its body entailed by
$B = \{b,c\}$. Since it is the only clause of $G$, it follows that $UCL(B,G)$
is empty. Since $IT$ cannot contain a clause with head $d$, the clause $ab
\rightarrow d \in BCL(B,F)$ cannot be entailed by $UCL(B,G) \cup IT$.

The problem is ``according to $G$''. The formula under construction is still
too weak. Selecting clauses from $G$ is correct, but using $G$ for deciding
body entailment is not. The clause $c \rightarrow d$ is expected to come out
from this selection because its body $c$ is entailed by $B = \{a,b\}$ and is
therefore relevant to the reconstruction of the clauses whose body is entailed
by $B$. However, $c$ is entailed by $B$ only according to $F$, not to $G$.

\

% first: SCL(B,F) is equivalent to UCL(B,G,F)

The solution is to use $G$ as the source of clauses and $F$ for entailment
between bodies. The first results are established on this modified selection
function.

\[
UCL(B,G,F) = \{B' \rightarrow x \in G \mid F \models B \rightarrow B'\}
\]

The condition $x \not\in B'$ is not necessary because $G$ is already proved not
to contain tautologies by Lemma~\ref{g-it-taut}.

This set is equivalent to $SCL(B,F)$. This is proved in two parts: first,
$UCL(B,G,F)$ is a subset of $SCL(B,F)$; second, it entails it.

\begin{lemma}
\label{ucl-in-scl}

At every iteration of the algorithm
(Algorithm~\ref{algorithm-autoreconstruction}),
$UCL(B,G,F) \subseteq SCL(B,F)$ holds.

\end{lemma}

\proof The definition of $A' \rightarrow x \in UCL(B,G,F)$ is $A' \rightarrow x
\in G$ and $F \models B \rightarrow A'$. By Lemma~\ref{g-subset-bcl}, $A'
\rightarrow x \in G$ implies $A' \rightarrow x \in BCL(A,F)$ for some $A$ such
that $B \not\leq_F A$. The definition of $A' \rightarrow x \in BCL(A,F)$ is $x
\not\in A'$, $F \models A' \rightarrow x$ and $F \models A \equiv A'$. Since $B
\not\leq_F A$, the latter implies $B \not\leq_F A'$, which means $F \not\models
A' \rightarrow B$. All parts of the definition of $A' \rightarrow x \in
SCL(B,F)$ are proved: $x \not\in A'$, $F \models A' \rightarrow x$, $F \models
B \rightarrow A'$ and $F \not\models A' \rightarrow B$.~\qed

This lemma trivially implies $SCL(B,F) \models UCL(B,G,F)$. The converse
entailment requires a preliminary lemma that shows that $G$ and $UCL(B,G,F)$
are the same when checking entailment of clauses whose body is entailed by $B$.

\begin{lemma}
\label{ucl-for-a}

For every two formulae $F$ and $G$,
if $G \models A \rightarrow x$,
$F \models G$ and
$F \models B \rightarrow A$
then
$UCL(B,G,F) \models A \rightarrow x$.

\end{lemma}

\proof Proved by induction on the number of variables of $G$. The base case is
when $G$ only contains a single variable. The only definite clauses of one
variable are $x \rightarrow x$ and $\emptyset \rightarrow x$. The first is a
tautology and is therefore entailed by $UCL(B,G,F)$. The second is in
$UCL(B,G,F)$ since $F \models B \rightarrow \emptyset$.

The induction conclusion is that $UCL(B,G,F) \models A \rightarrow x$ follows
from $G \models A \rightarrow x$, $F \models G$ and $F \models B \rightarrow
A$. If $x$ is in $A$ then $A \rightarrow x$ is a tautology and is therefore
entailed by $UCL(B,G,F)$. Otherwise, $x \not\in A$ and $G \models A \rightarrow
x$ are the premises of Lemma~\ref{set-implies-set}, which proves the existence
of a clause $A' \rightarrow x \in G$ such that $G^x \models A \rightarrow A'$.
Since $F$ entails $G$ it also entails its subset $G^x$ and its consequences,
including $A \rightarrow A'$. This entailment $F \models A \rightarrow A'$
forms with $A' \rightarrow x \in G$ the definition of $A' \rightarrow x \in
UCL(B,G,F)$.

Since $G^x$ entails $A \rightarrow A'$, it entails $A \rightarrow a$ for every
$a \in A'$. The other two assumptions of the induction hypothesis have already
been proved: $F \models G^x$ and $F \models B \rightarrow A$. Its conclusion is
$UCL(B,G^x,F) \models A \rightarrow a$. Since this is the case for each $a \in
A'$, $UCL(B,G^x,F) \models A \rightarrow A'$ follows. By construction,
$UCL(B,G^x,F)$ is a subset of $UCL(B,G,F)$ since the formula $F$ that dictates
with clauses of $G$ or $G^x$ are selected is the same and $G^x$ is a subset of
$G$. As a result, $UCL(B,G,F) \models UCL(B,G^x,F)$. Therefore, $UCL(B,G,F)
\models A \rightarrow A'$. Along with $A' \rightarrow x \in UCL(B,G,F)$, this
implies $UCL(B,G,F) \models A \rightarrow x$.~\qed

This lemma allows proving the converse entailment of Lemma~\ref{ucl-in-scl}.

\begin{lemma}
\label{ucl-implies-scl}

At every iteration of the algorithm, $UCL(B,G,F) \models SCL(B,F)$.

\end{lemma}

\proof Let $A \rightarrow x \in SCL(B,F)$. The claim is that $UCL(B,G,F)
\models A \rightarrow x$. By Lemma~\ref{g-scl}, $A \rightarrow x \in SCL(B,F)$
implies $G \models A \rightarrow x$. The definition of $A \rightarrow x \in
SCL(B,F)$ includes $F \models B \rightarrow A$. Since $F \models G \cup IT$ by
Lemma~\ref{f-implies-g-it}, also $F \models G$ holds. All preconditions of
Lemma~\ref{ucl-for-a} are met. Its conclusion is $UCL(B,G,F) \models A
\rightarrow x$.~\qed

The final destination of this three lemmas is proving that $UCL(B,G,F)$ and
$SCL(B,F)$ are equivalent in the algorithm. This is an obvious consequence of
the above lemmas, and is formally stated in the following.

\begin{lemma}
\label{ucl-is-scl}

At every iteration of the autoreconstruction algorithm
(Algorithm~\ref{algorithm-autoreconstruction}),
$UCL(B,G,F) \equiv SCL(B,F)$ holds.

\end{lemma}

\proof By Lemma~\ref{ucl-in-scl}, $UCL(B,G,F) \subseteq SCL(B,F)$, which
implies $SCL(B,F) \models UCL(B,G,F)$. Lemma~\ref{ucl-implies-scl} proves the
other direction: $UCL(B,G,F) \models SCL(B,F)$. Implication in both directions
is equivalence.~\qed

This lemma could directly be applied to the program by checking
{} $UCL(B,G,F) \cup IT \models BCL(B,F)$
instead of
{} $SCL(B,F) \cup IT \models BCL(B,F)$
and modifying the function that calculates $UCL(B,F)$ by the addition of a
third argument. This is however not necessary. While $SCL(B,F)$ is not
equivalent to $G$, the whole formula $SCL(B,F) \cup IT$ can be replaced by
$UCL(B,G \cup IT)$.

These formulae are not equivalent. Yet, they are the same as long as entailment
of $BCL(B,F)$ is concerned. This is proved in two parts: first,
{} $UCL(B,G \cup IT) \models BCL(B,F)$
implies
{} $SCL(B,F) \cup IT \models BCL(B,F)$;
second, the converse.

\

% second, UCL(B,G u IT) |= BCL(B,F) implies SCL(B,F) u IT |= BCL(B,F)

The assumption of the first part is
{} $UCL(B,G \cup IT) \models BCL(B,F)$.
Its claim is
{} $SCL(B,F) \cup IT \models BCL(B,F)$.
The first step is that $IT$ can be factored out from $UCL(B,G \cup IT)$.

\begin{lemma}
\label{ucl-g-it}

At every iteration of the autoreconstruction algorithm
(Algorithm~\ref{algorithm-autoreconstruction}),
$SCL(B,F) \cup IT \equiv UCL(B,G \cup IT,F)$ holds.

\end{lemma}

\proof An immediate consequence of the definition is
$UCL(B,G \cup IT,F) = UCL(B,G,F) \cup UCL(B,IT,F)$.

By Lemma~\ref{ucl-is-scl}, $UCL(B,G,F)$ is equivalent to $SCL(B,F)$.

Since $IT \subseteq BCL(B,F)$ by Lemma~\ref{it-bcl}, it holds $F \models B
\rightarrow B'$ for all clauses $B' \rightarrow x$ of $IT$. As a result,
$UCL(B,IT,F)$ contains all of them: $UCL(B,IT,F) = IT$.~\qed

Since $SCL(B,F) \cup IT$ is equivalent to $UCL(B,G \cup IT,F)$, if the latter
is equivalent to $UCL(B,G \cup IT,G \cup IT)$, the target is hit because this
formula is the same as $UCL(B,G \cup IT)$.

The difference between $UCL(B,G \cup IT,F)$ and $UCL(B,G \cup IT,G \cup IT)$ is
the selection criteria:
{} $F \models B \rightarrow B'$
versus
{} $G \cup IT \models B \rightarrow B'$
where $B'$ is the body of a clause. These conditions differ in general. Yet,
they coincide under the current assumption $UCL(B,G \cup IT) \models BCL(B,F)$.

\begin{lemma}
\label{ucl-b-bprime}

If $UCL(B,G \cup IT) \models BCL(B,F)$ then
{} $F \models B \rightarrow B'$
is equivalent to
{} $G \cup IT \models B \rightarrow B'$
for every set of variables $B'$.

\end{lemma}

\proof By Lemma~\ref{f-implies-g-it} and Lemma~\ref{f-implies-g-it-h}, $F$
entails $G \cup IT$. As a result, if $G \cup IT \models B \rightarrow B'$, then
$F \models B \rightarrow B'$.

The rest of the proof obtains $G \cup IT \models B \rightarrow B'$ from $F
\models B \rightarrow B'$. Since $F \models B \rightarrow B'$, then $F \models
B \rightarrow b$ holds for every $b \in B'$. If $b \in B$ then $B \rightarrow
b$ is a tautology and is therefore entailed by $BCL(B,F)$. Otherwise, $b
\not\in B$ and $F \models B \rightarrow b$ with the trivially true $F \models B
\equiv B$ define $B \rightarrow b \in BCL(B,F)$. Either way, $B \rightarrow b$
is entailed by $BCL(B,F)$ for every $b \in B'$. By assumption, $UCL(B,G \cup
IT) \models BCL(B,F)$, which implies $G \cup IT \models B \rightarrow b$ since
$UCL(B,G \cup IT)$ is defined as a subset of $G \cup IT$. This holds for every
$b \in B'$, proving $G \cup IT \models B \rightarrow B'$.~\qed

The first part of the proof can now be completed.

\begin{lemma}
\label{ucl-scl-it}

If $UCL(B,G \cup IT) \models BCL(B,F)$
then $SCL(B,F) \cup IT \models BCL(B,F)$.

\end{lemma}

\proof The only difference between $UCL(B,G \cup IT)$ and $UCL(B,G \cup IT,G
\cup IT)$ is that the former does not contain tautologies. The latter does not
as well because it is a subset of $G \cup IT$, which does not by
Lemma~\ref{g-it-taut}.

The set $UCL(B,G \cup IT,G \cup IT)$ contains some clauses $G \cup IT$ like
$UCL(B,G \cup IT,F)$ does, but differs from it in the selection criteria: $G
\cup IT \models B \rightarrow B'$ instead of $F \models B \rightarrow B'$. The
assumption $UCL(B,G \cup IT) \models BCL(B,F)$ proves that $F$ and $G \cup IT$
entail the same clauses $B \rightarrow B'$ by Lemma~\ref{ucl-b-bprime}.
Therefore, $UCL(B,G \cup IT,G \cup IT)$ is identical to $UCL(B,G \cup IT,F)$.

Lemma~\ref{ucl-g-it} proves that $UCL(B,G \cup IT,F)$ is equivalent to
$SCL(B,F) \cup IT$.

Summarizing, the assumption $UCL(B,G \cup IT) \models BCL(B,F)$ implies that
$UCL(B,G \cup IT)$ is the same as $UCL(B,G \cup IT,F)$, which is the same as
$SCL(B,F) \cup IT$; this equality and the assumption $UCL(B,G \cup IT) \models
BCL(B,F)$ imply $SCL(B,F) \cup IT \models BCL(B,F)$.~\qed

The proof looks like a sequence of equivalences. As such, it would also prove
the opposite direction, saving the cost of a further proof. But this is a false
impression: $UCL(B,G \cup IT)$ is not always equivalent to $SCL(B,F) \cup IT$,
it is only when $UCL(B,G \cup IT) \models BCL(B,F)$ holds.

\

% third, SCL(B,F) u IT |= BCL(B,F) implies UCL(B,G u IT) |= BCL(B,F)

The converse direction has to be proved separately:
{} $SCL(B,F) \cup IT \models BCL(B,F)$
implies
{} $UCL(B,G \cup IT) \models BCL(B,F)$.

The first step in this direction is that $UCL()$ maintains certain entailments
of the formula it is applied to.

\begin{lemma}
\label{survive-ucl}

For every formulae $F$ and $G$, variable $x$ and sets of variables $B$ and $B'$
such that $F \models G$ and $F \models B \rightarrow B'$, it holds $G \models
B' \rightarrow x$ if and only if $UCL(B,G,F) \models B' \rightarrow x$.

\end{lemma}

\proof Since $UCL(B,G,F)$ only comprises clauses of $G$ by definition, it is
entailed by $G$. This proves the first direction of the claim: if $UCL(B,G,F)
\models B' \rightarrow x$ then $G \models B' \rightarrow x$.

The converse direction is proved by contradiction: $G \models B' \rightarrow x$
and $UCL(B,G,F) \not\models B' \rightarrow x$ do not both hold at the same
time. The latter is equivalent to the existence of a model $M$ that satisfies
$UCL(B,G,F) \cup B' \cup \{\neg x\}$.

Let $M'$ be the model that evaluates all variables of $BCN(B,F)$ like $M$ and
all others to false: if $F \models B \rightarrow z$ then $z$ has the same value
in $M$ and $M'$; otherwise, $z$ is false in $M'$. This model $M'$ is proved to
satisfy every element of $G \cup B' \cup \{\neg x\}$, contradicting the
assumption $G \models B' \rightarrow x$. The proof is done separately for the
elements of $G \backslash UCL(B,G,F)$, of $UCL(B,F,F)$, of $B'$ and of $\{\neg
x\}$.

Let $B'' \rightarrow y$ be a clause of $G$ but not of $UCL(B,G,F)$. Since
$UCL(B,G,F)$ contains all clauses of $G$ except those whose body is not
entailed by $B$, this is only possible if $F \not\models B \rightarrow B''$. As
a result, $F \not\models B \rightarrow b$ for some $b \in B''$. By
construction, $M'$ evaluates $b$ to false and therefore satisfies $B''
\rightarrow y$.

The definition of $B'' \rightarrow y \in UCL(B,G,F)$ is $B'' \rightarrow y \in
G$ and $F \models B \rightarrow B''$. The latter implies $F \models B
\rightarrow b$ for every $b \in B''$. Therefore, $M'$ evaluates every $b \in
B''$ in the same way as $M$ does. Since $F \models G$ holds by
Lemma~\ref{f-implies-g-it}, $B'' \rightarrow y \in G$ implies $F \models B''
\rightarrow y$. With $F \models B \rightarrow B''$, it implies $F \models B
\rightarrow y$. Therefore, $y$ is evaluated by $M'$ in the same way as $M$
does. This proves that all variables in every $B'' \rightarrow y \in
UCL(B,G,F)$ is evaluated by $M'$ in the same way as $M$ does. Since $M$
satisfies $UCL(B,G,F)$, also $M'$ does.

Since $B' \rightarrow b$ is a tautology for every $b \in B$, it is entailed by
$F$. Which also entails $B \rightarrow B'$. By transitivity, $F$ entails $B
\rightarrow b$ for every $b \in B'$. As a result, $M'$ evaluates all variables
of $B'$ in the same way as $M$. Since $M$ satisfies $B'$ by assumption, also
$M'$ does.

A consequence of $F \models B \rightarrow B'$ and $G \models B' \rightarrow x$
is $F \models B \rightarrow x$ since $F \models G$. It implies that $M'$
evaluates $\neg x$ in the same way as $M$, which satisfies it.

This proves that $M'$ satisfies $G \backslash UCL(B,G,F) \cup UCL(B,G,F) \cup
B' \cup \{\neg b\}$, which is the same as $G \cup B' \cup \{\neg x\}$. The
satisfiabily of this set is the same as $G \not\models B' \rightarrow x$, which
contradicts the assumption $G \models B' \rightarrow x$.~\qed

This lemma tells that $UCL(B,G,F)$ is the same as $G$ when checking entailment
of clauses whose body is entailed by $B$ according to $F$. It also applies to
the case where the two formulae $G$ and $F$ are the same.

\begin{lemma}
\label{scl-it-ucl}

If $SCL(B,F) \cup IT \models BCL(B,F)$
then $UCL(B,G \cup IT) \models BCL(B,F)$.

\end{lemma}

\proof Lemma~\ref{g-scl} proves $G \models SCL(B,F)$. With the assumption
{} $SCL(B,F) \cup IT \models BCL(B,F)$,
it implies
{} $G \cup IT \models BCL(B,F)$.

Let $B' \rightarrow x \in BCL(B,F)$. By definition, $F \models B \rightarrow
B'$. This implies $F \models B \rightarrow b$ for every $b \in B'$. If $b \in
B$ then $B \rightarrow b$ is a tautology and is therefore entailed by
$BCL(B,F)$. Otherwise, $b \not\in B$, $F \models B \rightarrow b$ and the
trivially true $F \models B \rightarrow B$ imply $B \rightarrow b \in
BCL(B,F)$. Either way, $B \rightarrow b$ is entailed by $BCL(B,F)$. Since $G
\cup IT$ entails this set, it also entails $B \rightarrow b$ as well. This hold
for all $b \in B'$; therefore, $G \cup IT \models B \rightarrow B'$.

Lemma~\ref{survive-ucl} applies to $UCL(B,G \cup IT,G \cup IT)$: since $G \cup
IT \models G \cup IT$, $G \cup IT \models B \rightarrow B'$ and $G \cup IT
\models B' \rightarrow x$, it follows $UCL(B,G \cup IT,G \cup IT) \models B'
\rightarrow x$. This is the same as $UCL(B,G \cup IT) \models B' \rightarrow x$
since the only difference between the definitions of $UCL(B,G \cup IT,G \cup
IT)$ and $UCL(B,G \cup IT)$ is that the latter does not contain tautologies,
but the former does not as well since it is a subset of $G \cup IT$, which does
not by Lemma~\ref{g-it-taut}.~\qed

This lemma proves that
{} $SCL(B,F) \cup IT \models BCL(B,F)$
implies
{} $UCL(B,G \cup IT) \models BCL(B,F)$.
The converse implication is Lemma~\ref{ucl-scl-it}. The conclusion is that
these two conditions are the same.

\begin{lemma}
\label{scl-it-g-it}

The conditions $SCL(B,F) \cup IT \models BCL(B,F)$
and $UCL(B,G \cup IT) \models BCL(B,F)$
are equivalent.

\end{lemma}

\proof Immediate consequence of Lemma~\ref{ucl-scl-it} and
Lemma~\ref{scl-it-ucl}.~\qed

\subsection{Ubi minor major cessat}

The previous section proves that the dominant operation of the algorithm
$SCL(B,F) \cup IT \models BCL(B,F)$ is equivalent to $UCL(B,G \cup IT) \models
BCL(B,F)$. While $SCL(B,F)$ is a set of consequences and can therefore be very
large, $UCL(B,G \cup IT)$ is a subset of $G \cup IT$ and is therefore bounded
by the size of the formula under construction $G$ and the candidate set $IT$.
The latter is in turn bounded because it is single-head: it has no more clauses
than the variables in $F$.

The problem remains on the other side of the implication: $BCL(B,F)$ contains
all clauses $B' \rightarrow x$ entailed by $F$ such that $F \models B
\rightarrow B'$. Many such clauses may be entailed even from a small formula
$F$.

The solution is to use $UCL(B,F)$ in place of $BCL(B,F)$. This is the set of
clauses of $F$ whose body is entailed by $B$ according to $F$. It is monotonic
with respect to entailment of the formula, as the following lemma proves.

\begin{lemma}
\label{ucl-entails}

If $F \models F'$ then $UCL(B,F) \models UCL(B,F')$.

\end{lemma}

\proof Every clause $B' \rightarrow x$ of $UCL(B,F')$ satisfies $B' \rightarrow
x \in F'$ and $F' \models B \rightarrow B'$ by definition. The former implies
$F' \models B' \rightarrow x$. Since $F \models F'$, these two entailments
carry over to the other formula: $F \models B' \rightarrow x$ and $F \models B
\rightarrow B'$. By Lemma~\ref{ucl-only}, these two conditions imply $UCL(B,F)
\models B' \rightarrow x$. This argument shows that every clause of $UCL(B,F')$
is entailed by $UCL(B,F)$.~\qed

Another result about $UCL()$ is that $RCN()$ is invariant with respect to
replacing $F$ with $UCL(B,F)$, as it is $BCN()$ as proved by
Lemma~\ref{ucl-bcn}.

\begin{lemma}
\label{ucl-rcn}

For every formula $F$ and set of variables $B$ it holds
$RCN(B,F) = RCN(B,UCL(B,F))$

\end{lemma}

\proof The first step of the proof is $F \models B \rightarrow (BCN(B,F)
\backslash \{x\})$. The definition of $y \in BCN(B,F)$ is $F \models B
\rightarrow y$. This is the case for every $y \in BCN(B,F)$; therefore, $F
\models B \rightarrow BCN(B,F)$ follows. Since $BCN(B,F) \backslash \{x\}$ is a
subset of $BCN(B,F)$, also
{} $F \models B \rightarrow (BCN(B,F) \backslash \{x\})$ holds.

This entailment is the precondition of Lemma~\ref{ucl-only}, which proves that
$F$ and $UCL(B,F)$ entail the same clauses whose body is entailed by $B$. In
this case,
{} $F \models B \rightarrow (BCN(B,F) \backslash \{x\})$
proves that $F \models (BCN(B,F) \backslash \{x\}) \rightarrow x$ is the same
as $UCL(B,F) \models (BCN(B,F) \backslash \{x\}) \rightarrow x$. These two
entailments define $x \in RCN(B,F)$ and $x \in RCN(B,UCL(B,F))$. Since they are
the same, $RCN(B,F)$ and $RCN(B,UCL(B,F))$ coincide.~\qed

This lemma allows proving $UCL(B,F)$ can replace $BCL(B,F)$ in a certain
situation.

A semantical counterpart of Lemma~\ref{ucl-f-scl-bcl} shows that $UCL(B,F)$ is
equivalent to the union of the two other considered sets of clauses.

\begin{lemma}
\label{ucl-scl-bcl}

For every formula $F$ and set of variables $B$,
it holds $UCL(B,F) \equiv SCL(B,F) \cup BCL(B,F)$.

\end{lemma}

\proof Lemma~\ref{ucl-f-scl-bcl} proves $UCL(B,F) = F \cap (SCL(B,F) \cup
BCL(B,F))$. As a subset of $SCL(B,F) \cup BCL(B,F)$, the set $UCL(B,F)$ is
entailed by it: $SCL(B,F) \cup BCL(B,F) \models UCL(B,F)$.

The converse implication $UCL(B,F) \models SCL(B,F) \cup BCL(B,F)$ is proved by
showing that every clause $B' \rightarrow x$ of $SCL(B,F) \cup BCL(B,F)$ is
entailed by $UCL(B,F)$. The definition of $B' \rightarrow x \in SCL(B,F) \cup
BCL(B,F)$ includes $F \models B' \rightarrow x$ and either $B' <_F B$ or $B'
\equiv_F B$. The two alternatives are equivalent to $B' \leq_F B$, which is
defined as $F \models B \rightarrow B'$. Lemma~\ref{ucl-only} proves that $F
\models B \rightarrow B'$ and $F \models B' \rightarrow x$ imply $UCL(B,F)
\models B' \rightarrow x$.~\qed

Lemma~\ref{g-scl} proves that $G$ entails $SCL(B,F)$. This fact cannot be
pushed from $G$ to $UCL(B,G)$ since $UCL(B,G)$ does not contain the clauses of
$G$ whose body is not entailed by $G$. Adding $IT$ and requiring $BCL(B,F)$ to
be entailed fills the gap.

\begin{lemma}
\label{ucl-bcl-scl}

If $UCL(B,G \cup IT) \models BCL(B,F)$
then $UCL(B,G \cup IT) \models SCL(B,F)$.

\end{lemma}

\proof The claim is proved by showing that every clause $B' \rightarrow x$ of
$SCL(B,F)$ is entailed by $UCL(B,G \cup IT)$.

The definition of $B' \rightarrow x \in SCL(B,F)$ includes $B' <_F B$, which
includes $F \models B \rightarrow B'$. This is the same as $F \models B
\rightarrow b$ for every $b \in B'$. If $b \in B$ then $B \rightarrow b$ is a
tautology and is therefore entailed by $BCL(B,F)$. Otherwise, $b \not\in B$ and
$F \models B \rightarrow b$ and the trivially true $F \models B \equiv B$
define $B \rightarrow b \in BCL(B,F)$. Either way, $BCL(B,F)$ entails $B
\rightarrow b$. Since this is the case for every $b \in B'$, it proves
$BCL(B,F) \models B \rightarrow B'$. Since $UCL(B,G \cup IT)$ entails
$BCL(B,F)$ by assumption, it also entails its consequence $B \rightarrow B'$.
Since $UCL(B,G \cup IT)$ is by definition a subset of $G \cup IT$, monotonicity
implies $G \cup IT \models B \rightarrow B'$.

Lemma~\ref{g-scl} proves $G \models SCL(B,F)$. Therefore, $G \models B'
\rightarrow x$. By monotonicity, $G \cup IT \models B' \rightarrow x$. Since $G
\cup IT \models B \rightarrow B'$ as proved in the previous paragraph, $G \cup
IT \models B' \rightarrow x$ implies $UCL(B,G \cup IT) \models B' \rightarrow
x$ by Lemma~\ref{ucl-only}.~\qed

If $UCL(B,G \cup IT)$ implies $BCL(B,F)$, it also implies $SCL(B,F)$. If it
implies the first, it implies both. Their union is $SCL(B,F) \cup BCL(B,F)$,
which is semantically the same as $UCL(B,F)$. This is proved by the next lemma.

\begin{lemma}
\label{ucl-bcl-ucl}

The conditions
{} $UCL(B,G \cup IT) \models BCL(B,F)$
and
{} $UCL(B,G \cup IT) \equiv UCL(B,F)$
are equivalent.

\end{lemma}

\proof Lemma~\ref{f-implies-g-it} states $F \models G \cup IT$. This is the
precondition of Lemma~\ref{ucl-entails}, which proves
{} $UCL(B,F) \models UCL(B,G \cup IT)$.
This is half of the second condition in the claim,
{} $UCL(B,G \cup IT) \equiv UCL(B,F)$.
It holds regardless of the first.

The claim is therefore proved if the first condition
{} $UCL(B,G \cup IT) \models BCL(B,F)$
is the same as the other half of the second:
{} $UCL(B,G \cup IT) \models UCL(B,F)$.

Lemma~\ref{ucl-scl-bcl} proves $UCL(B,F) \equiv SCL(B,F) \cup BCL(B,F)$;
therefore, the claim is that
{} $UCL(B,G \cup IT) \models BCL(B,F)$
and
{} $UCL(B,G \cup IT) \models SCL(B,F) \cup BCL(B,F)$
are the same. The first entailment follows from the second because its
consequent is a subset of that of the second. The converse is a consequence of
Lemma~\ref{ucl-bcl-scl}:
{} $UCL(B,G \cup IT) \models BCL(B,F)$
implies
{} $UCL(B,G \cup IT) \models SCL(B,F)$.
Therefore, it implies
{} $UCL(B,G \cup IT) \models SCL(B,F) \cup BCL(B,F)$.~\qed

Lemma~\ref{scl-it-g-it} proves that
{} $SCL(B,F) \cup IT \models BCL(B,F)$
is the same as
{} $UCL(B,G \cup IT) \models BCL(B,F)$,
which Lemma~\ref{ucl-bcl-ucl} proves the same as
{} $UCL(B,G \cup IT) \equiv UCL(B,F)$.
The formulae in the equivalence are both bounded in size by $F$ and $G \cup
IT$; the latter is single-head by construction, and is therefore polynomially
bounded by the number of variables.

\

Yet, equivalence is still to be checked. It is the same as mutual inference:
$UCL(B,G \cup IT)$ entails every clause of $UCL(B,F)$ and vice versa. Each
inference check is linear in time, making the total quadratic.

An alternative is to trade space for time. Equivalence is equality in some
cases. For example, if both $A$ and $B$ are deductively closed, they are
equivalent if and only if they coincide. Equivalence is equality on the
deductive closure of the two formulae.

Checking equality is easy, but the deductive closure may be very large. Working
on the sets of prime implicants is a more efficient alternative. In the present
case, a still better alternative is to use the minimal implicants with heads in
$H$. The formula under construction $G$ already entails $SCL(B,F)$; the target
of the current iteration is to entail $BCL(B,F)$ with clauses that have heads
in $H$. Only these clauses matter. The minimal such clauses are $HCLOSE()$.

The final step of this section is to prove that equality is enough if checked
on $HCLOSE()$ when its first argument is $RCN(B,F)$ and the second is either
$UCL(B,G \cup IT)$ or $UCL(B,F)$. These formulae have something in common: the
premise of the following lemma.

\begin{lemma}
\label{hclose-rcn}

If $F \models B \rightarrow B'$ holds for every clause $B' \rightarrow x \in
F$, then $F \equiv HCLOSE(RCN(B,F),F)$.

\end{lemma}

\proof The definition of $B' \rightarrow x \in HCLOSE(H,F)$ includes $F \models
B' \rightarrow x$: every clause of $HCLOSE(H,F)$ is entailed by $F$. This also
happens in the particular case $H=RCN(B,F)$, and proves $F \models
HCLOSE(RCN(B,F),F)$ holds even when the premise of the lemma does not.

The premise is necessary to the converse implication $HCLOSE(RCN(B,F),F)
\models F$. Every clause $B' \rightarrow x$ of $F$ is proved to be entailed by
$HCLOSE(RCN(B,F),F)$.

The premise of the lemma is that $F \models B \rightarrow B'$ holds for every
clause $B' \rightarrow x \in F$. It implies $B' \subseteq BCN(B,F)$. If $x \in
B'$ then $B' \rightarrow x$ is tautological and is therefore entailed by
$HCLOSE(RCN(B,F),F)$. Otherwise, $x \not\in B'$ strengthens $B' \subseteq
BCN(B,F)$ to $B' \subseteq BCN(B,F) \backslash \{x\}$. Since $F$ contains $B'
\rightarrow x$ it also entails it: $F \models B' \rightarrow x$. Monotonicity
implies $F \models (BCN(B,F) \backslash \{x\}) \rightarrow x$. This is the
definition of $x \in RCN(B,F)$.

The definition of $B' \rightarrow x \in HCLOSEALL(RCN(B,F),F)$ is met: $x \in
RCN(B,F)$, $x \not\in B'$ and $F \models B' \rightarrow x$.
Lemma~\ref{hcloseall-hclose} proves that $HCLOSE(RCN(B,F),F)$ contains the
minimal clauses of $HCLOSEALL(RCN(B,F),F)$. If $B' \rightarrow x$ is minimal it
is in $HCLOSE(RCN(B,F),F)$; otherwise, one of its subclauses is in
$HCLOSE(RCN(B,F),F)$, and a subclause always entails its superclauses.~\qed

A particular case where the premise of this lemma holds is when the formula is
$UCL(B,F)$.

\begin{lemma}
\label{hclose-ucl}

It holds $UCL(B,F) \equiv HCLOSE(RCN(B,F),UCL(B,F))$.

\end{lemma}

\proof By definition, $UCL(B,F)$ contains only clauses $B' \rightarrow x$ such
that $F \models B \rightarrow B'$. This entailment is the same as $F \models B
\rightarrow b$ for every $b \in B'$. Lemma~\ref{ucl-body} tells that $F \models
B \rightarrow b$ is the same as $UCL(B,F) \models B \rightarrow b$. This
implies $UCL(B,F) \models B \rightarrow b$ for every $b \in B'$, which is the
same as $UCL(B,F) \models B \rightarrow B'$.

This condition $UCL(B,F) \models B \rightarrow B'$ holds for every $B'
\rightarrow x \in UCL(B,F)$. Lemma~\ref{hclose-rcn} proves $UCL(B,F) \equiv
HCLOSE(RCN(B,F),UCL(B,F))$.~\qed

Another step is the equality of the real consequences of $F$ and $G \cup IT$.

\begin{lemma}
\label{ucl-same-rcn}

If
{} $UCL(B,G \cup IT) \equiv UCL(B,F)$
then $RCN(B,F) = RCN(B,G \cup IT)$.

\end{lemma}

\proof Since $RCN()$ is defined semantically, it is unaffected by replacing a
formula with an equivalent one:
{} $UCL(B,G \cup IT) \equiv UCL(B,F)$
implies
{} $RCN(B,UCL(B,G \cup IT)) = RCN(B,UCL(B,F))$.
The claim follows from Lemma~\ref{ucl-rcn}, which states
{} $RCN(B,F) = RCN(B,UCL(B,F))$ and
{} $RCN(B,G \cup IT) = RCN(B,UCL(B,G \cup IT))$.~\qed

The latest version of the check
{} $SCL(B,F) \cup IT \models BCL(B,F)$
was
{} $UCL(B,G \cup IT) \equiv UCL(B,F)$.
This equivalence can be turned into an equality.

\begin{lemma}
\label{ucl-rcn-hclose}

The condition
{} $UCL(B,G \cup IT) \equiv UCL(B,F)$
is the same as
{} $HCLOSE(RCN(B,F),UCL(B,F)) = HCLOSE(RCN(B,G \cup IT),UCL(B,G \cup IT))$.

\end{lemma}

\proof Lemma~\ref{hclose-ucl} proves that
{} $UCL(B,F)$ is equivalent to $HCLOSE(RCN(B,F),UCL(B,F))$
and
{} $UCL(B,G \cup IT)$ to $HCLOSE(RCN(B,G \cup IT),UCL(B,G \cup IT))$.
If $HCLOSE(RCN(B,F),UCL(B,F))$ is equal to $HCLOSE(RCN(B,G \cup IT),UCL(B,G
\cup IT))$ then $UCL(B,F)$ and $UCL(B,G \cup IT)$ are equivalent to the same
formula. Therefore, they are equivalent.

The other direction of the claim is now proved. Since $HCLOSE()$ is defined
semantically, it has the same value when applied to two identical sets and two
equivalent formulae. Since $RCN(B,F) = RCN(B,G \cup IT)$ as proved by
Lemma~\ref{ucl-same-rcn} and $UCL(B,F) \equiv UCL(B,G \cup IT)$ by assumption,
the claim
{} $HCLOSE(RCN(B,F),UCL(B,F)) = HCLOSE(RCN(B,G \cup IT),UCL(B,G \cup IT))$
follows.~\qed

The chain of Lemma~\ref{scl-it-g-it}, Lemma~\ref{ucl-bcl-ucl} and
Lemma~\ref{ucl-rcn-hclose} prove that the dominant operation of the algorithm
$SCL(B,F) \cup IT \models BCL(B,F)$ is equivalent to
{} $HCLOSE(RCN(B,F),UCL(B,F)) = HCLOSE(RCN(B,G \cup IT),UCL(B,G \cup IT))$.
The first set can be determined from $F$ and $B$ alone, outside the loop over
the possible sets $IT$.

The algorithm that calculates $UCL()$ also produces $RCN()$. As a result,
$RCN(B,G \cup IT)$ is determined at no additional cost when $UCL(B,G \cup IT)$
is calculated.

A slight simplification for $HCLOSE(RCN(B,F),UCL(B,F))$ is that
$HCLOSE(HEADS(B,F),UCL(B,F))$ is already known as a part of the mechanism for
determining the bodies of the clauses of $IT$. Since $HEADS(B,F) \subseteq
RCN(B,F)$ by definition, the required set $HCLOSE(RCN(B,F),UCL(B,F))$ can be
calculated as $HCLOSE(HEADS(B,F),UCL(B,F)) \cup HCLOSE(RCN(B,F) \backslash
HEADS(B,F),UCL(B,F))$.

\subsection{Necessity is the mother of invention}

The dominating operation
{} $SCL(B,F) \cup IT \models BCL(B,F)$
is simplified as much as possible to improve the algorithm efficiency. Yet, even
in the form
{} $HCLOSE(RCN(B,F),UCL(B,F)) = HCLOSE(RCN(B,G \cup IT),UCL(B,G \cup IT))$
it is expensive as it requires computing a resolution closure, albeit bounded
by the heads.

The good news is that sometimes it is not necessary. For example, if $RCN(B,F)
\not= RCN(B,G \cup IT)$ then $G \cup IT$ fails at replicating $F$ since they
differ on their consequences of $B$; the candidate $IT$ can be discarded
without further analysis. The additional cost of this check is only that of
comparing two sets, since both $RCN(B,F)$ and $RCN(B,G \cup IT)$ are necessary
anyway.

Other necessary conditions exist.

They are all proved as:
{} $SCL(B,F) \cup IT \models BCL(B,F)$
implies something. By contraposition, not something implies not
{} $SCL(B,F) \cup IT \models BCL(B,F)$.
If that something is easy to determine, it saves from the expensive entailment
test when false: $IT$ is discarded straight away.

Any of the conditions that are equivalent to
{} $SCL(B,F) \cup IT \models BCL(B,F)$
can be used in its place:
{} $UCL(B,G \cup IT) \models BCL(B,F)$,
{} $UCL(B,G \cup IT) \equiv UCL(B,F)$,
and
{} $HCLOSE(RCN(B,F),UCL(B,F)) = HCLOSE(RCN(B,G \cup IT),UCL(B,G \cup IT))$.

The trick can be repeated as many times as useful: if
{} $SCL(B,F) \cup IT \models BCL(B,F)$
implies condition one, condition two and condition three, they are all checked;
failure in any stops the iteration over $IT$ since it fails
{} $SCL(B,F) \cup IT \models BCL(B,F)$.

\subsubsection{The first necessary condition}

The first necessary condition to
{} $SCL(B,F) \cup IT \models BCL(B,F)$
is that every variable in the bodies of
{} $HCLOSE(RCN(B,F),UCL(B,F))$
occurs in the body of a clause of $G \cup IT$.

A formal proof is an overshooting. The condition
{} $SCL(B,F) \cup IT \models BCL(B,F)$
is the same as the equality of
{} $HCLOSE(RCN(B,F),UCL(B,F))$
and
{} $HCLOSE(RCN(B,G \cup IT),UCL(B,G \cup IT))$.
Equality implies that every variable in a body of the former is in a body of
the latter. Since the latter can be obtained by resolving clauses of $UCL(B,G
\cup IT)$ and resolution does not create literals, that variable is in a body
of $UCL(B,G \cup IT)$. Since $UCL(B,G \cup IT)$ is defined as a subset of $G
\cup IT$, that variable is also in a body of $G \cup IT$.

In the other way around, if $G \cup IT$ contains no body with a variable that
is in a body of $HCLOSE(RC(B,F),UCL(B,F))$ then
{} $HCLOSE(RCN(B,F),UCL(B,F)) = HCLOSE(RCN(B,G \cup IT),UCL(B,G \cup IT))$
is false.

This necessary condition can be seen as the version on bodies of the condition
over the heads. But the condition on the heads is strict: the heads of
$ITERATION(B,F)$ are exactly $HEADS(B,F)$. Instead, a specific body of
$HCLOSE(HEADS(B,F),UCL(B,F))$ may occur in $ITERATION(B,F)$ or not. The example
in {\tt nobody.py} shows such a case: $F = \{a \rightarrow b, b \rightarrow a,
c \rightarrow d, d \rightarrow c, ac \rightarrow e\}$ and $B = \{a,b\}$; it
makes $a$ equivalent to $b$ and $c$ to $d$; while $ac \rightarrow e$ is in
$HCLOSE(HEADS(B,F),UCL(B,F))$, the iteration function $ITERATION(B,F) = \{bd
\rightarrow e\}$ is valid in spite of not containing a body $ac$. The example
{\tt twobodies.py} instead shows a case when the same body occurs multiple
times: $F = \{a \rightarrow b, a \rightarrow c\}$ and $B=\{a\}$. The only valid
iteration function has $ITERATION(B,F)$ equal to $F$ itself, which contains the
body $a$ twice.

The {\tt reconstruct.py} program exploits this condition by checking whether
the union of the bodies of $G \cup IT$ contains the union of the bodies of
$HCLOSE(RCN(B,F),UCL(B,F))$, failing $IT$ if it does not.

\

This condition involves $IT$: the bodies of $G \cup IT$ contain all variables
in the bodies of $HCLOSE(RCN(B,F),UCL(B,F))$. Therefore, it can only be checked
inside the loop over the possible sets $IT$. However, the bodies of $IT$ are by
construction bodies of $HCLOSE(HEADS(B,F),UCL(B,F))$. This provides a further
necessary condition: the bodies of $G$ and $HCLOSE(HEADS(B,F),UCL(B,F))$
contain all variables in the bodies of $HCLOSE(RCN(B,F),UCL(B,F))$. This other
necessary condition involves $IT$ no longer. Therefore, it can be checked
before entering the loop over the possible sets $IT$.

\

One outside the loop, one inside the loop. Both necessary conditions are based
on the bodies of $G \cup IT$ containing those of $HCLOSE(RCN(B,F),UCL(B,F))$;
they differ on employing the actual bodies of $IT$ (only given inside the loop)
or their upper bound $HCLOSE(HEADS(B,F),UCL(B,F))$ (known outside the loop).

Given the length of these formulae, the following shortcuts are employed:
{} $H = HEADS(B,F)$,
{} $R = RCN(B,F)$ and
{} $U = UCL(B,F)$.
If $BD(F)$ is the set of variables in the bodies of a formula, the
conditions are:

\begin{eqnarray*}
BD(HCLOSE(R,U))
& \subseteq &
BD(G) \cup
BD(IT) \\
BD(HCLOSE(R,U))
& \subseteq &
BD(G) \cup
BD(HCLOSE(H,U))
\end{eqnarray*}

They share some components. Since $H = HEADS(B,F)$ is a subset of $R =
RCN(B,F)$, the latter is the union of the former and their set difference:
{} $R = H \cup (R \backslash H)$.

This reformulation propagates to the head closure of these sets: by definition,
$HCLOSE(A \cup B,F)$ comprises the clauses $B' \rightarrow x$ that obey certain
conditions including $x \in A \cup B$ and not involving $A \cup B$ in any other
way. As a result, $HCLOSE(A \cup B,F) = HCLOSE(A,F) \cup HCLOSE(B,F)$.

In the present case,
{} $HCLOSE(R,U)$ 
is equal to
{} $HCLOSE(H,U) \cup
{}  HCLOSE(R \backslash H,U)$.
If two sets of clauses are the same, the variables in their bodies are the
same:
{} $BD(HCLOSE(R,U) =
{}  BD(HCLOSE(H,U)) \cup
{}  BD(HCLOSE(R \backslash H,U))$.

This equality allows reformulating the two necessary conditions as follows.

\begin{eqnarray*}
BD(HCLOSE(H,U)) \cup
BD(HCLOSE(R \backslash H,U))
& \subseteq &
BD(G) \cup
BD(IT) \\
BD(HCLOSE(H,U)) \cup
BD(HCLOSE(R \backslash H,U))
& \subseteq &
BD(G) \cup
BD(HCLOSE(H,U))
\end{eqnarray*}

In the second containment, $BD(HCLOSE(H,U))$ is in both sides. Since a set
is always contained in itself, it can be removed from the left-hand side.

\begin{eqnarray*}
BD(HCLOSE(H,U)) \cup
BD(HCLOSE(R \backslash H,U))
& \subseteq &
BD(G) \cup
BD(IT) \\
BD(HCLOSE(R \backslash H,U))
& \subseteq &
BD(G) \cup
BD(HCLOSE(H,U))
\end{eqnarray*}

The {\tt reconstruct.py} program checks these two conditions employing the
following two sets.

\begin{eqnarray*}
IB
&=&
BD(HCLOSE(H,U)) \backslash 
BD(G)
\\
HL
&=&
BD(HCLOSE(R \backslash H,U)) \backslash
BD(G) \backslash
IB
\end{eqnarray*}

% involved in implying = implies with other variables

The first set is called {\tt inbodies} in the program; it comprises the
variables that are involved in implying a variable in $HEADS(B,F)$ and are not
in a body of $G$. The second is called {\tt headlessbodies}; it comprises the
variables that are involved in implying a variable that is a head in $G$, but
cannot do that because they are neither in $G$ nor in $IB$.

The second set $HL$ is used to determine the second necessary condition, the
one that is independent of $IT$. Making its definition explicit in the
definition tells why.

\begin{eqnarray*}
HL
&=&
BD(HCLOSE(R \backslash H,U)) \backslash
BD(G) \backslash
IB
\\
&=&
BD(HCLOSE(R \backslash H,U)) \backslash
BD(G) \backslash
(
BD(HCLOSE(H,U)) \backslash 
BD(G)
)
\\
&=&
BD(HCLOSE(R \backslash H,U)) \backslash
BD(G) \backslash
BD(HCLOSE(H,U))
\end{eqnarray*}

The last step is a consequence of the equality between
{} $A \backslash B \backslash C$
and
{} $A \backslash B \backslash (C \backslash B)$.
Since $A \backslash B$ does contain any element of $B$, subtracting $C$ or
subtracting $C$ without the elements of $B$ is the same.

This reformulation shows that the second necessary condition is the same as $HL
\subseteq \emptyset$. The only subset of an empty set is itself: $HL =
\emptyset$. Since $HL$ does not depend on $IT$, its emptiness can be checked
before entering the loop over the possible sets $IT$.

The first condition is also expressible in terms of $IB$ and $HL$. The first
step is to reformulate the union by placing the intersection in one of the two
sets only: $A \cup B$ is the same as
{} $((A \backslash B) \cup (A \cap B)) \cup
{}  ((B \backslash A) \cup (A \cap B))$;
the duplication of $A \cap B$ is not necessary, making this union the same as
{} $(A \backslash B) \cup (A \cap B) \cup (B \backslash A)$.
This is the same as
{} $A \cup (B \backslash A)$.
The first necessary conditions is reformulated using this property.

\begin{eqnarray*}
&
BD(HCLOSE(H,U)) \cup
(BD(HCLOSE(R \backslash H,U))
\backslash
BD(HCLOSE(H,U)))
& \\
& \subseteq
BD(G) \cup
BD(IT) \hfill
\end{eqnarray*}

The union of two sets is contained in the union of $BD(G)$ and
$BD(IT)$. Stated differently, the variables in the two sets that are not in
$BD(G)$ are in $BD(IT)$.

\begin{eqnarray*}
&
\begin{array}{l}
(BD(HCLOSE(H,U)) \backslash BD(G))
\cup \\
(BD(HCLOSE(R \backslash H,U))
\backslash
BD(HCLOSE(H,U))
\backslash BD(G))
\end{array}
& \\
& \subseteq
BD(IT)
\end{eqnarray*}

The two sets in the union are exactly $IB$ and $HL$. The first necessary
condition becomes $IB \cup HL \subseteq BD(IT)$. Since it involves $IT$, it
can only be checked inside the loop. But the loop is not entered at all if $HL$
is not empty. Therefore, it could be reformulated as $IB \subseteq BD(IT)$.
The program checks the condition including $HL$ to allow disabling the other
necessary condition $HL = \emptyset$ for testing.

\

Both necessary conditions involve $HEADS(B,F)$, but the program never
calculates it. Rather, it exploits Lemma~\ref{heads-h} by replacing
$HEADS(B,F)$ with
{} $RCN(B,F) \backslash \{x \mid \exists B' ~.~ B' \rightarrow x \in G\}$.
The lemma proves that either they coincide or the autoreconstruction algorithm
fails and $F$ is not single-head equivalent. The replacing set is not always
the same as the replaced set, but can take its place when checking the two
necessary conditions. This is proved by looking at the two cases separately:

\begin{itemize}

\item if $HEADS(B,F)$ and
{} $RCN(B,F) \backslash \{x \mid \exists B' ~.~ B' \rightarrow x \in G\}$
coincide the two conditions are unchanged by replacing the former with the
latter;

\item if $HEADS(B,F)$ and
{} $RCN(B,F) \backslash \{x \mid \exists B' ~.~ B' \rightarrow x \in G\}$
differ the two conditions may change when replacing the former with the latter;
this is not a problem because these conditions are only necessary; if they are
made true, the program proceeds as if it did not check them, and fails as
proved by Lemma~\ref{heads-h}; if they are made false they make the program
fail immediately. In both cases the program fails sooner or later. This is
correct behavior because $F$ is not single-head equivalent as proved by
Lemma~\ref{heads-h}.

\end{itemize}

\subsubsection{The second necessary condition}

The aim of the inner loop of the algorithm is to find a set $IT$ that imitates
a part of $F$. A necessary condition is to optimistically assume that
everything that could possibly be entailed by $IT$ is entailed: the set of its
heads $H$. Since $IT$ is used with $G$, these heads are also plugged into $G$
for further entailments, producing $RCN(B \cup H,G)$. The union $H \cup RCN(B
\cup H,G)$ bounds the variables that $B$ can imply in $G \cup IT$. If it does
not include all of $RCN(B,F)$ in spite of the optimistic assumption, $G \cup
IT$ does not imitate $F$.

This condition only involves the heads of $IT$, not all of it. By construction,
these heads are exactly $HEADS(B,F)$. They are the same for all sets $IT$. As a
result, the condition can be checked before entering the loop over the possible
candidate sets $IT$.

The complication is that $RCN()$ is not the set of variables that can be
entailed. That is $BCN()$. Rather, it is the set of real consequences, the
variables that are entailed not just because they are in $B$. This is why a
formal proof is required.

\begin{lemma}
\label{optimistic}

If $H$ is the set of heads of the formula $E$, then
{} $RCN(B,G \cup E) \subseteq H \cup RCN(B \cup H,G)$
holds for every formula $G$ that contains no tautology.

\end{lemma}

\proof Let $x$ be a variable of $RCN(B,G \cup E)$. By definition, $G \cup E$
entails $(BCN(B,G \cup E) \backslash \{x\}) \rightarrow x$. Let $B'= BCN(B,G
\cup E) \backslash \{x\}$ be the body of this clause. The entailment turns into
$G \cup E \models B' \rightarrow x$, with $x \not\in B'$.

By Lemma~\ref{set-implies-set}, it holds $G \cup E \models B' \rightarrow B''$
for some clause $B'' \rightarrow x \in G \cup E$. If this clause is in $E$ its
head $x$ is in $H$ by the premise of the lemma and the claim is proved.

Otherwise, $B'' \rightarrow x$ is in $G$. In this case, $x$ is proved to belong
to $RCN(B \cup H,G)$.

Since $B'$ is $BCN(B,G \cup E) \backslash \{x\}$, it is a subset of $BCN(B,G
\cup E)$. Therefore, all its variables are entailed by $G \cup E \cup B$.
Equivalently, $G \cup E \models B \rightarrow B'$. With $G \cup E \models B'
\rightarrow B''$, it implies $G \cup E \models B \rightarrow B''$.

The head of every clause of $E$ is in $H$ since $H$ is the set of heads of $E$.
As a result, $H \models E$. Together with $G \cup E \models B \rightarrow B''$,
it implies $G \cup H \models B \rightarrow B''$. This entailment can be
rewritten as $G \models (B \cup H) \rightarrow B''$: every variable of $B''$ is
implied by $B \cup H$ according to $G$. Therefore, $B'' \subseteq BCN(B \cup
H,G)$.

Since $G$ contains no tautology by assumption, and $B'' \rightarrow x$ is one
of its clauses, it is not a tautology: $x \not\in B''$. The containment
{} $B'' \subseteq BCN(B \cup H,G)$
strengthens to
{} $B'' \subseteq BCN(B \cup H,G) \backslash \{x\}$.
Since $B'' \rightarrow x$ is in $G$, it is also entailed by it:
{} $G \models B'' \rightarrow x$.
By monotonicity, the body $B''$ of this clause can be replaced by any of its
supersets like $BCN(B \cup H,G) \backslash \{x\}$, leading to
{} $G \models (BCN(B \cup H,G) \backslash \{x\}) \rightarrow x$,
which defines $x \in RCN(B \cup H,G)$.~\qed

A condition is necessary if it follows from $SCL(B,F) \cup IT \models BCL(B,F)$
or an equivalent formulation of it, like $UCL(B,F) \equiv UCL(B,G \cup IT)$.

\begin{lemma}

If $UCL(B,F) \equiv UCL(B,G \cup IT)$
then
{} $RCN(B,F) \subseteq HEADS(B,F) \cup RCN(B \cup HEADS(B,F),F)$.

\end{lemma}

\proof Lemma~\ref{ucl-same-rcn} proves that $UCL(B,F) \equiv UCL(B,G \cup IT)$
implies $RCN(B,F) = RCN(B,G \cup IT)$. Lemma~\ref{optimistic} tells that the
second formula of the equality $RCN(B,G \cup IT)$ is a subset of $H \cup RCN(B
\cup H,G)$ where $H$ is the set of heads of $IT$. The heads of $IT$ are
$HEADS(B,F)$. A consequence is $RCN(B,F) \subseteq HEADS(B,F) \cup RCN(B \cup
HEADS(B,F),G)$.~\qed

When using
{} $H = RCN(B,F) \backslash \{x \mid \exists B' . B' \rightarrow x \in G\}$
as the heads of $IT$, Lemma~\ref{heads-h} applies: if $F$ is single-head
equivalent, then $H$ coincides with $HEADS(B,F)$ by Lemma~\ref{heads-h}. As a
result,
{} $RCN(B,F) \subseteq H \cup RCN(B \cup HEADS(B,F),F)$
is another consequence of $F$ being single-head equivalent. If it does not
hold, $F$ is not single-head equivalent and the algorithm fails as required.

The set $H \cup RCN(B \cup HEADS(B,F),G)$ is called {\em maxit} in the
{\tt reconstruct.py} program because it is an upper bound to the set of
consequences achievable by adding whichever set of clauses $IT$ with heads
$HEADS(B,F)$ to $G$. Yet, it does not involve $IT$ itself. Therefore, it can be
checked once for all before entering the loop rather than once for each
iteration.

\subsubsection{The third necessary condition}

Since $G \cup IT$ aims at imitating $F$ on the sets $B'$ that are equivalent to
$B$, it must have the same consequences. A specific case is already proved:
$RCN(B,G \cup IT) = RCN(B,F)$ as proved by Lemma~\ref{ucl-same-rcn}. This
condition extends from $B$ to every equivalent set:
{} $RCN(B',G \cup IT) = RCN(B,F)$ for every $B' \equiv_F B$.
This is mostly proved by the following lemma.

\begin{lemma}
\label{equivalent-same-rcn}

If $B \equiv_F B'$ then $RCN(B,F) = RCN(B',F)$.

\end{lemma}

\proof The definition of $B \equiv_F B'$ is $F \models B \equiv B'$. This is
equivalent to $F \cup B \equiv F \cup B'$. The definition
{} $BCN(B,F) = \{x \mid F \models B \rightarrow x\}$
can be rewritten as
{} $BCN(B,F) = \{x \mid F \cup B \models x\}$.
Since $F \cup B$ is equivalent to $F \cup B'$, this set is equal to
{} $\{x \mid F \cup B' \models x\}$,
or $BCN(B',F)$. This proves $BCN(B,F) = BCN(B',F)$.

As a result,
{} $RCN(B,F) = \{x \mid F \models (BCN(B,F) \backslash \{x\}) \rightarrow x\}$
can be rewritten as
{} $\{x \mid F \models (BCN(B',F) \backslash \{x\}) \rightarrow x\}$,
the definition of $RCN(B',F)$.~\qed

\iffalse

Lemma~\ref{scl-it-g-it}:
$SCL(B,F) \cup IT \models BCL(B,F)$ is equivalent to
$UCL(B,G \cup IT) \models BCL(B,F)$

Lemma~\ref{ucl-bcl-ucl}:
$UCL(B,G \cup IT) \models BCL(B,F)$ is equivalent to
$UCL(B,G \cup IT) \equiv UCL(B,F)$

Lemma~\ref{ucl-same-rcn}:
$UCL(B,G \cup IT) \equiv UCL(B,F)$
$RCN(B,F) = RCN(B,G \cup IT)$.

\fi

The goal is a necessary condition to the dominant operation of the loop, the
check
{} $SCL(B,F) \cup IT \models BCL(B,F)$.
Lemma~\ref{scl-it-g-it} and Lemma~\ref{ucl-bcl-ucl} prove it equivalent to
{} $UCL(B,G \cup IT) \equiv UCL(B,F)$,
which Lemma~\ref{ucl-same-rcn} proves to imply
{} $RCN(B,F) = RCN(B,G \cup IT)$.
Transitivity with
{} $RCN(B',G \cup IT) = RCN(B,G \cup IT)$
would prove the claim
{} $RCN(B',G \cup IT) = RCN(B,F)$.
Unfortunately, the premise
{} $RCN(B',G \cup IT) = RCN(B,G \cup IT)$
is only proved by Lemma~\ref{equivalent-same-rcn} if $B \equiv_{G \cup IT} B'$
holds, but the assumption is $B \equiv_F B'$. The following lemma provides this
missing bit.

\begin{lemma}
\label{ucl-same-equiv}

If $UCL(B,G \cup IT) \equiv UCL(B,F)$ then
$B' \equiv_F B$ is equivalent to $B' \equiv_{G \cup IT} B$.

\end{lemma}

\proof If $G \cup IT$ entails $B' \equiv B$ then $F$ entails it as well because
Lemma~\ref{f-implies-g-it} proves $F \models G \cup IT$.

The converse is now proved: if $F \models B' \equiv B$ then $G \cup IT \models
B' \equiv B$.

Lemma~\ref{ucl-only} applies to $F \models B \rightarrow B$ and $F \models B
\rightarrow b$ for every $b \in B'$, proving $UCL(B,F) \models B \rightarrow b$
for every $b \in B'$, which implies $UCL(B,F) \models B \rightarrow B'$. It
also applies to $F \models B \rightarrow B'$ and $F \models B' \rightarrow b$
for every $b \in B$, proving $UCL(B,F) \models B' \rightarrow b$, which implies
$UCL(B,F) \models B' \rightarrow B$. The conclusion is $UCL(B,F) \models B
\equiv B'$.

Since $UCL(B,F)$ is equivalent to $UCL(B,G \cup IT)$ by assumption, also
$UCL(B,G \cup IT) \models B' \equiv B$ holds. Since $UCL(B,G \cup IT)$ is by
definition a subset of $G \cup IT$, the claim $G \cup IT \models B' \equiv B$
follows.~\qed

The third necessary condition can now be proved.

\begin{lemma}

If $UCL(B,G \cup IT) \equiv UCL(B,F)$, then $RCN(B',G \cup IT) = RCN(B,F)$
holds for every $B'$ such that $B' \equiv_F B$.

\end{lemma}

\proof Given the premise $UCL(B,G \cup IT) \equiv UCL(B,F)$,
Lemma~\ref{ucl-same-rcn} proves $RCN(B,F) = RCN(B,G \cup IT)$.
Lemma~\ref{ucl-same-equiv} proves that $B' \equiv_F B$ implies $B \equiv_{G
\cup IT} B'$. Lemma~\ref{equivalent-same-rcn} applies: $RCN(B,G \cup IT) =
RCN(B',G \cup IT)$. The claim $RCN(B,F) = RCN(B',G \cup IT)$ follows by
transitivity.~\qed

Lemma~\ref{scl-it-g-it} and Lemma~\ref{ucl-bcl-ucl} prove that
{} $SCL(B,F) \cup IT \models BCL(B,F)$
is equivalent to
{} $UCL(B,G \cup IT) \equiv UCL(B,F)$,
which implies $RCN(B',G \cup IT) = RCN(B,F)$ for every $B'$ such that $B'
\equiv_F B$. This is a necessary condition. It is used like the others: in
reverse. If $RCN(B,F) \not= RCN(B',G \cup IT)$ for some $B'$ that is equivalent
to $B$, any further processing of $IT$ is unnecessary since $SCL(B,F) \cup IT
\models BCL(B,F)$ is bound to fail.

From the computational point of view, $RCN(B,F)$ is already known from the very
beginning. The second set $RCN(B',G \cup IT)$ can be easily determined given a
body $B'$ of a clause in $BCL(B,F)$. The problem is the possibly large number
of such sets $B'$. Restricting to
$HCLOSE(HEADS(B,F),UCL(B,F))$ or even to
$MINBODIES(HCLOSE(HEADS(B,F),UCL(B,F)),UCL(B,F) \cap SCL(B,F))$ still leaves
this condition necessary: weaker but easier to check. The {\tt singlehead.py}
program uses this restriction in the {\tt noteq} loop.

\subsection{Too many optimizations, too little time}

Some further directions for improvement to the algorithm have not been
investigated enough to prove their correctness and are left unimplemented.

\

The first undeveloped optimization is turning the third necessary condition
into a sufficient condition. It looks like checking $RCN(B',G \cup IT) =
RCN(B,F)$ for all $B' \in HCLOSE(HEADS(B,F),UCL(B,F) \cap SCL(B,F))$ is not
just necessary to $SCL(B,F) \cup IT \models BCL(B,F)$ but also sufficient. If
so, the following check
{} $HCLOSE(RCN(B,F),UCL(B,F)) = HCLOSE(RCN(B,G \cup IT),UCL(B,G \cup IT))$
is redundant.

\

The second undeveloped optimization is a simplification of the check
{} $HCLOSE(RCN(B,F),UCL(B,F)) = HCLOSE(RCN(B,G \cup IT),UCL(B,G \cup IT))$.
A subset of $HCLOSE(RCN(B,F),UCL(B,F))$ is already known:
$HCLOSE(HEADS(B,F),UCL(B,F))$ is computed to determine the possible sets $IT$
to test. Restricting the check to
{} $HCLOSE(HEADS(B,F),UCL(B,F)) = HCLOSE(HEADS(B,F),UCL(B,G \cup IT))$
saves time both because the first formula is already known and because the
second is easier to compute due to the smaller first argument.

This is not correct in general, as various examples in {\tt bnotheads.py}
show. One of them is the following.

\begin{eqnarray*}
F &=& \{ab \equiv ef, a \equiv c\} \\
&=&
\{ab \rightarrow e, ab \rightarrow f, ef \rightarrow a, ef \rightarrow b,
a \rightarrow c, c \rightarrow a\}
\end{eqnarray*}

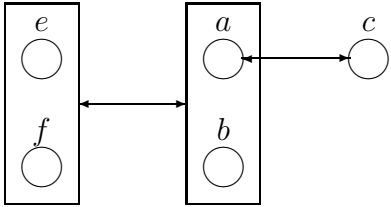
\setlength{\unitlength}{5000sp}%
\begingroup\makeatletter\ifx\SetFigFont\undefined%
\gdef\SetFigFont#1#2#3#4#5{%
  \reset@font\fontsize{#1}{#2pt}%
  \fontfamily{#3}\fontseries{#4}\fontshape{#5}%
  \selectfont}%
\fi\endgroup%
\begin{picture}(1921,1014)(4669,-4393)
{\color[rgb]{0,0,0}\thinlines
\put(5761,-3661){\circle{202}}
}%
{\color[rgb]{0,0,0}\put(5761,-4201){\circle{202}}
}%
{\color[rgb]{0,0,0}\put(6481,-3661){\circle{202}}
}%
{\color[rgb]{0,0,0}\put(4861,-3661){\circle{202}}
}%
{\color[rgb]{0,0,0}\put(4861,-4201){\circle{202}}
}%
{\color[rgb]{0,0,0}\put(5851,-3661){\vector(-1, 0){  0}}
\put(5851,-3661){\vector( 1, 0){540}}
}%
{\color[rgb]{0,0,0}\put(5581,-4381){\framebox(360,990){}}
}%
{\color[rgb]{0,0,0}\put(4681,-4381){\framebox(360,990){}}
}%
{\color[rgb]{0,0,0}\put(5041,-3886){\vector(-1, 0){  0}}
\put(5041,-3886){\vector( 1, 0){540}}
}%
\put(5761,-3526){\makebox(0,0)[b]{\smash{{\SetFigFont{12}{24.0}
{\rmdefault}{\mddefault}{\updefault}{\color[rgb]{0,0,0}$a$}%
}}}}
\put(5761,-4066){\makebox(0,0)[b]{\smash{{\SetFigFont{12}{24.0}
{\rmdefault}{\mddefault}{\updefault}{\color[rgb]{0,0,0}$b$}%
}}}}
\put(6481,-3526){\makebox(0,0)[b]{\smash{{\SetFigFont{12}{24.0}
{\rmdefault}{\mddefault}{\updefault}{\color[rgb]{0,0,0}$c$}%
}}}}
\put(4861,-4066){\makebox(0,0)[b]{\smash{{\SetFigFont{12}{24.0}
{\rmdefault}{\mddefault}{\updefault}{\color[rgb]{0,0,0}$f$}%
}}}}
\put(4861,-3526){\makebox(0,0)[b]{\smash{{\SetFigFont{12}{24.0}
{\rmdefault}{\mddefault}{\updefault}{\color[rgb]{0,0,0}$e$}%
}}}}
\end{picture}%
\nop{
+---+    +---+
| e |    | a<--->c
|   |<-->|   |
| f |    | b |
+---+    +---+
}

Since $SCL(B,F)$ is $\{a \rightarrow c, c \rightarrow a\}$ when $B = \{a,b\}$,
the variable $a$ is not in $SFREE(B,F)$, and therefore not in $HEADS(B,F)$.
Therefore, $ef \rightarrow a$ is never in $IT$. Yet, its absence goes
undetected because it is not in $HCLOSE(HEADS(B,F),UCL(B,F))$ either.

The third necessary condition catches the anomaly: $a$ is in $RCN(B,F)$ but not
in $RCN(\{e,f\},G \cup IT)$ since no $IT$ contains a clause of head $a$. This
is however specific to this case, as the following counterexample still from
{\tt bnotheads.py} shows.

\begin{eqnarray*}
F &=& \{ab \equiv ef, a \equiv c, b \equiv d) \\
&=& \{ab \rightarrow e, ab \rightarrow f, ef \rightarrow a, ef \rightarrow b,
a \rightarrow c, c \rightarrow a, b \rightarrow d, d \rightarrow b\}
\end{eqnarray*}

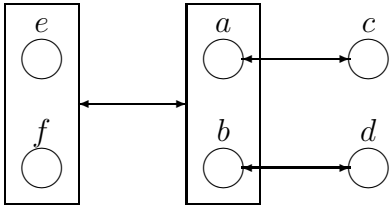
\setlength{\unitlength}{5000sp}%
\begingroup\makeatletter\ifx\SetFigFont\undefined%
\gdef\SetFigFont#1#2#3#4#5{%
  \reset@font\fontsize{#1}{#2pt}%
  \fontfamily{#3}\fontseries{#4}\fontshape{#5}%
  \selectfont}%
\fi\endgroup%
\begin{picture}(1921,1014)(4669,-4393)
{\color[rgb]{0,0,0}\thinlines
\put(5761,-3661){\circle{202}}
}%
{\color[rgb]{0,0,0}\put(5761,-4201){\circle{202}}
}%
{\color[rgb]{0,0,0}\put(6481,-3661){\circle{202}}
}%
{\color[rgb]{0,0,0}\put(4861,-3661){\circle{202}}
}%
{\color[rgb]{0,0,0}\put(4861,-4201){\circle{202}}
}%
{\color[rgb]{0,0,0}\put(6481,-4201){\circle{202}}
}%
{\color[rgb]{0,0,0}\put(5851,-3661){\vector(-1, 0){  0}}
\put(5851,-3661){\vector( 1, 0){540}}
}%
{\color[rgb]{0,0,0}\put(5581,-4381){\framebox(360,990){}}
}%
{\color[rgb]{0,0,0}\put(4681,-4381){\framebox(360,990){}}
}%
{\color[rgb]{0,0,0}\put(5041,-3886){\vector(-1, 0){  0}}
\put(5041,-3886){\vector( 1, 0){540}}
}%
{\color[rgb]{0,0,0}\put(5851,-4201){\vector(-1, 0){  0}}
\put(5851,-4201){\vector( 1, 0){540}}
}%
\put(5761,-3526){\makebox(0,0)[b]{\smash{{\SetFigFont{12}{24.0}
{\rmdefault}{\mddefault}{\updefault}{\color[rgb]{0,0,0}$a$}%
}}}}
\put(5761,-4066){\makebox(0,0)[b]{\smash{{\SetFigFont{12}{24.0}
{\rmdefault}{\mddefault}{\updefault}{\color[rgb]{0,0,0}$b$}%
}}}}
\put(6481,-3526){\makebox(0,0)[b]{\smash{{\SetFigFont{12}{24.0}
{\rmdefault}{\mddefault}{\updefault}{\color[rgb]{0,0,0}$c$}%
}}}}
\put(4861,-4066){\makebox(0,0)[b]{\smash{{\SetFigFont{12}{24.0}
{\rmdefault}{\mddefault}{\updefault}{\color[rgb]{0,0,0}$f$}%
}}}}
\put(4861,-3526){\makebox(0,0)[b]{\smash{{\SetFigFont{12}{24.0}
{\rmdefault}{\mddefault}{\updefault}{\color[rgb]{0,0,0}$e$}%
}}}}
\put(6481,-4066){\makebox(0,0)[b]{\smash{{\SetFigFont{12}{24.0}
{\rmdefault}{\mddefault}{\updefault}{\color[rgb]{0,0,0}$d$}%
}}}}
\end{picture}%
\nop{
+---+    +---+
| e |    | a<--->c
|   |<-->|   |
| f |    | b<--->d
+---+    +---+
}

Again, $a \not\in HEADS(B,F)$ causes $ef \rightarrow a$ not to be in
$HCLOSE(HEADS(B,F),UCL(B,F))$. However, $b$, $c$ and $d$ are not in
$HEADS(B,F)$ either, making $ef \rightarrow b$, $ef \rightarrow c$, $ef
\rightarrow d$ not being in $HCLOSE(HEADS(B,F),UCL(B,F))$. As a result,
$\{e,f\}$ is not a body of $HCLOSE(HEADS(B,F),UCL(B,F))$. The equality
$RCN(\{e,f\},G \cup IT) = RCN(B,F)$ is not even checked.

An undetected inequality like this is due to some sets $B'$ not being
recognized as equivalent to $B$. In turns, this seems to be only possible if $B
\cap HEADS(B,F) = \emptyset$, but this is yet to be proved. If proved, the
optimization can be employed if $B \cap HEADS(B,F) \not= \emptyset$, resorting
to the general condition
{} $HCLOSE(RCN(B,F),UCL(B,F)) = HCLOSE(RCN(B,G \cup IT),UCL(B,G \cup IT))$
otherwise.

\

The third undeveloped optimization is that associating a body to a variable $x
\in HEADS(B,F)$ is easy if $x$ is not in any of the possible bodies of $IT$.
This condition is easy to check: $x \not\in \bigcup T$.

All clauses in $ITERATION(B,F)$ are in $BCL(B,F)$ by definition: they are
nontautological clauses $B' \rightarrow x$ entailed by $F$ that meet $B
\equiv_F B'$. When building a clause for the head $x$, all acceptable bodies
are equivalent to $B$ according to $F$, and are therefore all equivalent to
each other. Since they are equivalent, they are the same. Any one will do. Even
just $B$.

Why the requirement $x \not\in \bigcup T$, then? Because $B \equiv_F B'$ is
equivalence according to $F$, not to $G \cup IT$. Equivalence according to $G
\cup IT$ is still to be achieved. The following formula (in {\tt outloop.py})
clarifies this point.

\[
F = \{a \rightarrow b, b \rightarrow c, c \rightarrow a,
      b \rightarrow x, c \rightarrow x\}
\]

The heads for $B = \{a\}$ are $HEADS(B,F) = \{a, b, c, x\}$. The task is to
produce a body for each of these variables. The simplified rule lets $B$ be the
body for all.

This is correct for $x$: the clause $a \rightarrow x$ is a valid choice for a
clause with head $x$. It is in the equivalent single-head formula $F'$, for
example:

\[
F' = \{a \rightarrow x, a \rightarrow b, b \rightarrow c, c \rightarrow x\}
\]

The same choice does not work for $b$ and $c$: no single-head formula
equivalent to $F$ contains both $a \rightarrow b$ and $a \rightarrow c$. For
example, the latter cannot replace $b \rightarrow c$ in $F'$ even if their
bodies are equivalent because the equivalence $a \equiv_F b$ extends to $F'$
only as consequence of $b \rightarrow c$.

This mechanism does not work because it relies on body equivalences that are
not achieved yet. Saying that $B \rightarrow x$ is as good as $B' \rightarrow
x$ because $B$ is equivalent to $B'$ anyway is incorrect because $B'
\rightarrow x$ may be necessary to ensure this equivalence.

This is not the case for $x$. Visually, $x$ is outside any loop.

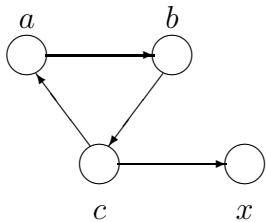
\setlength{\unitlength}{5000sp}%
\begingroup\makeatletter\ifx\SetFigFont\undefined%
\gdef\SetFigFont#1#2#3#4#5{%
  \reset@font\fontsize{#1}{#2pt}%
  \fontfamily{#3}\fontseries{#4}\fontshape{#5}%
  \selectfont}%
\fi\endgroup%
\begin{picture}(1298,1110)(5292,-4486)
{\color[rgb]{0,0,0}\thinlines
\put(5401,-3661){\circle{202}}
}%
{\color[rgb]{0,0,0}\put(6121,-3661){\circle{202}}
}%
{\color[rgb]{0,0,0}\put(5761,-4201){\circle{202}}
}%
{\color[rgb]{0,0,0}\put(6481,-4201){\circle{202}}
}%
{\color[rgb]{0,0,0}\put(5491,-3661){\vector( 1, 0){540}}
}%
{\color[rgb]{0,0,0}\put(6076,-3751){\vector(-3,-4){270}}
}%
{\color[rgb]{0,0,0}\put(5716,-4111){\vector(-3, 4){270}}
}%
{\color[rgb]{0,0,0}\put(5851,-4201){\vector( 1, 0){540}}
}%
\put(5401,-3526){\makebox(0,0)[b]{\smash{{\SetFigFont{12}{24.0}
{\rmdefault}{\mddefault}{\updefault}{\color[rgb]{0,0,0}$a$}%
}}}}
\put(6121,-3526){\makebox(0,0)[b]{\smash{{\SetFigFont{12}{24.0}
{\rmdefault}{\mddefault}{\updefault}{\color[rgb]{0,0,0}$b$}%
}}}}
\put(5761,-4471){\makebox(0,0)[b]{\smash{{\SetFigFont{12}{24.0}
{\rmdefault}{\mddefault}{\updefault}{\color[rgb]{0,0,0}$c$}%
}}}}
\put(6481,-4471){\makebox(0,0)[b]{\smash{{\SetFigFont{12}{24.0}
{\rmdefault}{\mddefault}{\updefault}{\color[rgb]{0,0,0}$x$}%
}}}}
\end{picture}%
\nop{
a ---> b ---> c ---> x
^             |
|             |
+-------------+
}

Technically, if $x$ is not in $\bigcup T$ it cannot contribute to the
equivalence of $B$ with any other set $B'$. Equivalence requires going from a
set to another and back. Instead, $x$ can only be entailed. It cannot entail.
Since $B' \rightarrow x$ is unnecessary to $B' \equiv B$, removing it does
invalidate this equivalence. Therefore, $B \rightarrow x$ still entails $B'
\rightarrow x$, constituting a valid replacement.

This optimization would use $H' = H \cap \bigcup T$ in place of $H$ and search
for a set $IT$ that meets
{} $HCLOSE(RCN(B,F) \cap \bigcup T,UCL(B,F)) =
{}  HCLOSE(RCN(B,G \cup IT) \cap \bigcup T,UCL(B,G \cup IT))$
instead of
{} $HCLOSE(RCN(B,F),UCL(B,F)) =
{}  HCLOSE(RCN(B,G \cup IT),UCL(B,G \cup IT))$.
On success, a clause $B \rightarrow x$ for every $x \in H \backslash H'$ is
added to $G$ along with $IT$.

Most of the other examples in this article and in the test directory of the
{\tt singlehead.py} program would not benefit from this optimization. This is
probably a distortion due to the nature of the technical analysis of the
problem and its algorithms. What makes single-head equivalence difficult from
the theoretical and implementation point of view is establishing equivalence of
sets $B'$ with $B$; this is why most examples revolve around this case.
Variables that are entailed but not in a loop are simple to deal with. Examples
that contain them are not significant.

Not significant to the technical analysis of the problem does not mean not
significant to the problem itself.

To the contrary, this optimization may dramatically increase performance. For
example, the algorithm is fast at telling that Formula~\ref{opposite} is not
single-head equivalent because it only needs to test the possible associations
of three heads $H = \{x, a, b\}$ with three bodies $\{b, d\}$, $\{a, c\}$ and
$\{x\}$. Adding easy clauses like $x \rightarrow y_i$ with $i=1,\ldots,n$
creates $n$ new heads, exponentially increasing the number of associations to
test. This is wasted time, since their bodies can for example be all $\{x\}$.

\begin{equation}
\label{opposite}
F = \{
	x \rightarrow a, a \rightarrow d,
	x \rightarrow b, b \rightarrow c,
	ac \rightarrow x,
	bd \rightarrow x
\}
\end{equation}

This formula is in the file {\tt disjointnotsingle.py}, the variant with the
two additional clauses in {\tt insignificant.py}.

\

The fourth unimplemented optimization would allow establishing that a formula
is not single-head equivalent without a single entailment test. It is based on
a counting argument, a calculation of the number of necessary clauses. The
following example illustrates its principle.

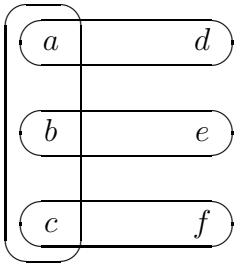
\setlength{\unitlength}{5000sp}%
\begingroup\makeatletter\ifx\SetFigFont\undefined%
\gdef\SetFigFont#1#2#3#4#5{%
  \reset@font\fontsize{#1}{#2pt}%
  \fontfamily{#3}\fontseries{#4}\fontshape{#5}%
  \selectfont}%
\fi\endgroup%
\begin{picture}(1149,1299)(7264,-6223)
\thinlines
{\color[rgb]{0,0,0}\put(7456,-5131){\oval(210,210)[bl]}
\put(7456,-5116){\oval(210,210)[tl]}
\put(8296,-5131){\oval(210,210)[br]}
\put(8296,-5116){\oval(210,210)[tr]}
\put(7456,-5236){\line( 1, 0){840}}
\put(7456,-5011){\line( 1, 0){840}}
\put(7351,-5131){\line( 0, 1){ 15}}
\put(8401,-5131){\line( 0, 1){ 15}}
}%
{\color[rgb]{0,0,0}\put(7456,-5581){\oval(210,210)[bl]}
\put(7456,-5566){\oval(210,210)[tl]}
\put(8296,-5581){\oval(210,210)[br]}
\put(8296,-5566){\oval(210,210)[tr]}
\put(7456,-5686){\line( 1, 0){840}}
\put(7456,-5461){\line( 1, 0){840}}
\put(7351,-5581){\line( 0, 1){ 15}}
\put(8401,-5581){\line( 0, 1){ 15}}
}%
{\color[rgb]{0,0,0}\put(7456,-6031){\oval(210,210)[bl]}
\put(7456,-6016){\oval(210,210)[tl]}
\put(8296,-6031){\oval(210,210)[br]}
\put(8296,-6016){\oval(210,210)[tr]}
\put(7456,-6136){\line( 1, 0){840}}
\put(7456,-5911){\line( 1, 0){840}}
\put(7351,-6031){\line( 0, 1){ 15}}
\put(8401,-6031){\line( 0, 1){ 15}}
}%
{\color[rgb]{0,0,0}\put(7381,-6106){\oval(210,210)[bl]}
\put(7381,-5041){\oval(210,210)[tl]}
\put(7546,-6106){\oval(210,210)[br]}
\put(7546,-5041){\oval(210,210)[tr]}
\put(7381,-6211){\line( 1, 0){165}}
\put(7381,-4936){\line( 1, 0){165}}
\put(7276,-6106){\line( 0, 1){1065}}
\put(7651,-6106){\line( 0, 1){1065}}
}%
\put(7501,-5161){\makebox(0,0)[b]{\smash{{\SetFigFont{12}{24.0}
{\rmdefault}{\mddefault}{\updefault}{\color[rgb]{0,0,0}$a$}%
}}}}
\put(8251,-5161){\makebox(0,0)[b]{\smash{{\SetFigFont{12}{24.0}
{\rmdefault}{\mddefault}{\updefault}{\color[rgb]{0,0,0}$d$}%
}}}}
\put(7501,-5611){\makebox(0,0)[b]{\smash{{\SetFigFont{12}{24.0}
{\rmdefault}{\mddefault}{\updefault}{\color[rgb]{0,0,0}$b$}%
}}}}
\put(8251,-5611){\makebox(0,0)[b]{\smash{{\SetFigFont{12}{24.0}
{\rmdefault}{\mddefault}{\updefault}{\color[rgb]{0,0,0}$e$}%
}}}}
\put(7501,-6061){\makebox(0,0)[b]{\smash{{\SetFigFont{12}{24.0}
{\rmdefault}{\mddefault}{\updefault}{\color[rgb]{0,0,0}$c$}%
}}}}
\put(8251,-6061){\makebox(0,0)[b]{\smash{{\SetFigFont{12}{24.0}
{\rmdefault}{\mddefault}{\updefault}{\color[rgb]{0,0,0}$f$}%
}}}}
\end{picture}%
\nop{
+--+
|+-----+
||a|  d|
|+-----+
|  |
|+-----+
||b|  e|
|+-----+
|  |
|+-----+
||c|  f|
|+-----+
+--+
}

A box stands for a set of variables that is equivalent to the others:
$\{a,b,c\} \equiv \{a,d\} \equiv \{b,e\} \equiv \{c,f\}$. The equivalence
between $\{a,b,c\}$ and $\{a,d\}$ is for example realized by $abc \rightarrow
d$, $ad \rightarrow b$ and $ad \rightarrow c$: each variable that is in a set
but not in the other is implied by the other. The file for this formula is {\tt
disjointemptynotsingle.py}.

The bodies of the formula are the equivalent sets; therefore, they are all
equivalent to each other. The first precondition chosen by the program could
for example be $\{a,b,c\}$. Regardless, $SCL(B,F)$ is empty. The {\tt
singlehead.py} program disproves the formula single-head equivalent testing
4096 combinations of heads and bodies.

% Among them, 1956 contains tautologies and 1996 are discarded because their
% bodies do not contain enough variables. Other 70 are discarded because
% $RCN(B,F)$ differs from $RCN(B,G \cup IT)$. The remaining ones are 50.

This computation can be skipped altogether by just counting the number of
necessary clauses. Since $\{a,d\}$ implies the other equivalent sets and only
the equivalent sets imply something, $G \cup IT$ must contain some clauses that
make $\{a,d\}$ directly imply at least another equivalent set: $\{a,b,c\}$,
$\{b,e\}$ or $\{c,f\}$. Each contains two variables not in $\{a,d\}$, requiring
two clauses. For example, $ad \rightarrow b$ and $ad \rightarrow c$ are
required to imply $\{b, c\}$ from $\{a,d\}$. They are irredundant as $a$ alone
does not imply anything in $F$, nor does $d$.

The same applies to the other three equivalent sets: two clauses each. The
total is eight clauses. They are all different: the clauses for different
equivalent sets are different because the equivalent sets all differ from each
other; the two clauses for the same equivalent set are different because they
have different heads. The eight clauses are all different.

The variables are only six; a single-head formula contains at most six clauses.
The formula is not single-head equivalent because no equivalent formula
contains less than eight clauses.

This mechanism generalizes to all formulae that contain equivalent sets of
variables. For each of them $B'$, the number of variables in the minimal
difference $B'' \backslash B' \backslash SFREE(B,F)$ to another equivalent set
$B''$ is counted. The variables not in $SFREE(B,F)$ are discarded, as
implicitly done on the example where $SCL(B,F) = \emptyset$. The total is the
minimal number of necessary clauses. If it is larger than the number of heads,
the formula is not single-head equivalent.

When determining the equivalent sets of variables, only the minimal ones count.
For example, the sets that matter in $F = \{a \rightarrow b, b \rightarrow a, a
\rightarrow c\}$ are only $\{a\}$ and $\{b\}$, not $\{a,c\}$ and $\{b,c\}$.

% If these minimally-equivalent sets are exponentially many, the formula cannot
% be single-head equivalent. This only applies to sets that are equivalent but
% only thanks to $SCL(B,F)$.

This condition is only necessary, as some formulae have less equivalent sets
than heads but are not single-head equivalent anyway. An example is the formula
{} $ab \equiv bc \equiv ca \equiv de \equiv ef \equiv fd$.
Each set can imply another with a single clause. For example, $\{c,a\}$ implies
$\{b,c\}$ thanks to $ca \rightarrow b$ only.

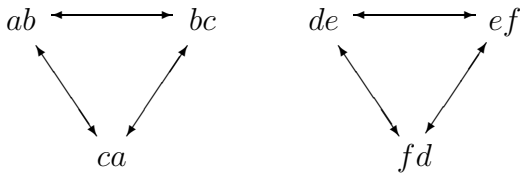
\setlength{\unitlength}{5000sp}%
\begingroup\makeatletter\ifx\SetFigFont\undefined%
\gdef\SetFigFont#1#2#3#4#5{%
  \reset@font\fontsize{#1}{#2pt}%
  \fontfamily{#3}\fontseries{#4}\fontshape{#5}%
  \selectfont}%
\fi\endgroup%
\begin{picture}(2430,840)(7636,-6001)
\thinlines
{\color[rgb]{0,0,0}\put(7726,-5386){\vector(-2, 3){  0}}
\put(7726,-5386){\vector( 2,-3){300}}
}%
{\color[rgb]{0,0,0}\put(8176,-5836){\vector(-2,-3){  0}}
\put(8176,-5836){\vector( 2, 3){300}}
}%
{\color[rgb]{0,0,0}\put(7801,-5236){\vector(-1, 0){  0}}
\put(7801,-5236){\vector( 1, 0){600}}
}%
{\color[rgb]{0,0,0}\put(9301,-5236){\vector(-1, 0){  0}}
\put(9301,-5236){\vector( 1, 0){600}}
}%
{\color[rgb]{0,0,0}\put(9226,-5386){\vector(-2, 3){  0}}
\put(9226,-5386){\vector( 2,-3){300}}
}%
{\color[rgb]{0,0,0}\put(9659,-5824){\vector(-2,-3){  0}}
\put(9659,-5824){\vector( 2, 3){300}}
}%
\put(8101,-5986){\makebox(0,0)[b]{\smash{{\SetFigFont{12}{24.0}
{\rmdefault}{\mddefault}{\updefault}{\color[rgb]{0,0,0}$ca$}%
}}}}
\put(7651,-5311){\makebox(0,0)[b]{\smash{{\SetFigFont{12}{24.0}
{\rmdefault}{\mddefault}{\updefault}{\color[rgb]{0,0,0}$ab$}%
}}}}
\put(9151,-5311){\makebox(0,0)[b]{\smash{{\SetFigFont{12}{24.0}
{\rmdefault}{\mddefault}{\updefault}{\color[rgb]{0,0,0}$de$}%
}}}}
\put(10051,-5311){\makebox(0,0)[b]{\smash{{\SetFigFont{12}{24.0}
{\rmdefault}{\mddefault}{\updefault}{\color[rgb]{0,0,0}$ef$}%
}}}}
\put(8551,-5311){\makebox(0,0)[b]{\smash{{\SetFigFont{12}{24.0}
{\rmdefault}{\mddefault}{\updefault}{\color[rgb]{0,0,0}$bc$}%
}}}}
\put(9601,-5986){\makebox(0,0)[b]{\smash{{\SetFigFont{12}{24.0}
{\rmdefault}{\mddefault}{\updefault}{\color[rgb]{0,0,0}$fd$}%
}}}}
\end{picture}%
\nop{
 ab ---- bc     de ---- ef
    \  /           \  /
     ca             fd
}

The counting condition is satisfied. Yet, the six clauses it counts as
necessary only realize the two loops of equivalences. Other clauses are
necessary to link the two, outnumbering the heads.

\

The final undeveloped variant is not an optimization of the algorithm but a way
to make it produce a meaningful result even if the input formula is not
single-head equivalent. At each step, the set of heads of the candidate set
$IT$ is $RCN(B,F)$ minus the heads of the formula under construction $G$. If
the formula is single-head equivalent, the result is equal to $HEADS(B,F)$.
Otherwise, the autoreconstruction algorithm fails as proved by
Lemma~\ref{heads-h}.

This mechanism presumes that a single-head formula equivalent to $F$ is the
only goal; if none exists, no output is necessary. This may not be the case. A
formula that contains just a pair of clauses with the same head is almost as
good as one containing none: forgetting requires only one true nondeterministic
choice, which only doubles the running time. Double is still polynomial.

The goal extends from finding a single-head formula to a formula containing few
duplicated heads. It is achieved by adding a clause $B' \rightarrow x$ for each
$x \in RCN(B,F) \backslash RCN(B,G \cup IT)$.

\section{Polynomial or \np-complete?}
\label{future}

The reconstruction algorithm requires a valid iteration function, which in turn
requires testing the possible association of some bodies to some heads. These
associations may be exponentially many. The algorithm takes exponential time if
so.

Is this specific to the reconstruction algorithm? Or is the problem itself that
requires exponential time to be solved?

While no proof of \np-hardness is found so far, at least the problem is not
harder than \np. The considered decision problem is: does the formula have an
equivalent formula that is single-head? Testing all equivalent formulae for
both equivalence and single-headness is not feasible even in nondeterministic
polynomial time since formulae may be arbitrarily large. Single-headness caps
that size, since it allows at most one clause for each head. A single-head
formula may at most contain a clause for each variable.

\begin{theorem}

Checking whether a formula is equivalent to a single-head formula is in \np.

\end{theorem}

\proof Equivalence to a single-head formula can be established by a
nondeterministic variant of the reconstruction algorithm, where $IT$ is not
looped over but is uniquely determined by guessing a body for each variable in
$HEADS(B,F)$. Technically, the loop ``for all $IT$...'' is replaced by
``nondeterministically assign an element of $T$ to each element of $H$''. The
resulting algorithm is polynomial.~\qed

Proving membership to \np\  only takes a short paragraph. Proving \np-hardness
is a completely different story. The rest of this section gives clues about it.

A proof of \np-hardness is a reduction from some other \np-hard problem to that
of telling whether $F$ is single-head equivalent. Given an instance of the
problem that is known \np-hard, it requires building a formula that is
single-head equivalent or not depending on that instance. This is generally
done by some sort of formula template, a scheme that has some parts left
partially unspecified. The specific \np-hard problem instance fills these so
that the resulting concrete formula is either single-head or not depending on
it.

While no such reduction is known, still something can be said about the formula
schemes that work and the ones that do not.

Lemma~\ref{disjoint-heads} tells that single-head equivalence requires two
things: first, $HEADS(B,F)$ and $HEADS(B',F)$ are disjoint whenever $B
\not\equiv_F B'$; second, $ITERATION(B,F)$ is single-head.

The first condition is polynomial. A reduction that only exploits it would
reduce a \np-hard problem to a polynomial one. Winning the million dollar
price at stake for it~\cite{cook-03}
% https://www.claymath.org/millennium-problems/p-vs-np-problem
would be certainly appreciated, but seems unlikely. Rather, it suggests that a
\np-hard proof is not based on the disjointness of heads.

The reduction, if any, is based on the other condition: the existence of a
valid iteration function such that $ITERATION(B,F)$ is single-head. This is
exactly where the autoreconstruction algorithm spends more time: testing all
associations of the bodies in $T$ to the heads in $H$. A reduction from the
propositional satisfiability of a formula $G$ to the single-head equivalence of
$F$ would probably hinge around a single, fixed set $B$:

\begin{itemize}

\item $HEADS(B',F)$ and $HEADS(B'',F)$ are disjoint for all $B' \not\equiv_F
B''$ regardless of $G$;

\item $SCL(B,F)$ is independent on $G$, or maybe only depends on the variables
of $G$;

\item $B$ is a fixed set of variables, maybe a singleton; maybe it is a set
that only depends on the variables of $G$;

\item $G$ is satisfiable if and only if a single-head subset of
$HCLOSE(HEADS(B,F),UCL(B,F))$ entails $BCL(B,F)$ with $SCL(B,F)$.

\end{itemize}

The core of the reduction is the last point: the satisfiability of $G$
corresponds to the existence of a single-head subset $IT$ of
$HCLOSE(HEADS(B,F),UCL(B,F))$ satisfying $SCL(B,F) \cup IT \models BCL(B,F)$,
where $B$ is fixed. The clauses of $F \cap SCL(B,F)$ may support this
construction; or maybe they are just not needed, and $F$ is entirely made of
clauses of $BCL(B,F)$. Whether $SCL(B,F)$ is necessary or not might further
restrict the range of possible reductions. The clauses of $SCL(B,F)$ and
$BCL(B,F)$ differ in an important way when it comes to ensuring $SCL(B,F) \cup
IT \models BCL(B,F)$, as explained in the next section.

\subsection{What do clauses do?}

While searching for a proof of \np-hardness for the single-head equivalence
problem, an observation emerged: clauses in $SCL(B,F)$ and in $ITERATION(B,F)$
differ in a fundamental point: which and how many clauses of $BCL(B,F)$ they
collaborate to imply.

All clauses of $BCL(B,F)$ have the form $B' \rightarrow x$ with $B' \equiv_F
B$. Apart the case where $B$ is only equivalent to itself, this means that $B'$
entails $B$ and all other sets equivalent to $B$. This is not possible if
$SCL(B,F) \cup ITERATION(B,F)$ does not contain at least a clause that can be
applied from $B'$: a clause whose body is contained in $B'$. This clause makes
$B'$ entail another equivalent set, which in turn entail another in the same
way. This clause may belong to $SCL(B,F)$ or to $BCL(B,F)$.

\begin{description}

\item[$B'' \rightarrow x \in SCL(B,F)$] may have a body $B''$ that is a subset
of multiple sets $B'$ that are equivalent to $B$; therefore, $B'' \rightarrow
x$ may be a starting point of implication from many differing $B'$;

\item[$B'' \rightarrow x \in ITERATION(B,F)$] implies that $B''$ is exactly one
of these sets that are equivalent to $B$; therefore, $B'' \rightarrow x$ may
only be the starting point of implications from $B' = B''$ itself.

\end{description}

The hard task for $ITERATION(B,F)$ is to realize all equivalences $B' \equiv B$
entailed by $F$. As explained in the previous sections, the clauses not
involved in implying such equivalences are easy to find. Equivalences are
mutual entailments and are therefore broken down into entailments like $B'
\rightarrow B''$ where both $B'$ and $B''$ are equivalent to $B$ according to
$F$. The difference between a clause of $SCL(B,F)$ and one of $ITERATION(B,F)$
is that the first can be used to realize many implications $B' \rightarrow
B''$, the second only the ones starting from its body.

The following example clarifies the difference:
{} $F = \{a \equiv b, ac \rightarrow d, ad \rightarrow e, ae \rightarrow c\}$
when $B=\{a,c\}$.

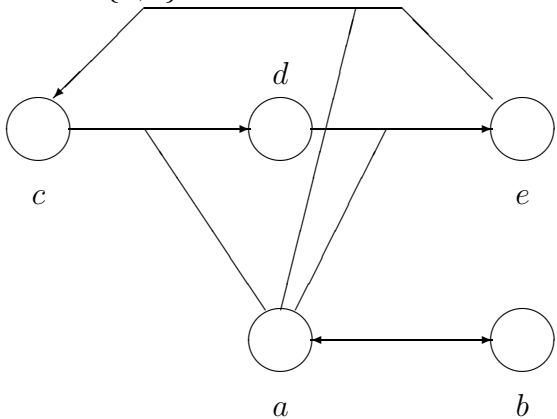
\setlength{\unitlength}{5000sp}%
\begingroup\makeatletter\ifx\SetFigFont\undefined%
\gdef\SetFigFont#1#2#3#4#5{%
  \reset@font\fontsize{#1}{#2pt}%
  \fontfamily{#3}\fontseries{#4}\fontshape{#5}%
  \selectfont}%
\fi\endgroup%
\begin{picture}(2752,2052)(3575,-4201)
{\color[rgb]{0,0,0}\thinlines
\put(3751,-2761){\circle{336}}
}%
{\color[rgb]{0,0,0}\put(4951,-2761){\circle{300}}
}%
{\color[rgb]{0,0,0}\put(6151,-2761){\circle{336}}
}%
{\color[rgb]{0,0,0}\put(4951,-3811){\circle{336}}
}%
{\color[rgb]{0,0,0}\put(6151,-3811){\circle{336}}
}%
{\color[rgb]{0,0,0}\put(3901,-2761){\vector( 1, 0){900}}
}%
{\color[rgb]{0,0,0}\put(5101,-2761){\vector( 1, 0){900}}
}%
{\color[rgb]{0,0,0}\put(6001,-2611){\line(-1, 1){450}}
\put(5551,-2161){\line(-1, 0){1275}}
\put(4276,-2161){\vector(-1,-1){450}}
}%
{\color[rgb]{0,0,0}\put(4876,-3661){\line(-2, 3){600}}
}%
{\color[rgb]{0,0,0}\put(5026,-3661){\line( 1, 2){450}}
}%
{\color[rgb]{0,0,0}\put(4951,-3661){\line( 1, 4){375}}
}%
{\color[rgb]{0,0,0}\put(5101,-3811){\vector(-1, 0){  0}}
\put(5101,-3811){\vector( 1, 0){900}}
}%
\put(4951,-4186){\makebox(0,0)[b]{\smash{{\SetFigFont{12}{24.0}
{\rmdefault}{\mddefault}{\updefault}{\color[rgb]{0,0,0}$a$}%
}}}}
\put(6151,-4186){\makebox(0,0)[b]{\smash{{\SetFigFont{12}{24.0}
{\rmdefault}{\mddefault}{\updefault}{\color[rgb]{0,0,0}$b$}%
}}}}
\put(3751,-3136){\makebox(0,0)[b]{\smash{{\SetFigFont{12}{24.0}
{\rmdefault}{\mddefault}{\updefault}{\color[rgb]{0,0,0}$c$}%
}}}}
\put(4951,-2536){\makebox(0,0)[b]{\smash{{\SetFigFont{12}{24.0}
{\rmdefault}{\mddefault}{\updefault}{\color[rgb]{0,0,0}$d$}%
}}}}
\put(6151,-3136){\makebox(0,0)[b]{\smash{{\SetFigFont{12}{24.0}
{\rmdefault}{\mddefault}{\updefault}{\color[rgb]{0,0,0}$e$}%
}}}}
\end{picture}%
\nop{
         +--------------+
         |              |
+--------+--------+     |
V                 |     |
c --+--> d --+--> e     |
    |        |          |
    |        |          |
    +--------+----------+
             |
             a <---> b
}

The clauses of strictly entailed bodies are $SCL(B,F) = \{a
\equiv b\}$. The bodies equivalent to $B$ are
{} $\{a,c\}$, $\{a,d\}$, $\{a,e\}$,
{} $\{b,c\}$, $\{b,d\}$, and $\{b,e\}$.
The clause $ad \rightarrow e \in BCL(B,F)$ can only be used when $B'=\{a,d\}$
is true. It only allows the direct implication $ad \rightarrow ae$. On the
other hand, the clause $a \rightarrow b \in SCL(B,F)$ entails $ac \rightarrow
bc$, $ad \rightarrow bd$ and $ae \rightarrow be$.

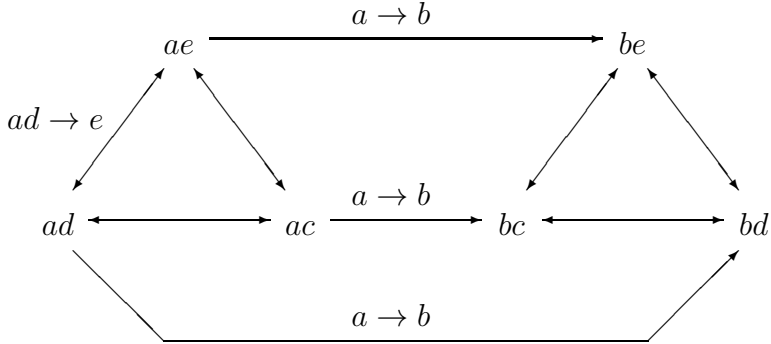
\setlength{\unitlength}{5000sp}%
\begingroup\makeatletter\ifx\SetFigFont\undefined%
\gdef\SetFigFont#1#2#3#4#5{%
  \reset@font\fontsize{#1}{#2pt}%
  \fontfamily{#3}\fontseries{#4}\fontshape{#5}%
  \selectfont}%
\fi\endgroup%
\begin{picture}(3480,1752)(4486,-4573)
\thinlines
{\color[rgb]{0,0,0}\put(4576,-3811){\vector(-3,-4){  0}}
\put(4576,-3811){\vector( 3, 4){450}}
}%
{\color[rgb]{0,0,0}\put(5176,-3211){\vector(-3, 4){  0}}
\put(5176,-3211){\vector( 3,-4){450}}
}%
{\color[rgb]{0,0,0}\put(4651,-3961){\vector(-1, 0){  0}}
\put(4651,-3961){\vector( 1, 0){900}}
}%
{\color[rgb]{0,0,0}\put(6826,-3811){\vector(-3,-4){  0}}
\put(6826,-3811){\vector( 3, 4){450}}
}%
{\color[rgb]{0,0,0}\put(7426,-3211){\vector(-3, 4){  0}}
\put(7426,-3211){\vector( 3,-4){450}}
}%
{\color[rgb]{0,0,0}\put(6901,-3961){\vector(-1, 0){  0}}
\put(6901,-3961){\vector( 1, 0){900}}
}%
{\color[rgb]{0,0,0}\put(5251,-3061){\vector( 1, 0){1950}}
}%
{\color[rgb]{0,0,0}\put(5851,-3961){\vector( 1, 0){750}}
}%
{\color[rgb]{0,0,0}\put(4576,-4111){\line( 1,-1){450}}
\put(5026,-4561){\line( 1, 0){2400}}
\put(7426,-4561){\vector( 1, 1){450}}
}%
\put(6151,-2986){\makebox(0,0)[b]{\smash{{\SetFigFont{12}{24.0}
{\rmdefault}{\mddefault}{\updefault}{\color[rgb]{0,0,0}$a \rightarrow b$}%
}}}}
\put(6151,-4486){\makebox(0,0)[b]{\smash{{\SetFigFont{12}{24.0}
{\rmdefault}{\mddefault}{\updefault}{\color[rgb]{0,0,0}$a \rightarrow b$}%
}}}}
\put(6151,-3886){\makebox(0,0)[b]{\smash{{\SetFigFont{12}{24.0}
{\rmdefault}{\mddefault}{\updefault}{\color[rgb]{0,0,0}$a \rightarrow b$}%
}}}}
\put(4501,-4036){\makebox(0,0)[b]{\smash{{\SetFigFont{12}{24.0}
{\rmdefault}{\mddefault}{\updefault}{\color[rgb]{0,0,0}$ad$}%
}}}}
\put(7951,-4036){\makebox(0,0)[b]{\smash{{\SetFigFont{12}{24.0}
{\rmdefault}{\mddefault}{\updefault}{\color[rgb]{0,0,0}$bd$}%
}}}}
\put(4726,-3511){\makebox(0,0)[rb]{\smash{{\SetFigFont{12}{24.0}
{\rmdefault}{\mddefault}{\updefault}{\color[rgb]{0,0,0}$ad \rightarrow e$}%
}}}}
\put(5101,-3136){\makebox(0,0)[b]{\smash{{\SetFigFont{12}{24.0}
{\rmdefault}{\mddefault}{\updefault}{\color[rgb]{0,0,0}$ae$}%
}}}}
\put(5701,-4036){\makebox(0,0)[b]{\smash{{\SetFigFont{12}{24.0}
{\rmdefault}{\mddefault}{\updefault}{\color[rgb]{0,0,0}$ac$}%
}}}}
\put(6751,-4036){\makebox(0,0)[b]{\smash{{\SetFigFont{12}{24.0}
{\rmdefault}{\mddefault}{\updefault}{\color[rgb]{0,0,0}$bc$}%
}}}}
\put(7351,-3136){\makebox(0,0)[b]{\smash{{\SetFigFont{12}{24.0}
{\rmdefault}{\mddefault}{\updefault}{\color[rgb]{0,0,0}$be$}%
}}}}
\end{picture}%
\nop{
                     a->b
             ae      --->      be
      ad->e /   \             /  \                  .
          ad -- ac   --->   bc -- bd
	             a->b
           |                      ^
           +----------------------+
	             a->b
}

\subsection{Counterexamples}

Lemma~\ref{disjoint-heads} implies that $HEADS(B,F)$ and $HEADS(B',F)$ are
always disjoint if $F$ does not entail $B \equiv B'$ and is single-head
equivalent. This is however only a necessary condition, as sufficiency also
requires the existence of a valid iteration function such that $ITERATION(B,F)$
is single-head.

Since $HEADS(B,F) \cap HEADS(B',F) = \emptyset$ is easy to check for fixed $B$
and $B'$, a proof of \np-hardness cannot be based on it. It should instead
hinge around the other condition. For example, it produces formulae that always
have disjoint $HEADS(B,F)$, but have a single-head $ITERATION(B,F)$ if the
given formula is satisfiable.

Such a reduction may result from generalizing one of the example formulae that
have disjoint $HEADS(B,F)$ but $ITERATION(B,F)$ is not single-head.

\begin{enumerate}

\item A set entails two equivalent sets; achieving equivalence requires them to
entail another equivalent set each, but this requires two clauses with the same
head.

Technically, $SCL(B,F)$ implies $A \rightarrow B$, $A \rightarrow C$ but no
other entailment among the minimal $F$-equivalent sets $A$, $B$, $C$ and $D$.
Some other entailments are required to make the four sets equivalent. For
example, to make $B$ equivalent to the other sets, it has to imply at least one
of them. Since the clauses of $SCL(B,F)$ do not suffice, some clauses from
$BCL(B,F)$ are needed. Such clauses have body $B$ because they are directly
applicable from $B$ (their body is a subset of $B$) and are not in $SCL(B,F)$
(their body is not a proper subset of $B$). At least one clause $B \rightarrow
x$ is necessary. For the same reason, $C \rightarrow y$ and $D \rightarrow z$
are also necessary, but cannot be entailed from single-head clauses.

%
%  two cycles of single variables, running in different directions 
%
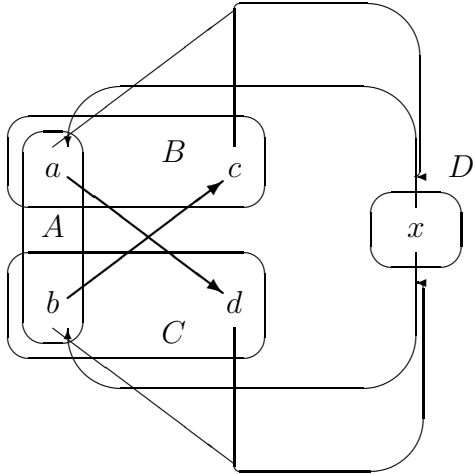
\setlength{\unitlength}{5000sp}%
\begingroup\makeatletter\ifx\SetFigFont\undefined%
\gdef\SetFigFont#1#2#3#4#5{%
  \reset@font\fontsize{#1}{#2pt}%
  \fontfamily{#3}\fontseries{#4}\fontshape{#5}%
  \selectfont}%
\fi\endgroup%
\begin{picture}(2277,2342)(7564,-6931)
\thinlines
{\color[rgb]{0,0,0}\put(7876,-6211){\vector( 0, 1){0}}
\put(8614,-6211){\oval(1476,600)[bl]}
\put(8614,-5836){\oval(1974,1350)[br]}
}%
{\color[rgb]{0,0,0}\put(8627,-5611){\oval(1948,1200)[tr]}
\put(8627,-5311){\oval(1502,600)[tl]}
\put(7876,-5311){\vector( 0,-1){0}}
}%
{\color[rgb]{0,0,0}\put(8926,-6886){\oval(450, 74)[bl]}
\put(8926,-6211){\oval(1424,1424)[br]}
\put(9601,-6211){\oval( 74,450)[tr]}
\put(9601,-5986){\vector(-1, 0){0}}
}%
{\color[rgb]{0,0,0}\put(9601,-5461){\vector(-1, 0){0}}
\put(9601,-5294){\oval( 40,334)[br]}
\put(8926,-5294){\oval(1390,1390)[tr]}
\put(8926,-4636){\oval(450, 74)[tl]}
}%
{\color[rgb]{0,0,0}\put(7756,-6181){\oval(210,210)[bl]}
\put(7756,-5341){\oval(210,210)[tl]}
\put(7846,-6181){\oval(210,210)[br]}
\put(7846,-5341){\oval(210,210)[tr]}
\put(7756,-6286){\line( 1, 0){ 90}}
\put(7756,-5236){\line( 1, 0){ 90}}
\put(7651,-6181){\line( 0, 1){840}}
\put(7951,-6181){\line( 0, 1){840}}
}%
{\color[rgb]{0,0,0}\put(9481,-5806){\oval(210,210)[bl]}
\put(9481,-5641){\oval(210,210)[tl]}
\put(9721,-5806){\oval(210,210)[br]}
\put(9721,-5641){\oval(210,210)[tr]}
\put(9481,-5911){\line( 1, 0){240}}
\put(9481,-5536){\line( 1, 0){240}}
\put(9376,-5806){\line( 0, 1){165}}
\put(9826,-5806){\line( 0, 1){165}}
}%
{\color[rgb]{0,0,0}\put(7681,-5506){\oval(210,210)[bl]}
\put(7681,-5266){\oval(210,210)[tl]}
\put(8746,-5506){\oval(210,210)[br]}
\put(8746,-5266){\oval(210,210)[tr]}
\put(7681,-5611){\line( 1, 0){1065}}
\put(7681,-5161){\line( 1, 0){1065}}
\put(7576,-5506){\line( 0, 1){240}}
\put(8851,-5506){\line( 0, 1){240}}
}%
{\color[rgb]{0,0,0}\put(7681,-6256){\oval(210,210)[bl]}
\put(7681,-5941){\oval(210,210)[tl]}
\put(8746,-6256){\oval(210,210)[br]}
\put(8746,-5941){\oval(210,210)[tr]}
\put(7681,-6361){\line( 1, 0){1065}}
\put(7681,-5836){\line( 1, 0){1065}}
\put(7576,-6256){\line( 0, 1){315}}
\put(8851,-6256){\line( 0, 1){315}}
}%
\thicklines
{\color[rgb]{0,0,0}\put(7876,-6061){\vector( 4, 3){768}}
}%
{\color[rgb]{0,0,0}\put(7876,-5461){\vector( 4,-3){768}}
}%
\thinlines
{\color[rgb]{0,0,0}\put(7801,-5311){\line( 4, 3){900}}
\put(8701,-4636){\line( 0,-1){675}}
}%
{\color[rgb]{0,0,0}\put(7801,-6211){\line( 4,-3){900}}
\put(8701,-6886){\line( 0, 1){675}}
}%
\put(7801,-5461){\makebox(0,0)[b]{\smash{{\SetFigFont{12}{24.0}
{\rmdefault}{\mddefault}{\updefault}{\color[rgb]{0,0,0}$a$}%
}}}}
\put(8701,-5461){\makebox(0,0)[b]{\smash{{\SetFigFont{12}{24.0}
{\rmdefault}{\mddefault}{\updefault}{\color[rgb]{0,0,0}$c$}%
}}}}
\put(7801,-6136){\makebox(0,0)[b]{\smash{{\SetFigFont{12}{24.0}
{\rmdefault}{\mddefault}{\updefault}{\color[rgb]{0,0,0}$b$}%
}}}}
\put(8701,-6136){\makebox(0,0)[b]{\smash{{\SetFigFont{12}{24.0}
{\rmdefault}{\mddefault}{\updefault}{\color[rgb]{0,0,0}$d$}%
}}}}
\put(9601,-5761){\makebox(0,0)[b]{\smash{{\SetFigFont{12}{24.0}
{\rmdefault}{\mddefault}{\updefault}{\color[rgb]{0,0,0}$x$}%
}}}}
\put(9826,-5461){\makebox(0,0)[b]{\smash{{\SetFigFont{12}{24.0}
{\rmdefault}{\mddefault}{\updefault}{\color[rgb]{0,0,0}$D$}%
}}}}
\put(8401,-6286){\makebox(0,0)[b]{\smash{{\SetFigFont{12}{24.0}
{\rmdefault}{\mddefault}{\updefault}{\color[rgb]{0,0,0}$C$}%
}}}}
\put(8401,-5386){\makebox(0,0)[b]{\smash{{\SetFigFont{12}{24.0}
{\rmdefault}{\mddefault}{\updefault}{\color[rgb]{0,0,0}$B$}%
}}}}
\put(7801,-5761){\makebox(0,0)[b]{\smash{{\SetFigFont{12}{24.0}
{\rmdefault}{\mddefault}{\updefault}{\color[rgb]{0,0,0}$A$}%
}}}}
\end{picture}%
\nop{
+----------------------+
|                      |
|   +--------+-----+   |
|   |        |     |   |
+-> a -\ +-> c     V --+               B=ac
        X    +---> x           A=ab            D=x
+-> b -/ +-> d     ^ --+               C=bd
|   |        |     |   |
|   +--------+-----+   |
|                      |
+----------------------+
}

This example is in the {\tt disjointnotsingle.py} file. As required, $SCL(B,F)$
implies $A \rightarrow B$ and $A \rightarrow C$ and nothing else. Realizing all
equivalences among $A$, $B$, $C$ and $D$ requires completing the loops: while
$A$ already implies other equivalent sets, $B$, $C$ and $D$ do not. This can
only be done with clauses of $BCL(B,F)$. Since tautologies are useless and
heads in $SCL(B,F)$ are forbidden, the only possible clauses for $B$ are $B
\rightarrow b$ and $B \rightarrow x$; this respectively leaves $C \rightarrow
x$ or $C \rightarrow a$ for $C$. Either $a$ or $b$ is the head of one such
clause. This leaves only the other one to $D$, but neither $D \rightarrow a$
alone nor $D \rightarrow b$ alone is enough to make $D$ entail the other
equivalent sets. The problem can be attributed to the presence of $D$, since
all equivalences are otherwise completed by $B \rightarrow b$ and $C
\rightarrow a$.

% \ttyfig{converge}{ }

The problem may show up even if $SCL(B,F)$ is empty. The counterexample is in
{\tt disjointemptynotsingle.py}. It makes $ad$, $be$, $cf$ and $abc$
equivalent. Each of them should imply another; since $SCL(B,F)$ is empty, this
can only be done by clauses having them as bodies. Since each body differs from
the others by two variables, the total is eight clauses. The variables are only
six, insufficient for eight single-head clauses.

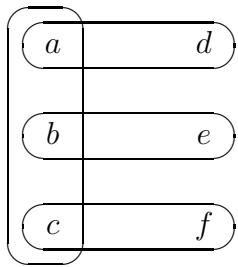
\setlength{\unitlength}{5000sp}%
\begingroup\makeatletter\ifx\SetFigFont\undefined%
\gdef\SetFigFont#1#2#3#4#5{%
  \reset@font\fontsize{#1}{#2pt}%
  \fontfamily{#3}\fontseries{#4}\fontshape{#5}%
  \selectfont}%
\fi\endgroup%
\begin{picture}(1149,1299)(7264,-6223)
\thinlines
{\color[rgb]{0,0,0}\put(7456,-5131){\oval(210,210)[bl]}
\put(7456,-5116){\oval(210,210)[tl]}
\put(8296,-5131){\oval(210,210)[br]}
\put(8296,-5116){\oval(210,210)[tr]}
\put(7456,-5236){\line( 1, 0){840}}
\put(7456,-5011){\line( 1, 0){840}}
\put(7351,-5131){\line( 0, 1){ 15}}
\put(8401,-5131){\line( 0, 1){ 15}}
}%
{\color[rgb]{0,0,0}\put(7456,-5581){\oval(210,210)[bl]}
\put(7456,-5566){\oval(210,210)[tl]}
\put(8296,-5581){\oval(210,210)[br]}
\put(8296,-5566){\oval(210,210)[tr]}
\put(7456,-5686){\line( 1, 0){840}}
\put(7456,-5461){\line( 1, 0){840}}
\put(7351,-5581){\line( 0, 1){ 15}}
\put(8401,-5581){\line( 0, 1){ 15}}
}%
{\color[rgb]{0,0,0}\put(7456,-6031){\oval(210,210)[bl]}
\put(7456,-6016){\oval(210,210)[tl]}
\put(8296,-6031){\oval(210,210)[br]}
\put(8296,-6016){\oval(210,210)[tr]}
\put(7456,-6136){\line( 1, 0){840}}
\put(7456,-5911){\line( 1, 0){840}}
\put(7351,-6031){\line( 0, 1){ 15}}
\put(8401,-6031){\line( 0, 1){ 15}}
}%
{\color[rgb]{0,0,0}\put(7381,-6106){\oval(210,210)[bl]}
\put(7381,-5041){\oval(210,210)[tl]}
\put(7546,-6106){\oval(210,210)[br]}
\put(7546,-5041){\oval(210,210)[tr]}
\put(7381,-6211){\line( 1, 0){165}}
\put(7381,-4936){\line( 1, 0){165}}
\put(7276,-6106){\line( 0, 1){1065}}
\put(7651,-6106){\line( 0, 1){1065}}
}%
\put(7501,-5161){\makebox(0,0)[b]{\smash{{\SetFigFont{12}{24.0}
{\rmdefault}{\mddefault}{\updefault}{\color[rgb]{0,0,0}$a$}%
}}}}
\put(8251,-5161){\makebox(0,0)[b]{\smash{{\SetFigFont{12}{24.0}
{\rmdefault}{\mddefault}{\updefault}{\color[rgb]{0,0,0}$d$}%
}}}}
\put(7501,-5611){\makebox(0,0)[b]{\smash{{\SetFigFont{12}{24.0}
{\rmdefault}{\mddefault}{\updefault}{\color[rgb]{0,0,0}$b$}%
}}}}
\put(8251,-5611){\makebox(0,0)[b]{\smash{{\SetFigFont{12}{24.0}
{\rmdefault}{\mddefault}{\updefault}{\color[rgb]{0,0,0}$e$}%
}}}}
\put(7501,-6061){\makebox(0,0)[b]{\smash{{\SetFigFont{12}{24.0}
{\rmdefault}{\mddefault}{\updefault}{\color[rgb]{0,0,0}$c$}%
}}}}
\put(8251,-6061){\makebox(0,0)[b]{\smash{{\SetFigFont{12}{24.0}
{\rmdefault}{\mddefault}{\updefault}{\color[rgb]{0,0,0}$f$}%
}}}}
\end{picture}%
\nop{
+--+
|+-----+
||a|  d|
|+-----+
|  |
|+-----+
||b|  e|
|+-----+
|  |
|+-----+
||c|  f|
|+-----+
+--+
}

\item The same variable is necessary for two entailments among equivalent sets.
For example, $F$ entails both $A \rightarrow B$ and $C \rightarrow D$, but $B$
and $D$ contain the same variable. An example is $A = \{a,b\}$, $B = \{c,d\}$,
$C = \{a,e\}$ and $D = \{c,f\}$ with the formula in the file {\tt
disjointemptynotsingle.py}.

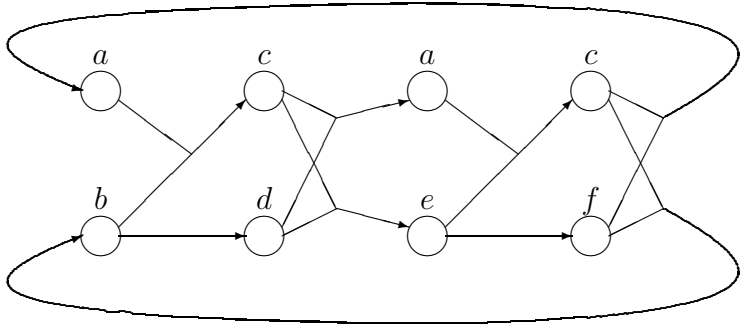
\setlength{\unitlength}{5000sp}%
\begingroup\makeatletter\ifx\SetFigFont\undefined%
\gdef\SetFigFont#1#2#3#4#5{%
  \reset@font\fontsize{#1}{#2pt}%
  \fontfamily{#3}\fontseries{#4}\fontshape{#5}%
  \selectfont}%
\fi\endgroup%
\begin{picture}(3648,1612)(5107,-4737)
{\color[rgb]{0,0,0}\thinlines
\put(5581,-3571){\circle{202}}
}%
{\color[rgb]{0,0,0}\put(5581,-4291){\circle{202}}
}%
{\color[rgb]{0,0,0}\put(6391,-3571){\circle{202}}
}%
{\color[rgb]{0,0,0}\put(6391,-4291){\circle{202}}
}%
{\color[rgb]{0,0,0}\put(7201,-3571){\circle{202}}
}%
{\color[rgb]{0,0,0}\put(7201,-4291){\circle{202}}
}%
{\color[rgb]{0,0,0}\put(8011,-3571){\circle{202}}
}%
{\color[rgb]{0,0,0}\put(8011,-4291){\circle{202}}
}%
{\color[rgb]{0,0,0}\put(5671,-4246){\vector( 1, 1){630}}
}%
{\color[rgb]{0,0,0}\put(5671,-3616){\line( 4,-3){360}}
}%
{\color[rgb]{0,0,0}\put(5671,-4291){\vector( 1, 0){630}}
}%
{\color[rgb]{0,0,0}\put(6751,-4156){\vector( 4,-1){360}}
}%
{\color[rgb]{0,0,0}\put(6481,-3616){\line( 1,-2){270}}
\put(6751,-4156){\line(-2,-1){270}}
}%
{\color[rgb]{0,0,0}\put(6481,-3571){\line( 2,-1){270}}
\put(6751,-3706){\line(-1,-2){270}}
}%
{\color[rgb]{0,0,0}\put(6751,-3706){\vector( 4, 1){360}}
}%
{\color[rgb]{0,0,0}\put(7291,-4291){\vector( 1, 0){630}}
}%
{\color[rgb]{0,0,0}\put(7291,-4246){\vector( 1, 1){630}}
}%
{\color[rgb]{0,0,0}\put(7291,-3616){\line( 4,-3){360}}
}%
{\color[rgb]{0,0,0}\put(8101,-3571){\line( 2,-1){270}}
\put(8371,-3706){\line(-1,-2){270}}
}%
{\color[rgb]{0,0,0}\put(8101,-3616){\line( 1,-2){270}}
\put(8371,-4156){\line(-2,-1){270}}
}%
{\color[rgb]{0,0,0}\put(8371,-3706){\line( 1, 0){  1}}
\multiput(8378,-3702)(5.00000,2.00000){2}{\makebox(0.7938,5.5563){\tiny.}}
\multiput(8383,-3700)(6.00000,4.00000){2}{\makebox(0.7938,5.5563){\tiny.}}
\multiput(8389,-3696)(4.00000,2.00000){3}{\makebox(0.7938,5.5563){\tiny.}}
\multiput(8397,-3692)(5.00000,3.00000){3}{\makebox(0.7938,5.5563){\tiny.}}
\multiput(8407,-3686)(4.00000,2.00000){4}{\makebox(0.7938,5.5563){\tiny.}}
\multiput(8419,-3680)(4.36273,2.61764){4}{\makebox(0.7938,5.5563){\tiny.}}
\multiput(8432,-3672)(3.80000,1.90000){5}{\makebox(0.7938,5.5563){\tiny.}}
\multiput(8447,-3664)(4.04413,2.42647){5}{\makebox(0.7938,5.5563){\tiny.}}
\multiput(8463,-3654)(3.52942,2.11765){6}{\makebox(0.7938,5.5563){\tiny.}}
\multiput(8481,-3644)(3.76470,2.25882){6}{\makebox(0.7938,5.5563){\tiny.}}
\multiput(8500,-3633)(3.30770,2.20513){7}{\makebox(0.7938,5.5563){\tiny.}}
\multiput(8520,-3620)(4.00000,2.40000){6}{\makebox(0.7938,5.5563){\tiny.}}
\multiput(8540,-3608)(3.38462,2.25641){7}{\makebox(0.7938,5.5563){\tiny.}}
\multiput(8560,-3594)(3.50000,2.33333){7}{\makebox(0.7938,5.5563){\tiny.}}
\multiput(8581,-3580)(3.50000,2.33333){7}{\makebox(0.7938,5.5563){\tiny.}}
\multiput(8602,-3566)(3.33333,2.50000){7}{\makebox(0.7938,5.5563){\tiny.}}
\multiput(8622,-3551)(3.15040,2.52032){7}{\makebox(0.7938,5.5563){\tiny.}}
\multiput(8641,-3536)(3.60000,3.00000){6}{\makebox(0.7938,5.5563){\tiny.}}
\multiput(8659,-3521)(3.08197,2.56831){7}{\makebox(0.7938,5.5563){\tiny.}}
\multiput(8677,-3505)(3.00000,3.00000){6}{\makebox(0.7938,5.5563){\tiny.}}
\multiput(8692,-3490)(2.72132,3.26558){6}{\makebox(0.7938,5.5563){\tiny.}}
\multiput(8706,-3474)(2.57560,3.21950){6}{\makebox(0.7938,5.5563){\tiny.}}
\multiput(8719,-3458)(2.50000,3.75000){5}{\makebox(0.7938,5.5563){\tiny.}}
\multiput(8729,-3443)(1.62070,4.05175){5}{\makebox(0.7938,5.5563){\tiny.}}
\multiput(8736,-3427)(1.66667,5.00000){4}{\makebox(0.7938,5.5563){\tiny.}}
\multiput(8741,-3412)(0.66215,3.97290){5}{\makebox(0.7938,5.5563){\tiny.}}
\put(8743,-3396){\line( 0, 1){ 15}}
\multiput(8742,-3381)(-1.25490,5.01960){4}{\makebox(0.7938,5.5563){\tiny.}}
\multiput(8738,-3366)(-2.00000,4.00000){5}{\makebox(0.7938,5.5563){\tiny.}}
\multiput(8730,-3350)(-2.54098,3.04918){6}{\makebox(0.7938,5.5563){\tiny.}}
\multiput(8717,-3335)(-3.10000,3.10000){6}{\makebox(0.7938,5.5563){\tiny.}}
\multiput(8701,-3320)(-3.69230,2.46153){7}{\makebox(0.7938,5.5563){\tiny.}}
\multiput(8679,-3305)(-3.77143,1.88571){8}{\makebox(0.7938,5.5563){\tiny.}}
\multiput(8653,-3291)(-3.95000,1.97500){9}{\makebox(0.7938,5.5563){\tiny.}}
\multiput(8621,-3276)(-3.75862,1.50345){11}{\makebox(0.7938,5.5563){\tiny.}}
\multiput(8583,-3262)(-3.90000,1.30000){12}{\makebox(0.7938,5.5563){\tiny.}}
\multiput(8540,-3248)(-3.85520,0.96380){14}{\makebox(0.7938,5.5563){\tiny.}}
\multiput(8490,-3235)(-4.07967,0.81593){15}{\makebox(0.7938,5.5563){\tiny.}}
\multiput(8433,-3223)(-4.12821,0.82564){16}{\makebox(0.7938,5.5563){\tiny.}}
\multiput(8371,-3211)(-4.06564,0.67761){15}{\makebox(0.7938,5.5563){\tiny.}}
\multiput(8314,-3202)(-4.05405,0.67568){16}{\makebox(0.7938,5.5563){\tiny.}}
\multiput(8253,-3193)(-4.16216,0.69369){16}{\makebox(0.7938,5.5563){\tiny.}}
\multiput(8190,-3186)(-4.02365,0.67061){17}{\makebox(0.7938,5.5563){\tiny.}}
\multiput(8125,-3179)(-4.07432,0.67905){17}{\makebox(0.7938,5.5563){\tiny.}}
\multiput(8059,-3173)(-4.07432,0.67905){17}{\makebox(0.7938,5.5563){\tiny.}}
\put(7993,-3167){\line(-1, 0){ 65}}
\put(7928,-3162){\line(-1, 0){ 65}}
\put(7863,-3158){\line(-1, 0){ 65}}
\put(7798,-3154){\line(-1, 0){ 62}}
\put(7736,-3150){\line(-1, 0){ 62}}
\put(7674,-3148){\line(-1, 0){ 60}}
\put(7614,-3145){\line(-1, 0){ 58}}
\put(7556,-3143){\line(-1, 0){ 56}}
\put(7500,-3141){\line(-1, 0){ 54}}
\put(7446,-3140){\line(-1, 0){ 53}}
\put(7393,-3139){\line(-1, 0){ 52}}
\put(7341,-3138){\line(-1, 0){ 49}}
\put(7292,-3138){\line(-1, 0){ 49}}
\put(7243,-3138){\line(-1, 0){ 47}}
\put(7196,-3137){\line(-1, 0){ 46}}
\put(7150,-3137){\line(-1, 0){ 45}}
\put(7105,-3138){\line(-1, 0){ 44}}
\put(7061,-3138){\line(-1, 0){ 44}}
\put(7017,-3139){\line(-1, 0){ 43}}
\put(6974,-3139){\line(-1, 0){ 43}}
\put(6931,-3140){\line(-1, 0){ 43}}
\put(6888,-3141){\line(-1, 0){ 43}}
\put(6845,-3141){\line(-1, 0){ 44}}
\put(6801,-3142){\line(-1, 0){ 44}}
\put(6757,-3143){\line(-1, 0){ 45}}
\put(6712,-3144){\line(-1, 0){ 46}}
\put(6666,-3146){\line(-1, 0){ 47}}
\put(6619,-3147){\line(-1, 0){ 49}}
\put(6570,-3148){\line(-1, 0){ 49}}
\put(6521,-3150){\line(-1, 0){ 52}}
\put(6469,-3151){\line(-1, 0){ 53}}
\put(6416,-3153){\line(-1, 0){ 54}}
\put(6362,-3155){\line(-1, 0){ 56}}
\put(6306,-3157){\line(-1, 0){ 58}}
\put(6248,-3159){\line(-1, 0){ 60}}
\put(6188,-3162){\line(-1, 0){ 62}}
\put(6126,-3165){\line(-1, 0){ 62}}
\put(6064,-3168){\line(-1, 0){ 65}}
\put(5999,-3171){\line(-1, 0){ 65}}
\put(5934,-3174){\line(-1, 0){ 65}}
\put(5869,-3178){\line(-1, 0){ 66}}
\put(5803,-3183){\line(-1, 0){ 66}}
\multiput(5737,-3187)(-4.01351,-0.66892){17}{\makebox(0.7938,5.5563){\tiny.}}
\put(5672,-3193){\line(-1, 0){ 63}}
\multiput(5609,-3198)(-4.02162,-0.67027){16}{\makebox(0.7938,5.5563){\tiny.}}
\multiput(5548,-3204)(-4.04247,-0.67375){15}{\makebox(0.7938,5.5563){\tiny.}}
\multiput(5491,-3211)(-3.97838,-0.66306){16}{\makebox(0.7938,5.5563){\tiny.}}
\multiput(5431,-3219)(-4.15385,-0.69231){14}{\makebox(0.7938,5.5563){\tiny.}}
\multiput(5377,-3228)(-3.99038,-0.79808){13}{\makebox(0.7938,5.5563){\tiny.}}
\multiput(5329,-3237)(-3.91608,-0.78322){12}{\makebox(0.7938,5.5563){\tiny.}}
\multiput(5286,-3246)(-4.13072,-1.03268){10}{\makebox(0.7938,5.5563){\tiny.}}
\multiput(5249,-3256)(-3.97500,-1.32500){9}{\makebox(0.7938,5.5563){\tiny.}}
\multiput(5217,-3266)(-3.81773,-1.52709){8}{\makebox(0.7938,5.5563){\tiny.}}
\multiput(5190,-3276)(-3.66667,-1.83333){7}{\makebox(0.7938,5.5563){\tiny.}}
\multiput(5168,-3287)(-4.22795,-2.53677){5}{\makebox(0.7938,5.5563){\tiny.}}
\multiput(5151,-3297)(-3.47560,-2.78048){5}{\makebox(0.7938,5.5563){\tiny.}}
\multiput(5137,-3308)(-2.86887,-3.44264){4}{\makebox(0.7938,5.5563){\tiny.}}
\multiput(5128,-3318)(-1.86667,-3.73333){4}{\makebox(0.7938,5.5563){\tiny.}}
\multiput(5122,-3329)(-1.65000,-4.95000){3}{\makebox(0.7938,5.5563){\tiny.}}
\put(5119,-3339){\line( 0,-1){ 11}}
\multiput(5119,-3350)(1.38235,-5.52940){3}{\makebox(0.7938,5.5563){\tiny.}}
\multiput(5122,-3361)(1.86667,-3.73333){4}{\makebox(0.7938,5.5563){\tiny.}}
\multiput(5128,-3372)(2.72000,-3.62667){4}{\makebox(0.7938,5.5563){\tiny.}}
\multiput(5136,-3383)(3.33333,-3.33333){4}{\makebox(0.7938,5.5563){\tiny.}}
\multiput(5146,-3393)(2.87500,-2.87500){5}{\makebox(0.7938,5.5563){\tiny.}}
\multiput(5158,-3404)(3.47560,-2.78048){5}{\makebox(0.7938,5.5563){\tiny.}}
\multiput(5172,-3415)(3.72000,-2.79000){5}{\makebox(0.7938,5.5563){\tiny.}}
\multiput(5187,-3426)(4.03845,-2.69230){5}{\makebox(0.7938,5.5563){\tiny.}}
\multiput(5203,-3437)(3.52942,-2.11765){6}{\makebox(0.7938,5.5563){\tiny.}}
\multiput(5221,-3447)(3.61764,-2.17058){6}{\makebox(0.7938,5.5563){\tiny.}}
\multiput(5239,-3458)(4.00000,-2.00000){6}{\makebox(0.7938,5.5563){\tiny.}}
\multiput(5259,-3468)(3.84000,-1.92000){6}{\makebox(0.7938,5.5563){\tiny.}}
\multiput(5278,-3478)(4.00000,-2.00000){6}{\makebox(0.7938,5.5563){\tiny.}}
\multiput(5298,-3488)(3.92000,-1.96000){6}{\makebox(0.7938,5.5563){\tiny.}}
\multiput(5318,-3497)(3.76000,-1.88000){6}{\makebox(0.7938,5.5563){\tiny.}}
\multiput(5337,-3506)(3.76000,-1.88000){6}{\makebox(0.7938,5.5563){\tiny.}}
\multiput(5356,-3515)(4.56898,-1.82759){5}{\makebox(0.7938,5.5563){\tiny.}}
\multiput(5374,-3523)(4.56898,-1.82759){5}{\makebox(0.7938,5.5563){\tiny.}}
\multiput(5392,-3531)(3.96552,-1.58621){5}{\makebox(0.7938,5.5563){\tiny.}}
\multiput(5408,-3537)(3.70000,-1.85000){5}{\makebox(0.7938,5.5563){\tiny.}}
\multiput(5423,-3544)(4.31033,-1.72413){4}{\makebox(0.7938,5.5563){\tiny.}}
\multiput(5436,-3549)(4.02300,-1.60920){4}{\makebox(0.7938,5.5563){\tiny.}}
\multiput(5448,-3554)(5.55000,-1.85000){3}{\makebox(0.7938,5.5563){\tiny.}}
\multiput(5459,-3558)(4.56895,-1.82758){3}{\makebox(0.7938,5.5563){\tiny.}}
\multiput(5468,-3562)(7.06900,-2.82760){2}{\makebox(0.7938,5.5563){\tiny.}}
\multiput(5475,-3565)(5.00000,-2.00000){2}{\makebox(0.7938,5.5563){\tiny.}}
\multiput(5480,-3567)(5.55000,-1.85000){3}{\makebox(0.7938,5.5563){\tiny.}}
\put(5491,-3571){\vector( 3,-1){0}}
}%
{\color[rgb]{0,0,0}\put(8371,-4156){\line( 1, 0){  1}}
\multiput(8378,-4160)(5.00000,-2.00000){2}{\makebox(0.7938,5.5563){\tiny.}}
\multiput(8383,-4162)(6.00000,-4.00000){2}{\makebox(0.7938,5.5563){\tiny.}}
\multiput(8389,-4166)(4.00000,-2.00000){3}{\makebox(0.7938,5.5563){\tiny.}}
\multiput(8397,-4170)(5.00000,-3.00000){3}{\makebox(0.7938,5.5563){\tiny.}}
\multiput(8407,-4176)(4.00000,-2.00000){4}{\makebox(0.7938,5.5563){\tiny.}}
\multiput(8419,-4182)(4.36273,-2.61764){4}{\makebox(0.7938,5.5563){\tiny.}}
\multiput(8432,-4190)(3.80000,-1.90000){5}{\makebox(0.7938,5.5563){\tiny.}}
\multiput(8447,-4198)(4.04413,-2.42647){5}{\makebox(0.7938,5.5563){\tiny.}}
\multiput(8463,-4208)(3.52942,-2.11765){6}{\makebox(0.7938,5.5563){\tiny.}}
\multiput(8481,-4218)(3.76470,-2.25882){6}{\makebox(0.7938,5.5563){\tiny.}}
\multiput(8500,-4229)(3.30770,-2.20513){7}{\makebox(0.7938,5.5563){\tiny.}}
\multiput(8520,-4242)(4.00000,-2.40000){6}{\makebox(0.7938,5.5563){\tiny.}}
\multiput(8540,-4254)(3.38462,-2.25641){7}{\makebox(0.7938,5.5563){\tiny.}}
\multiput(8560,-4268)(3.50000,-2.33333){7}{\makebox(0.7938,5.5563){\tiny.}}
\multiput(8581,-4282)(3.50000,-2.33333){7}{\makebox(0.7938,5.5563){\tiny.}}
\multiput(8602,-4296)(3.33333,-2.50000){7}{\makebox(0.7938,5.5563){\tiny.}}
\multiput(8622,-4311)(3.15040,-2.52032){7}{\makebox(0.7938,5.5563){\tiny.}}
\multiput(8641,-4326)(3.60000,-3.00000){6}{\makebox(0.7938,5.5563){\tiny.}}
\multiput(8659,-4341)(3.08197,-2.56831){7}{\makebox(0.7938,5.5563){\tiny.}}
\multiput(8677,-4357)(3.00000,-3.00000){6}{\makebox(0.7938,5.5563){\tiny.}}
\multiput(8692,-4372)(2.72132,-3.26558){6}{\makebox(0.7938,5.5563){\tiny.}}
\multiput(8706,-4388)(2.57560,-3.21950){6}{\makebox(0.7938,5.5563){\tiny.}}
\multiput(8719,-4404)(2.50000,-3.75000){5}{\makebox(0.7938,5.5563){\tiny.}}
\multiput(8729,-4419)(1.62070,-4.05175){5}{\makebox(0.7938,5.5563){\tiny.}}
\multiput(8736,-4435)(1.66667,-5.00000){4}{\makebox(0.7938,5.5563){\tiny.}}
\multiput(8741,-4450)(0.66215,-3.97290){5}{\makebox(0.7938,5.5563){\tiny.}}
\put(8743,-4466){\line( 0,-1){ 15}}
\multiput(8742,-4481)(-1.25490,-5.01960){4}{\makebox(0.7938,5.5563){\tiny.}}
\multiput(8738,-4496)(-2.00000,-4.00000){5}{\makebox(0.7938,5.5563){\tiny.}}
\multiput(8730,-4512)(-2.54098,-3.04918){6}{\makebox(0.7938,5.5563){\tiny.}}
\multiput(8717,-4527)(-3.10000,-3.10000){6}{\makebox(0.7938,5.5563){\tiny.}}
\multiput(8701,-4542)(-3.69230,-2.46153){7}{\makebox(0.7938,5.5563){\tiny.}}
\multiput(8679,-4557)(-3.77143,-1.88571){8}{\makebox(0.7938,5.5563){\tiny.}}
\multiput(8653,-4571)(-3.95000,-1.97500){9}{\makebox(0.7938,5.5563){\tiny.}}
\multiput(8621,-4586)(-3.75862,-1.50345){11}{\makebox(0.7938,5.5563){\tiny.}}
\multiput(8583,-4600)(-3.90000,-1.30000){12}{\makebox(0.7938,5.5563){\tiny.}}
\multiput(8540,-4614)(-3.85520,-0.96380){14}{\makebox(0.7938,5.5563){\tiny.}}
\multiput(8490,-4627)(-4.07967,-0.81593){15}{\makebox(0.7938,5.5563){\tiny.}}
\multiput(8433,-4639)(-4.12821,-0.82564){16}{\makebox(0.7938,5.5563){\tiny.}}
\multiput(8371,-4651)(-4.06564,-0.67761){15}{\makebox(0.7938,5.5563){\tiny.}}
\multiput(8314,-4660)(-4.05405,-0.67568){16}{\makebox(0.7938,5.5563){\tiny.}}
\multiput(8253,-4669)(-4.16216,-0.69369){16}{\makebox(0.7938,5.5563){\tiny.}}
\multiput(8190,-4676)(-4.02365,-0.67061){17}{\makebox(0.7938,5.5563){\tiny.}}
\multiput(8125,-4683)(-4.07432,-0.67905){17}{\makebox(0.7938,5.5563){\tiny.}}
\multiput(8059,-4689)(-4.07432,-0.67905){17}{\makebox(0.7938,5.5563){\tiny.}}
\put(7993,-4695){\line(-1, 0){ 65}}
\put(7928,-4700){\line(-1, 0){ 65}}
\put(7863,-4704){\line(-1, 0){ 65}}
\put(7798,-4708){\line(-1, 0){ 62}}
\put(7736,-4712){\line(-1, 0){ 62}}
\put(7674,-4714){\line(-1, 0){ 60}}
\put(7614,-4717){\line(-1, 0){ 58}}
\put(7556,-4719){\line(-1, 0){ 56}}
\put(7500,-4721){\line(-1, 0){ 54}}
\put(7446,-4722){\line(-1, 0){ 53}}
\put(7393,-4723){\line(-1, 0){ 52}}
\put(7341,-4724){\line(-1, 0){ 49}}
\put(7292,-4724){\line(-1, 0){ 49}}
\put(7243,-4724){\line(-1, 0){ 47}}
\put(7196,-4725){\line(-1, 0){ 46}}
\put(7150,-4725){\line(-1, 0){ 45}}
\put(7105,-4724){\line(-1, 0){ 44}}
\put(7061,-4724){\line(-1, 0){ 44}}
\put(7017,-4723){\line(-1, 0){ 43}}
\put(6974,-4723){\line(-1, 0){ 43}}
\put(6931,-4722){\line(-1, 0){ 43}}
\put(6888,-4721){\line(-1, 0){ 43}}
\put(6845,-4721){\line(-1, 0){ 44}}
\put(6801,-4720){\line(-1, 0){ 44}}
\put(6757,-4719){\line(-1, 0){ 45}}
\put(6712,-4718){\line(-1, 0){ 46}}
\put(6666,-4716){\line(-1, 0){ 47}}
\put(6619,-4715){\line(-1, 0){ 49}}
\put(6570,-4714){\line(-1, 0){ 49}}
\put(6521,-4712){\line(-1, 0){ 52}}
\put(6469,-4711){\line(-1, 0){ 53}}
\put(6416,-4709){\line(-1, 0){ 54}}
\put(6362,-4707){\line(-1, 0){ 56}}
\put(6306,-4705){\line(-1, 0){ 58}}
\put(6248,-4703){\line(-1, 0){ 60}}
\put(6188,-4700){\line(-1, 0){ 62}}
\put(6126,-4697){\line(-1, 0){ 62}}
\put(6064,-4694){\line(-1, 0){ 65}}
\put(5999,-4691){\line(-1, 0){ 65}}
\put(5934,-4688){\line(-1, 0){ 65}}
\put(5869,-4684){\line(-1, 0){ 66}}
\put(5803,-4679){\line(-1, 0){ 66}}
\multiput(5737,-4675)(-4.01351,0.66892){17}{\makebox(0.7938,5.5563){\tiny.}}
\put(5672,-4669){\line(-1, 0){ 63}}
\multiput(5609,-4664)(-4.02162,0.67027){16}{\makebox(0.7938,5.5563){\tiny.}}
\multiput(5548,-4658)(-4.04247,0.67375){15}{\makebox(0.7938,5.5563){\tiny.}}
\multiput(5491,-4651)(-3.97838,0.66306){16}{\makebox(0.7938,5.5563){\tiny.}}
\multiput(5431,-4643)(-4.15385,0.69231){14}{\makebox(0.7938,5.5563){\tiny.}}
\multiput(5377,-4634)(-3.99038,0.79808){13}{\makebox(0.7938,5.5563){\tiny.}}
\multiput(5329,-4625)(-3.91608,0.78322){12}{\makebox(0.7938,5.5563){\tiny.}}
\multiput(5286,-4616)(-4.13072,1.03268){10}{\makebox(0.7938,5.5563){\tiny.}}
\multiput(5249,-4606)(-3.97500,1.32500){9}{\makebox(0.7938,5.5563){\tiny.}}
\multiput(5217,-4596)(-3.81773,1.52709){8}{\makebox(0.7938,5.5563){\tiny.}}
\multiput(5190,-4586)(-3.66667,1.83333){7}{\makebox(0.7938,5.5563){\tiny.}}
\multiput(5168,-4575)(-4.22795,2.53677){5}{\makebox(0.7938,5.5563){\tiny.}}
\multiput(5151,-4565)(-3.47560,2.78048){5}{\makebox(0.7938,5.5563){\tiny.}}
\multiput(5137,-4554)(-2.86887,3.44264){4}{\makebox(0.7938,5.5563){\tiny.}}
\multiput(5128,-4544)(-1.86667,3.73333){4}{\makebox(0.7938,5.5563){\tiny.}}
\multiput(5122,-4533)(-1.65000,4.95000){3}{\makebox(0.7938,5.5563){\tiny.}}
\put(5119,-4523){\line( 0, 1){ 11}}
\multiput(5119,-4512)(1.38235,5.52940){3}{\makebox(0.7938,5.5563){\tiny.}}
\multiput(5122,-4501)(1.86667,3.73333){4}{\makebox(0.7938,5.5563){\tiny.}}
\multiput(5128,-4490)(2.72000,3.62667){4}{\makebox(0.7938,5.5563){\tiny.}}
\multiput(5136,-4479)(3.33333,3.33333){4}{\makebox(0.7938,5.5563){\tiny.}}
\multiput(5146,-4469)(2.87500,2.87500){5}{\makebox(0.7938,5.5563){\tiny.}}
\multiput(5158,-4458)(3.47560,2.78048){5}{\makebox(0.7938,5.5563){\tiny.}}
\multiput(5172,-4447)(3.72000,2.79000){5}{\makebox(0.7938,5.5563){\tiny.}}
\multiput(5187,-4436)(4.03845,2.69230){5}{\makebox(0.7938,5.5563){\tiny.}}
\multiput(5203,-4425)(3.52942,2.11765){6}{\makebox(0.7938,5.5563){\tiny.}}
\multiput(5221,-4415)(3.61764,2.17058){6}{\makebox(0.7938,5.5563){\tiny.}}
\multiput(5239,-4404)(4.00000,2.00000){6}{\makebox(0.7938,5.5563){\tiny.}}
\multiput(5259,-4394)(3.84000,1.92000){6}{\makebox(0.7938,5.5563){\tiny.}}
\multiput(5278,-4384)(4.00000,2.00000){6}{\makebox(0.7938,5.5563){\tiny.}}
\multiput(5298,-4374)(3.92000,1.96000){6}{\makebox(0.7938,5.5563){\tiny.}}
\multiput(5318,-4365)(3.76000,1.88000){6}{\makebox(0.7938,5.5563){\tiny.}}
\multiput(5337,-4356)(3.76000,1.88000){6}{\makebox(0.7938,5.5563){\tiny.}}
\multiput(5356,-4347)(4.56898,1.82759){5}{\makebox(0.7938,5.5563){\tiny.}}
\multiput(5374,-4339)(4.56898,1.82759){5}{\makebox(0.7938,5.5563){\tiny.}}
\multiput(5392,-4331)(3.96552,1.58621){5}{\makebox(0.7938,5.5563){\tiny.}}
\multiput(5408,-4325)(3.70000,1.85000){5}{\makebox(0.7938,5.5563){\tiny.}}
\multiput(5423,-4318)(4.31033,1.72413){4}{\makebox(0.7938,5.5563){\tiny.}}
\multiput(5436,-4313)(4.02300,1.60920){4}{\makebox(0.7938,5.5563){\tiny.}}
\multiput(5448,-4308)(5.55000,1.85000){3}{\makebox(0.7938,5.5563){\tiny.}}
\multiput(5459,-4304)(4.56895,1.82758){3}{\makebox(0.7938,5.5563){\tiny.}}
\multiput(5468,-4300)(7.06900,2.82760){2}{\makebox(0.7938,5.5563){\tiny.}}
\multiput(5475,-4297)(5.00000,2.00000){2}{\makebox(0.7938,5.5563){\tiny.}}
\multiput(5480,-4295)(5.55000,1.85000){3}{\makebox(0.7938,5.5563){\tiny.}}
\put(5491,-4291){\vector( 3, 1){0}}
}%
\put(5581,-3436){\makebox(0,0)[b]{\smash{{\SetFigFont{12}{24.0}
{\rmdefault}{\mddefault}{\updefault}{\color[rgb]{0,0,0}$a$}%
}}}}
\put(7201,-3436){\makebox(0,0)[b]{\smash{{\SetFigFont{12}{24.0}
{\rmdefault}{\mddefault}{\updefault}{\color[rgb]{0,0,0}$a$}%
}}}}
\put(6391,-3436){\makebox(0,0)[b]{\smash{{\SetFigFont{12}{24.0}
{\rmdefault}{\mddefault}{\updefault}{\color[rgb]{0,0,0}$c$}%
}}}}
\put(5581,-4156){\makebox(0,0)[b]{\smash{{\SetFigFont{12}{24.0}
{\rmdefault}{\mddefault}{\updefault}{\color[rgb]{0,0,0}$b$}%
}}}}
\put(6391,-4156){\makebox(0,0)[b]{\smash{{\SetFigFont{12}{24.0}
{\rmdefault}{\mddefault}{\updefault}{\color[rgb]{0,0,0}$d$}%
}}}}
\put(7201,-4156){\makebox(0,0)[b]{\smash{{\SetFigFont{12}{24.0}
{\rmdefault}{\mddefault}{\updefault}{\color[rgb]{0,0,0}$e$}%
}}}}
\put(8011,-3436){\makebox(0,0)[b]{\smash{{\SetFigFont{12}{24.0}
{\rmdefault}{\mddefault}{\updefault}{\color[rgb]{0,0,0}$c$}%
}}}}
\put(8011,-4156){\makebox(0,0)[b]{\smash{{\SetFigFont{12}{24.0}
{\rmdefault}{\mddefault}{\updefault}{\color[rgb]{0,0,0}$f$}%
}}}}
\end{picture}%
\nop{
+--------------------------+
V                          |
a -+--> c  +--> a -+--> c  |
   |     |_|       |     |_|
  /      | |      /      | |
b ----> d  +--> e ----> f  |
^                          |
+--------------------------+
}

Starting from $\{a,b\}$, the $SCL(B,F)$ part of $F$ already entails $d$, which
is part of the equivalent set $\{c,d\}$. This calls for $ab \rightarrow c$. The
same applies to $\{a,e\}$ and $ae \rightarrow c$, which has the same head.

The variables $a$ and $c$ and shown twice to emphasize the structure of the
closed chain $A$, $B$, $C$ and $D$. Without the repetition the formula looks
like the following figure. The two clauses with head $a$ can be replaced by $c
\rightarrow a$.

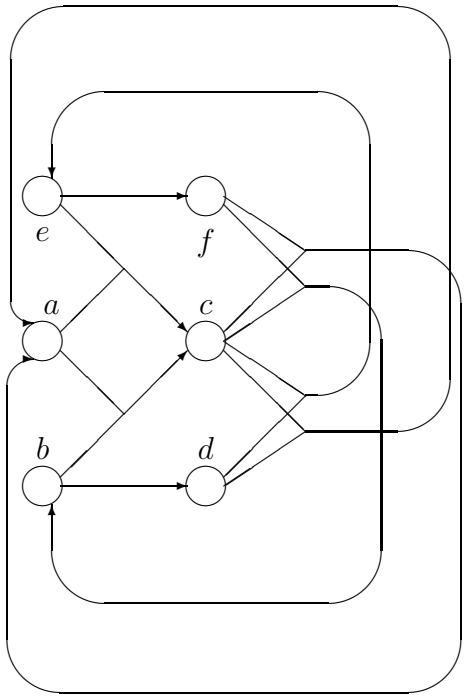
\setlength{\unitlength}{5000sp}%
\begingroup\makeatletter\ifx\SetFigFont\undefined%
\gdef\SetFigFont#1#2#3#4#5{%
  \reset@font\fontsize{#1}{#2pt}%
  \fontfamily{#3}\fontseries{#4}\fontshape{#5}%
  \selectfont}%
\fi\endgroup%
\begin{picture}(2270,3424)(4945,-5593)
\thinlines
{\color[rgb]{0,0,0}\put(5176,-4651){\vector( 0, 1){0}}
\put(5952,-4651){\oval(1552,980)[bl]}
\put(5952,-4281){\oval(1720,1720)[br]}
\put(6436,-4281){\oval(752,1420)[tr]}
}%
{\color[rgb]{0,0,0}\put(6436,-3449){\oval(638,1324)[br]}
\put(5910,-3449){\oval(1690,1690)[tr]}
\put(5910,-3031){\oval(1468,854)[tl]}
\put(5176,-3031){\vector( 0,-1){0}}
}%
{\color[rgb]{0,0,0}\put(6436,-3268){\oval(1432,2046)[br]}
\put(6062,-3268){\oval(2180,2178)[tr]}
\put(6062,-3268){\oval(2178,2178)[tl]}
\put(5086,-3268){\oval(226,966)[bl]}
\put(5086,-3751){\vector( 1, 0){0}}
}%
{\color[rgb]{0,0,0}\put(5086,-3931){\vector( 1, 0){0}}
\put(5086,-4459){\oval(264,1056)[tl]}
\put(6080,-4459){\oval(2252,2252)[bl]}
\put(6080,-4459){\oval(2252,2252)[br]}
\put(6436,-4459){\oval(1540,2136)[tr]}
}%
{\color[rgb]{0,0,0}\put(5131,-3841){\circle{202}}
}%
{\color[rgb]{0,0,0}\put(5131,-4561){\circle{202}}
}%
{\color[rgb]{0,0,0}\put(5941,-3841){\circle{202}}
}%
{\color[rgb]{0,0,0}\put(5941,-4561){\circle{202}}
}%
{\color[rgb]{0,0,0}\put(5131,-3121){\circle{202}}
}%
{\color[rgb]{0,0,0}\put(5941,-3121){\circle{202}}
}%
{\color[rgb]{0,0,0}\put(5221,-3121){\vector( 1, 0){630}}
}%
{\color[rgb]{0,0,0}\put(5221,-4561){\vector( 1, 0){630}}
}%
{\color[rgb]{0,0,0}\put(5221,-3886){\line( 1,-1){315}}
\put(5536,-4201){\line(-1,-1){315}}
}%
{\color[rgb]{0,0,0}\put(5536,-4201){\vector( 1, 1){315}}
}%
{\color[rgb]{0,0,0}\put(5221,-3166){\line( 1,-1){315}}
\put(5536,-3481){\line(-1,-1){315}}
}%
{\color[rgb]{0,0,0}\put(5536,-3481){\vector( 1,-1){315}}
}%
{\color[rgb]{0,0,0}\put(6031,-3166){\line( 1,-1){405}}
\put(6436,-3571){\line(-3,-2){405}}
}%
{\color[rgb]{0,0,0}\put(6031,-3121){\line( 3,-2){405}}
\put(6436,-3391){\line(-1,-1){405}}
}%
{\color[rgb]{0,0,0}\put(6031,-3886){\line( 1,-1){405}}
\put(6436,-4291){\line(-3,-2){405}}
}%
{\color[rgb]{0,0,0}\put(6031,-3841){\line( 3,-2){405}}
\put(6436,-4111){\line(-1,-1){405}}
}%
\put(5941,-4426){\makebox(0,0)[b]{\smash{{\SetFigFont{12}{24.0}
{\rmdefault}{\mddefault}{\updefault}{\color[rgb]{0,0,0}$d$}%
}}}}
\put(5131,-4426){\makebox(0,0)[b]{\smash{{\SetFigFont{12}{24.0}
{\rmdefault}{\mddefault}{\updefault}{\color[rgb]{0,0,0}$b$}%
}}}}
\put(5941,-3706){\makebox(0,0)[b]{\smash{{\SetFigFont{12}{24.0}
{\rmdefault}{\mddefault}{\updefault}{\color[rgb]{0,0,0}$c$}%
}}}}
\put(5131,-3346){\makebox(0,0)[b]{\smash{{\SetFigFont{12}{24.0}
{\rmdefault}{\mddefault}{\updefault}{\color[rgb]{0,0,0}$e$}%
}}}}
\put(5941,-3391){\makebox(0,0)[b]{\smash{{\SetFigFont{12}{24.0}
{\rmdefault}{\mddefault}{\updefault}{\color[rgb]{0,0,0}$f$}%
}}}}
\put(5176,-3706){\makebox(0,0)[b]{\smash{{\SetFigFont{12}{24.0}
{\rmdefault}{\mddefault}{\updefault}{\color[rgb]{0,0,0}$a$}%
}}}}
\end{picture}%
\nop{
   +----------------------+
   |                      |
   |   +--> e -----> f    |
   |   |     \        +---+
   +------>    \      |
   |   |    a --+--> c
   |   +-->    /      |
   |   |     /        +---+
   +---|--> b -----> d    |
       |                  |
       +------------------+
}

\item The formula requires connecting too many equivalent sets. All examples
shown so far show some ``local'' issues: a path of equivalent sets that cannot
be completed to form a loop. Other counterexamples allow each set to entail
another, but the resulting graph of equivalent sets of variables is
disconnected, like the formula in {\tt disconnected.py}:

\[
F = \{
abc \equiv def, def \equiv ghi, ghi \equiv jk, jk \equiv mn, mn \equiv abc,
a \rightarrow d, e \rightarrow g, i \rightarrow c,
j \rightarrow m, q \rightarrow n
\}
\]

The binary clauses in $SCL(B,F)$ push $ITERATION(B,F)$ toward entailing $abc
\rightarrow def$, $def \rightarrow ghi$ and $ghi \rightarrow abc$, closing the
loop short of including $jm$ and $qn$.

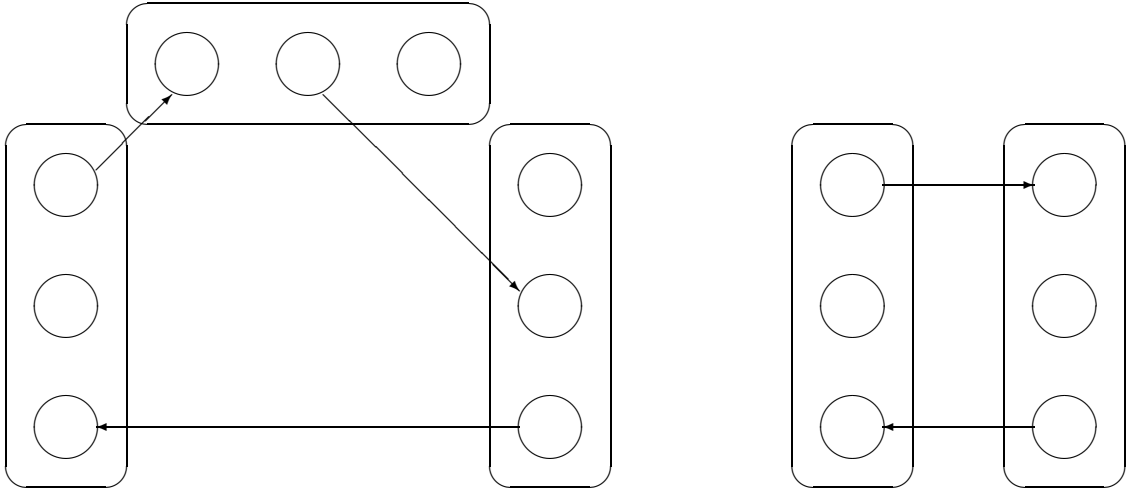
\setlength{\unitlength}{5000sp}%
\begingroup\makeatletter\ifx\SetFigFont\undefined%
\gdef\SetFigFont#1#2#3#4#5{%
  \reset@font\fontsize{#1}{#2pt}%
  \fontfamily{#3}\fontseries{#4}\fontshape{#5}%
  \selectfont}%
\fi\endgroup%
\begin{picture}(5574,2424)(5539,-6523)
{\color[rgb]{0,0,0}\thinlines
\put(7051,-4411){\circle{336}}
}%
{\color[rgb]{0,0,0}\put(6451,-4411){\circle{336}}
}%
{\color[rgb]{0,0,0}\put(7651,-4411){\circle{336}}
}%
{\color[rgb]{0,0,0}\put(5851,-5011){\circle{336}}
}%
{\color[rgb]{0,0,0}\put(5851,-5611){\circle{336}}
}%
{\color[rgb]{0,0,0}\put(5851,-6211){\circle{336}}
}%
{\color[rgb]{0,0,0}\put(8251,-5011){\circle{336}}
}%
{\color[rgb]{0,0,0}\put(8251,-5611){\circle{336}}
}%
{\color[rgb]{0,0,0}\put(8251,-6211){\circle{336}}
}%
{\color[rgb]{0,0,0}\put(9751,-5011){\circle{336}}
}%
{\color[rgb]{0,0,0}\put(9751,-5611){\circle{336}}
}%
{\color[rgb]{0,0,0}\put(9751,-6211){\circle{336}}
}%
{\color[rgb]{0,0,0}\put(10801,-5011){\circle{336}}
}%
{\color[rgb]{0,0,0}\put(10801,-5611){\circle{336}}
}%
{\color[rgb]{0,0,0}\put(10801,-6211){\circle{336}}
}%
{\color[rgb]{0,0,0}\put(8056,-6406){\oval(210,210)[bl]}
\put(8056,-4816){\oval(210,210)[tl]}
\put(8446,-6406){\oval(210,210)[br]}
\put(8446,-4816){\oval(210,210)[tr]}
\put(8056,-6511){\line( 1, 0){390}}
\put(8056,-4711){\line( 1, 0){390}}
\put(7951,-6406){\line( 0, 1){1590}}
\put(8551,-6406){\line( 0, 1){1590}}
}%
{\color[rgb]{0,0,0}\put(5656,-6406){\oval(210,210)[bl]}
\put(5656,-4816){\oval(210,210)[tl]}
\put(6046,-6406){\oval(210,210)[br]}
\put(6046,-4816){\oval(210,210)[tr]}
\put(5656,-6511){\line( 1, 0){390}}
\put(5656,-4711){\line( 1, 0){390}}
\put(5551,-6406){\line( 0, 1){1590}}
\put(6151,-6406){\line( 0, 1){1590}}
}%
{\color[rgb]{0,0,0}\put(6256,-4606){\oval(210,210)[bl]}
\put(6256,-4216){\oval(210,210)[tl]}
\put(7846,-4606){\oval(210,210)[br]}
\put(7846,-4216){\oval(210,210)[tr]}
\put(6256,-4711){\line( 1, 0){1590}}
\put(6256,-4111){\line( 1, 0){1590}}
\put(6151,-4606){\line( 0, 1){390}}
\put(7951,-4606){\line( 0, 1){390}}
}%
{\color[rgb]{0,0,0}\put(6001,-4936){\vector( 1, 1){375}}
}%
{\color[rgb]{0,0,0}\put(7126,-4561){\vector( 1,-1){975}}
}%
{\color[rgb]{0,0,0}\put(8101,-6211){\vector(-1, 0){2100}}
}%
{\color[rgb]{0,0,0}\put(9556,-6406){\oval(210,210)[bl]}
\put(9556,-4816){\oval(210,210)[tl]}
\put(9946,-6406){\oval(210,210)[br]}
\put(9946,-4816){\oval(210,210)[tr]}
\put(9556,-6511){\line( 1, 0){390}}
\put(9556,-4711){\line( 1, 0){390}}
\put(9451,-6406){\line( 0, 1){1590}}
\put(10051,-6406){\line( 0, 1){1590}}
}%
{\color[rgb]{0,0,0}\put(10606,-6406){\oval(210,210)[bl]}
\put(10606,-4816){\oval(210,210)[tl]}
\put(10996,-6406){\oval(210,210)[br]}
\put(10996,-4816){\oval(210,210)[tr]}
\put(10606,-6511){\line( 1, 0){390}}
\put(10606,-4711){\line( 1, 0){390}}
\put(10501,-6406){\line( 0, 1){1590}}
\put(11101,-6406){\line( 0, 1){1590}}
}%
{\color[rgb]{0,0,0}\put(9901,-5011){\vector( 1, 0){750}}
}%
{\color[rgb]{0,0,0}\put(10651,-6211){\vector(-1, 0){750}}
}%
\end{picture}%
\nop{
+-> 3 -> 3 -> 3 -+
|                |
+----------------+
                                .
+-> 2 -> 2 -+
|           |
+-----------+
}

Looking at every single part of the formula in isolation it looks single-head
equivalent, since each equivalent set can entail another. The problem is
producing a single loop that reaches all of them. This shows that single-head
equivalence is not a local property that can be checked by looking only at the
individual equivalent sets of variables.

\end{enumerate}

\section{Python implementation}
\label{python}

The {\tt reconstruct.py} Python program implements the algorithm described in
Sections~\ref{construction} and~\ref{iteration}. It is available at
{} {\tt https://github.com/paololiberatore/reconstruct}.

It takes either a sequence of clauses or a file name as commandline options.
Formulae like {\tt ab->cd} and {\tt df=gh} are translated into clauses.
Examples invocations are:

\begin{verbatim}
reconstruct.py -f 'ab->cd' 'df=gh'
reconstruct.py -t tests/chains.py
\end{verbatim}

After turning {\tt ab->cd} into $ab \rightarrow c$ and $ab \rightarrow d$ and
turning {\tt df=gh} into $df \rightarrow g$, $hg \rightarrow f$, $df
\rightarrow h$ and $hg \rightarrow d$, it removes the tautologies from them, if
any.

The reconstruction algorithm described in Section~\ref{construction} with the
iteration function in Section~\ref{iteration} is implemented in the {\tt
reconstruct(f)} function. The loops over the bodies of the formula and over the
candidate set of clauses for the iteration function are in this function. This
is because the optimized method for the latter uses a number of sets accumulated
by the former: the set of clauses under construction, their heads, their bodies
and the clauses of the original formula used so far. Only the calculation of
$RCN()$, $UCL()$, $HCLOSE()$ and the search for the minimal bodies are split
into separate functions: {\tt rcnucl(b, f)}, {\tt hclose(heads, usable)} and
{\tt minbodies(f)}. The first is copied verbatim from the implementation of the
incomplete algorithm described in a previous article~\cite{libe-20-b}.

The function {\tt rcnucl(b,f)} is first called on each body of the formula, and
the results stored. These are used to check for example $A \leq_F B$. This
condition is indeed the same as $F \models B \rightarrow A$, which is the same
as $BCN(A,F) \subseteq BCN(B,F)$, in turn equivalent to as $A \cup RCN(A,F)
\subseteq B \cup RCN(B,F)$.

For each precondition $B$, the candidate values for $ITERATION(B,F)$ are
produced by first calculating the required heads $HEADS(B,F)$ and attaching the
allowed bodies to them in all possible ways. All property of valid iteration
functions hold by construction except $G \cup ITERATION(B,F) \models BCL(B,F)$,
which is checked as 
{} $HCLOSE(RCN(B,F),UCL(B,F)) = HCLOSE(RCN(B,G \cup IT),UCL(B,G \cup IT))$
using {\tt hclose(heads, usable)} and {\tt rcnucl(b, f)}.

The algorithm is complete, but slower than the incomplete algorithm presented
in the previous article~\cite{libe-20-b}. The number of candidate sets for
$ITERATION(B,F)$ may exponentially increase with the number of variables. This
happens when a set of variables $B$ is equivalent to many minimal equivalent
sets. Such equivalences are due to loops of clauses. They are the reason why
the previous algorithm is incomplete, and they make the present one exponential
in time.

%input{segregate.tex}

\section{Conclusions}
\label{conclusions}

The algorithm for turning a formula in single-head form if possible is correct
and complete, but may take exponential time. It complements the previous
algorithm, which is polynomial but incomplete~\cite{libe-20-b}, rather than
overcoming it. Which of them is the best depends on the goal.

If the goal is to reduce the size of forgetting rather than computing it
quickly, completeness wins over efficiency. The formulae that are not turned
into single-head form by the previous algorithm are by the present one.
Forgetting from them produces polynomially-sized formulae. Running time is
sacrificed on the altar of size reduction.

If the goal is to forget quickly, spending exponential time on preprocessing is
self-harming. Forgetting may take polynomial time on the single-head formula,
but producing it took exponential time. Better spend less time preparing the
formula, and rather jump straight to forgetting.

The present algorithm may still be of use in the second case. What it does is
to reconstruct the formula by groups of clauses of equivalent bodies. If a
group takes too long, the reconstruction can be cut short by just producing the
clauses with these bodies in the original formula. This part of the output
formula may not be single-head, but the rest will. Since the efficiency of
forgetting depends on the number of duplicated heads, this is still an
improvement.

The two algorithms can be applied in turn. The old computes a formula
$SHMIN(F)$ that is always single-head but may not be equivalent to the
original. Only if it is not the second, slower algorithm is run. It may take
advantage of $SHMIN(F)$. When searching for the clauses with a certain set of
heads, the ones from $SHMIN(F)$ may be tested first.

Given that the only source of exponentiality in time are cycles of clauses, it
makes sense to simplify them as much as possible. A way to do it is to merge
equivalent sets of variables. As an example, if the formula under construction
entails the equivalence of $A$ and $B$, only one of them is really necessary.
For example, every occurrence of $B$ can be replaced by $A$. This is a
simplification because it negates the need to test both $(A \cup \{y\})
\rightarrow x$ and $(B \cup \{y\}) \rightarrow x$ as possible clauses for the
head $x$. A followup article~\cite{libe-20-c} further analyzes this trick.

A specific formula that turned out important during the analysis of the
algorithm is
{} $F = \{x \rightarrow a, a \rightarrow d, x \rightarrow b, b \rightarrow c,
{}        ac \rightarrow x, bd \rightarrow x \}$,
in the {\tt disjointnotsingle.py} test file of the {\tt singlehead.py} program.

%
%  two cycles of single variables, running in different directions 
%
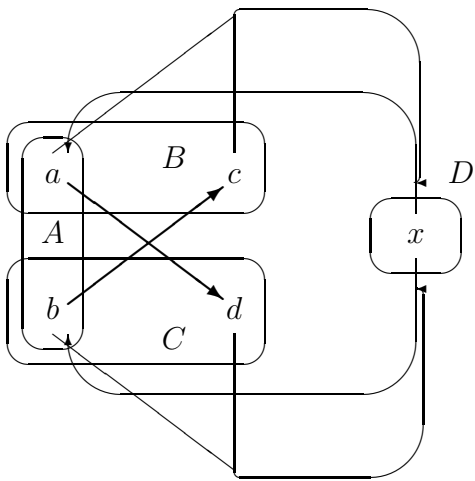
\setlength{\unitlength}{5000sp}%
\begingroup\makeatletter\ifx\SetFigFont\undefined%
\gdef\SetFigFont#1#2#3#4#5{%
  \reset@font\fontsize{#1}{#2pt}%
  \fontfamily{#3}\fontseries{#4}\fontshape{#5}%
  \selectfont}%
\fi\endgroup%
\begin{picture}(2277,2342)(7564,-6931)
\thinlines
{\color[rgb]{0,0,0}\put(7876,-6211){\vector( 0, 1){0}}
\put(8614,-6211){\oval(1476,600)[bl]}
\put(8614,-5836){\oval(1974,1350)[br]}
}%
{\color[rgb]{0,0,0}\put(8627,-5611){\oval(1948,1200)[tr]}
\put(8627,-5311){\oval(1502,600)[tl]}
\put(7876,-5311){\vector( 0,-1){0}}
}%
{\color[rgb]{0,0,0}\put(8926,-6886){\oval(450, 74)[bl]}
\put(8926,-6211){\oval(1424,1424)[br]}
\put(9601,-6211){\oval( 74,450)[tr]}
\put(9601,-5986){\vector(-1, 0){0}}
}%
{\color[rgb]{0,0,0}\put(9601,-5461){\vector(-1, 0){0}}
\put(9601,-5294){\oval( 40,334)[br]}
\put(8926,-5294){\oval(1390,1390)[tr]}
\put(8926,-4636){\oval(450, 74)[tl]}
}%
{\color[rgb]{0,0,0}\put(7756,-6181){\oval(210,210)[bl]}
\put(7756,-5341){\oval(210,210)[tl]}
\put(7846,-6181){\oval(210,210)[br]}
\put(7846,-5341){\oval(210,210)[tr]}
\put(7756,-6286){\line( 1, 0){ 90}}
\put(7756,-5236){\line( 1, 0){ 90}}
\put(7651,-6181){\line( 0, 1){840}}
\put(7951,-6181){\line( 0, 1){840}}
}%
{\color[rgb]{0,0,0}\put(9481,-5806){\oval(210,210)[bl]}
\put(9481,-5641){\oval(210,210)[tl]}
\put(9721,-5806){\oval(210,210)[br]}
\put(9721,-5641){\oval(210,210)[tr]}
\put(9481,-5911){\line( 1, 0){240}}
\put(9481,-5536){\line( 1, 0){240}}
\put(9376,-5806){\line( 0, 1){165}}
\put(9826,-5806){\line( 0, 1){165}}
}%
{\color[rgb]{0,0,0}\put(7681,-5506){\oval(210,210)[bl]}
\put(7681,-5266){\oval(210,210)[tl]}
\put(8746,-5506){\oval(210,210)[br]}
\put(8746,-5266){\oval(210,210)[tr]}
\put(7681,-5611){\line( 1, 0){1065}}
\put(7681,-5161){\line( 1, 0){1065}}
\put(7576,-5506){\line( 0, 1){240}}
\put(8851,-5506){\line( 0, 1){240}}
}%
{\color[rgb]{0,0,0}\put(7681,-6256){\oval(210,210)[bl]}
\put(7681,-5941){\oval(210,210)[tl]}
\put(8746,-6256){\oval(210,210)[br]}
\put(8746,-5941){\oval(210,210)[tr]}
\put(7681,-6361){\line( 1, 0){1065}}
\put(7681,-5836){\line( 1, 0){1065}}
\put(7576,-6256){\line( 0, 1){315}}
\put(8851,-6256){\line( 0, 1){315}}
}%
\thicklines
{\color[rgb]{0,0,0}\put(7876,-6061){\vector( 4, 3){768}}
}%
{\color[rgb]{0,0,0}\put(7876,-5461){\vector( 4,-3){768}}
}%
\thinlines
{\color[rgb]{0,0,0}\put(7801,-5311){\line( 4, 3){900}}
\put(8701,-4636){\line( 0,-1){675}}
}%
{\color[rgb]{0,0,0}\put(7801,-6211){\line( 4,-3){900}}
\put(8701,-6886){\line( 0, 1){675}}
}%
\put(7801,-5461){\makebox(0,0)[b]{\smash{{\SetFigFont{12}{24.0}
{\rmdefault}{\mddefault}{\updefault}{\color[rgb]{0,0,0}$a$}%
}}}}
\put(8701,-5461){\makebox(0,0)[b]{\smash{{\SetFigFont{12}{24.0}
{\rmdefault}{\mddefault}{\updefault}{\color[rgb]{0,0,0}$c$}%
}}}}
\put(7801,-6136){\makebox(0,0)[b]{\smash{{\SetFigFont{12}{24.0}
{\rmdefault}{\mddefault}{\updefault}{\color[rgb]{0,0,0}$b$}%
}}}}
\put(8701,-6136){\makebox(0,0)[b]{\smash{{\SetFigFont{12}{24.0}
{\rmdefault}{\mddefault}{\updefault}{\color[rgb]{0,0,0}$d$}%
}}}}
\put(9601,-5761){\makebox(0,0)[b]{\smash{{\SetFigFont{12}{24.0}
{\rmdefault}{\mddefault}{\updefault}{\color[rgb]{0,0,0}$x$}%
}}}}
\put(9826,-5461){\makebox(0,0)[b]{\smash{{\SetFigFont{12}{24.0}
{\rmdefault}{\mddefault}{\updefault}{\color[rgb]{0,0,0}$D$}%
}}}}
\put(8401,-6286){\makebox(0,0)[b]{\smash{{\SetFigFont{12}{24.0}
{\rmdefault}{\mddefault}{\updefault}{\color[rgb]{0,0,0}$C$}%
}}}}
\put(8401,-5386){\makebox(0,0)[b]{\smash{{\SetFigFont{12}{24.0}
{\rmdefault}{\mddefault}{\updefault}{\color[rgb]{0,0,0}$B$}%
}}}}
\put(7801,-5761){\makebox(0,0)[b]{\smash{{\SetFigFont{12}{24.0}
{\rmdefault}{\mddefault}{\updefault}{\color[rgb]{0,0,0}$A$}%
}}}}
\end{picture}%
\nop{
+----------------------+
|                      |
|   +--------+-----+   |
|   |        |     |   |
+-> a -\ +-> c     V --+               B=ac
        X    +---> x           A=ab            D=x
+-> b -/ +-> d     ^ --+               C=bd
|   |        |     |   |
|   +--------+-----+   |
|                      |
+----------------------+
}

It is not single-head because the four sets $A$, $B$, $C$ and $D$ are
equivalent, but only three heads are available: $a$, $b$ and $x$. The other two
variables are already the heads of $a \rightarrow d$ and $b \rightarrow c$.

A formula may also not be single-head equivalent for other reasons, but
examples for these cases are much simpler than this. They range from two to
three binary clauses. They are quite immediate. The relative size of this one,
with six clauses including two ternary ones, witnesses the complexity of this
case.

Given that the first algorithm is polynomial but incomplete and the second is
complete but not polynomial, a natural question is whether a complete and
polynomial algorithm exists. If so, the problem of recognizing whether a
formula is single-head equivalent would be tractable. This problem is in the
complexity class \np, but whether it is \np-complete or not is an open
question.

Another open question is the applicability of the algorithms outside the Horn
fragment of propositional logic.

In the general propositional case, resolving all occurrences of a variable
forgets the variable. This is the same as viewing a clause like $\neg a \vee b
\vee c$ as an implication $a \neg b \rightarrow c$. If no other occurrence of
$c$ is positive, it is forgotten by replacing every negative occurrence of it
with $a \neg b$. The size increase is only linear in the single-head case. The
question is whether the algorithms for turning a formula in single-head form
extends to the general propositional case.

The problem complicates when forgetting multiple variables, since the
replacement may multiply the number of positive occurrences of other variables,
like $a$ in this case. An alternative is to apply the replacing algorithm only
to the definite Horn clauses of the formula. In the example, a clause like
$\neg c \vee d \vee e$ would be left alone, and $c$ only replaced by definite
Horn clauses like $\neg c \vee \neg d \vee e$, equivalent to $dc \rightarrow
e$. The remaining occurrences of $c$ are then forgotten by resolving them out.
If the formula contains few non-Horn clauses, this procedure may be relatively
efficient.

This second mechanism would probably apply to every logic that includes the
definite Horn fragment. Information like ``this, this and this imply that'' is
common, so definite Horn clauses are likely to be present in large number.
Turning them into single-head formula would allow to efficiently forget them
before turning to the others.

\let\c=\cedilla
\bibliographystyle{alpha}
\newcommand{\etalchar}[1]{$^{#1}$}

\end{document}